%% file: main.tex
\documentclass[10pt,twocolumn,letterpaper]{article}

\usepackage[pagenumbers]{cvpr} 

\input{preamble}
\definecolor{cvprblue}{rgb}{0.21,0.49,0.74}
\usepackage[pagebackref,breaklinks,colorlinks,allcolors=cvprblue]{hyperref}

\usepackage{amsmath}
\usepackage{amssymb}
\usepackage{mathtools}
\usepackage{amsthm}
\usepackage{multirow}
\usepackage[table]{xcolor}
\usepackage[most]{tcolorbox}
\usepackage{microtype}
\usepackage{graphicx}
\usepackage{subcaption}
\usepackage{booktabs}
\usepackage{algorithm}
\usepackage{algorithmic}
\usepackage{listings}
\usepackage{tablefootnote}

\theoremstyle{plain}
\newtheorem{theorem}{Theorem}[section]

\theoremstyle{definition}
\newtheorem{definition}[theorem]{Definition}

\theoremstyle{remark}

\theoremstyle{definition}
\newtheorem{observation}[theorem]{Observation}
\tcolorboxenvironment{observation}{
    breakable,
    colback=gray!10!white,
    boxrule=0pt
}
\tcolorboxenvironment{theorem}{
    breakable,
    colback=gray!10!white,
    boxrule=0pt
}

\makeatletter

\newcommand{\FUNCTION}[1]{
    \STATE \textbf{function} #1
}

\newcommand{\ENDFUNCTION}{
    \STATE \textbf{end function}%
}
\makeatother

\title{GraDE: A Graph Diffusion Estimator \\ for Frequent Subgraph Discovery in Neural Architectures}

\author{
  \textbf{Yikang Yang}$^{1,3}$ \quad
  \textbf{Zhengxin Yang}$^{1,2}$\thanks{Zhengxin Yang is the corresponding author} \quad
  \textbf{Minghao Luo}$^{4}$ \quad
  \textbf{Luzhou Peng}$^{1,3}$ \\
  \textbf{Hongxiao Li}$^{1,3}$ \quad
  \textbf{Wanling Gao}$^{1,2}$ \quad
  \textbf{Lei Wang}$^{1,2}$ \quad
  \textbf{Jianfeng Zhan}$^{1,2}$ \\
  Institute of Computing Technology, Chinese Academy of Sciences$^{1}$\\
  BenchCouncil (International Open Benchmark Council)$^{2}$\\
  University of Chinese Academy of Sciences$^{3}$\\
  Department of Computer Science McCormick School of Engineering Northwestern University Evanston, IL, USA$^{4}$\\
  \texttt{\{yangyikang23s, yangzhengxin\}@ict.ac.cn}
}

\begin{document}
\maketitle
\begin{abstract}
    Finding frequently occurring subgraph patterns or network motifs in neural architectures is crucial for optimizing efficiency, accelerating design, and uncovering structural insights.
    However, as the subgraph size increases, enumeration-based methods are perfectly accurate but computationally prohibitive, while sampling-based methods are computationally tractable but suffer from a severe decline in discovery capability.
    To address these challenges, this paper proposes GraDE, a diffusion-guided search framework that ensures both computational feasibility and discovery capability.
    The key innovation is the $\underline{\textnormal{Gra}}\textnormal{ph}$  $\underline{\textnormal{D}}\textnormal{iffusion}$ $\underline{\textnormal{E}}\textnormal{stimator}$ (GraDE), which is the first to introduce graph diffusion models to identify frequent subgraphs by scoring their typicality within the learned distribution.
    Comprehensive experiments demonstrate that the estimator achieves superior ranking accuracy, with up to 114\% improvement compared to sampling-based baselines.
    Benefiting from this, the proposed framework successfully discovers large-scale frequent patterns, achieving up to 30$\times$ higher median frequency than sampling-based methods.
\end{abstract}
\input{sections/Introduction.tex}
\input{sections/Related_Work.tex}
\input{sections/Preliminary.tex}
\input{sections/Method.tex}
\input{sections/Experiments.tex}
\input{sections/Conclusion.tex}
{
    \small
    \bibliographystyle{ieeenat_fullname}
    \bibliography{main}
}

\newpage
\appendix
\onecolumn

\input{sections/Appendix}

\end{document}

%% file: sections/Introduction.tex
\section{Introduction}

\begin{figure*}[!t]
    \centering
    \includegraphics[width=0.86\textwidth]{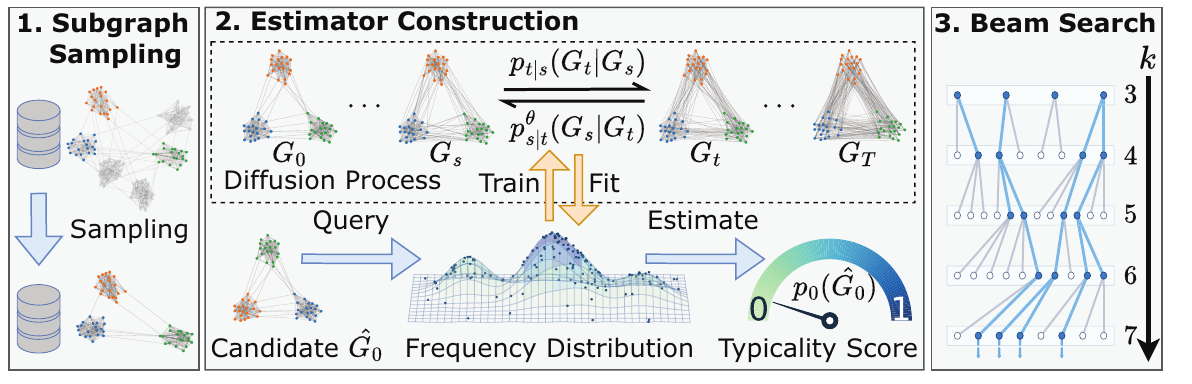}
    \caption{Overview of the GraDE framework. The proposed pipeline operates in three stages: (1) \textbf{Subgraph Sampling}, where a representative training set is collected from target architectures; (2) \textbf{Estimator Construction}, where a graph diffusion model is trained to fit the subgraph frequency distribution, enabling the GraDE estimator to serve as a frequency surrogate by scoring the structural typicality of any query candidate instance; (3) \textbf{Beam Search}, where motifs are discovered by iteratively expanding candidates and leveraging GraDE scores to retain only the top-$N$ promising subgraphs at each size $k$ to prune the search space.}
    \label{fig:method_overview}
\end{figure*}

Finding frequent subgraph patterns or network motifs~\cite{network_motifs, ying_representation_2024} is significant for characterizing the fundamental structural properties of neural architectures.
From a computational perspective, these motifs reveal recurring execution patterns that are essential for optimizing efficiency~\cite{salameh_autogo_2023, Plus}.
From a design perspective, the discovery of such patterns provides a modular library of reusable components to accelerate architecture design~\cite{pmlr-v119-li20c, 10013693}.
From an analytical perspective, these motifs serve as structural fingerprints that uncover the intrinsic regularities governing neural networks~\cite{wan2022on, yu2025mage}.
These collective advantages advance both theoretical understanding and the practical engineering of neural networks.

However, finding network motifs is extremely complex, inherently challenged by a trade-off between accuracy and scalability.
Enumeration-based methods offer absolute accuracy by exhaustively counting all possible subgraphs of size $k$~\cite{hocevar_combinatorial_2014, ying_representation_2024}. 
Unfortunately, this process is NP-hard, becoming computationally prohibitive as $k$ increases~\cite{ribeiro_survey_2022}.
Sampling-based methods attempt to ensure scalability by discovering patterns from a sampled subset of subgraphs~\cite{wernicke_efficient_2006, ribeiro_efficient_2010, lu_sampling_2012, paredes_rand-fase_2015}.
Although these methods are computationally tractable, they suffer from a severe decline in discovery capability as $k$ increases.
Specifically, as the search space expands exponentially with $k$, the extreme sparsity of samples leads to high variance and poor representativeness, causing a sharp decline in discovery capability.

To address these challenges, this paper proposes GraDE, a diffusion-guided search framework to ensure computational feasibility and high discovery capability.
The core of this framework is the \underline{Gra}ph \underline{D}iffusion \underline{E}stimator (GraDE), which is inspired by the capability of generative models in distribution learning~\cite{BojchevskiSZG18, DeCao2018MolGANAI, Simonovsky2018GraphVAETG, LiuKBKS19, NiuSSZGE20, Shi2021LearningGF, JoLH22, vignac2023digress, xu_discrete-state_2024, qin2025defog}.
This estimator is built upon the key insight that generative models can be inverted to grade structural typicality of any instance, thereby serving as a reliable surrogate for subgraph frequency.
Unlike sampling-based methods that treat each subgraph as an isolated entity, GraDE uses graph diffusion models to capture structural correlations, thereby generalizing from sparse data to a continuous probabilistic landscape.

As shown in Figure \ref{fig:method_overview}, the GraDE framework follows a three-stage pipeline:
\textbf{(1) Subgraph Sampling} collects a representative training set from the architecture distribution;
\textbf{(2) Estimator Construction} builds the Graph Diffusion Estimator by training a graph diffusion model to capture the structural distribution;
\textbf{(3) Beam Search} discovers motifs by iteratively expanding candidates and pruning unpromising branches via GraDE scores.
By replacing exponential subgraph counting with efficient model inference, GraDE ensures computational feasibility for large-scale patterns.

Comprehensive experiments are conducted on synthesized datasets including NAS-Bench-101~\cite{Ying2019NASBench101TR}, NAS-Bench-201~\cite{Dong020}, NAS-Bench-301~\cite{9762315}, and NAS-Bench-NLP~\cite{zela2022surrogatenasbenchmarksgoing}, and a real-world Younger dataset~\cite{Yang2024YoungerTF}.
The evaluation firstly confirms the effectiveness of the graph diffusion estimator as a subgraph frequency surrogate.
Results show that GraDE achieves significantly higher ranking accuracy than baseline methods, especially under sample-sparse conditions, with Spearman's rank correlation coefficient improving by up to 114\% on synthesized datasets and 33\% on the real-world dataset.
When integrated into the search pipeline, GraDE identifies motifs with up to 30$\times$ higher median frequency than sampling-based methods for large-sized subgraphs.
These results collectively demonstrate the superior discovery capability of the proposed framework.

The main contributions of this paper are:
(1) a novel \textbf{subgraph frequency estimator}, GraDE, which is the first to repurpose graph diffusion models for grading structural typicality; (2) a \textbf{search framework} that integrates GraDE with a beam search strategy to ensure computational feasibility and high discovery capability; (3) \textbf{comprehensive experiments} across various datasets confirming the effectiveness of both the estimator and the framework. 

%% file: sections/Related_Work.tex
\section{Related Work}
\label{sec:Related Work}
\subsection{Network Motif Discovery}
Network motifs are high-frequency and representative subgraph patterns in a graph~\cite{network_motifs}. 
Traditionally, discovery methods are divided into two primary paradigms: enumeration-based and sampling-based methods.

Enumeration-based methods, represented by ESU~\cite{wernicke_efficient_2006}, count all subgraphs of size $k$ to get exact results. 
While providing absolute precision, these methods suffer from exponential computational complexity. 
Although subsequent optimizations~\cite{Gtries, FASE, SCMD} have been proposed to accelerate execution, they remain fundamentally constrained by the growth of the search space, making exact enumeration intractable for large subgraphs.

Sampling-based methods were introduced to solve these efficiency issues. 
Many of these approaches add random sampling to the enumeration process to avoid high costs while keeping the results unbiased, such as Rand-ESU~\cite{wernicke_efficient_2006}, Rand-FaSE~\cite{paredes_rand-fase_2015}, and Rand-Gries~\cite{ribeiro_efficient_2010}. 
Other methods sample subgraphs based on local structures~\cite{kashtan_efficient_2004, saha_finding_2015}, such as NRS~\cite{lu_sampling_2012}, but these are often biased. 
However, as $k$ increases, the subgraph space grows exponentially, causing the sampled data to become extremely sparse. 
This sparsity leads to a significant decline in discovery capability, making these methods unreliable for identifying large-scale motifs in complex neural architecture datasets.

\subsection{Graph Generative Models}
Graph generative models aim to learn the distribution of graph data to generate new samples with similar structures~\cite{Fan2023GenerativeDM}.
Early approaches include GraphVAE~\cite{Simonovsky2018GraphVAETG}, which uses variational inference for graph encoding and decoding, and GraphRNN~\cite{You2018GraphRNNGR}, which treats generation as a sequential decision process.
Other methods leverage adversarial training, such as NetGAN~\cite{BojchevskiSZG18} and MolGAN~\cite{DeCao2018MolGANAI}, or utilize invertible mappings in normalizing flow-based models~\cite{LiuKBKS19}.

Recently, diffusion models have achieved state-of-the-art performance in graph generation.
Initial diffusion-based works~\cite{NiuSSZGE20, Shi2021LearningGF} directly applied Gaussian noise to adjacency matrices, which often destroyed the discrete structural constraints of graphs. 
To resolve this, DiGress~\cite{vignac2023digress} introduced discrete diffusion processes, adding noise to categorical node and edge attributes to better preserve structural fidelity.
Based on this, DisCo and other methods~\cite{xu_discrete-state_2024, siraudin2025cometh} extended the process into continuous time, significantly enhancing the flexibility and efficiency of the generation process.
More recently, research has shifted toward flow-matching methods~\cite{qin2025defog}, which define vector fields to transform noise into target distributions.

Despite their success in generation, these models have not yet been utilized as a subgraph frequency estimator for motif discovery, which is the focus of this work.

%% file: sections/Preliminary.tex
\section{Preliminaries}
\label{sec:Preliminaries}

This work studies directed graphs whose nodes and edges have categorical attributes.
A $k$-(sub)\textit{graph} $G$ is defined as a graph consisting of exactly $k$ nodes\footnote{For clarity, calligraphic typeface ($\mathcal{G}$) denotes sets; bold typeface ($\mathbf{G}$) denotes random variables, vectors or matrices; and non-bold typeface ($G$) denotes specific values or instances.}.
Such a $k$-graph is formally represented by a \textit{node type matrix} $X=(x^{i})_{1 \leq i \leq k} \in \{1, 2, \dots, a\}^k$ and an \textit{edge type matrix} $E=(e^{ij})_{1 \leq i, j \leq k} \in \{1, 2, \dots, b\}^{k \times k}$.
The scalars $a$ and $b$ denote the number of node and edge types, respectively, while the absence of an edge is regarded as an edge type.

The remainder of this section presents the research objective and the theoretical foundation.
To facilitate these discussions, several necessary graph concepts are introduced below.
A (sub)graph is \textit{connected} if every pair of nodes is linked by a path, regardless of edge directions.
A subgraph $S$ of $G$ is \textit{induced} if every edge of $G$ between nodes in $S$ is also an edge of $S$.
Two graphs are \textit{isomorphic} if there is a one-to-one mapping between their node sets that preserves all node and edge attributes.
An \textit{induced occurrence} of $S$ in $G$ refers to a specific instance of an induced subgraph within $G$ that is isomorphic to $S$.

\subsection{Research Objective: Frequent Subgraph Discovery}
This work aims to discover frequent subgraphs within a target graph set.
To formally characterize this objective, it is necessary to define the subgraph counting problem:
\begin{definition}[Subgraph Counting Problem]
  \label{def:subgraph_counting}
  Given a (sub)graph $S$ and a graph $G$, the \textit{subgraph counting problem}, denoted by $\text{SC}(S, G)$, is to determine the number of induced occurrences of $S$ in $G$.
  Two occurrences are considered distinct if their node sets differ by at least one node.
\end{definition}
For a graph set $\mathcal{G}$, the subgraph counting problem extends to the total occurrences found across all $G \in \mathcal{G}$:
\begin{equation}
  \label{eq:SC}
  \text{SC}(S, \mathcal{G})=\sum_{G \in \mathcal{G}} \text{SC}(S, G).
\end{equation}

This work uses $\mathcal{D}_k$ to denote all connected induced $k$-subgraphs in $\mathcal{G}$, which can be partitioned into multiple sets, such that instances are isomorphic within the same set and non-isomorphic across different sets.
Each set is represented by one subgraph pattern shared by its instances, and the collection of these representatives is denoted by $\mathcal{S}_k=\{S_k^1, S_k^2, \dots\}$.
The \textit{relative frequency} of the $i$-th subgraph pattern $S_k^i$ is defined as:
\begin{equation}
    \label{eq:RF}
    \text{RF}_k(S_k^i, \mathcal{G})=\text{SC}(S_k^i, \mathcal{G}) \Big(\sum_{j=1}^{|\mathcal{S}_k|}\text{SC}(S_k^j, \mathcal{G})\Big)^{-1} .
\end{equation}

The research objective is thus defined as:
{
\tcolorboxenvironment{definition}{
      colback=gray!10!white,
      boxrule=0pt,
      breakable
  }
\begin{definition}[Frequent Subgraph Discovery]
    Given a graph set $\mathcal{G}$ and a subgraph size $k$, the goal is to efficiently find subgraphs $S\in\mathcal{S}_k$ with high $\text{RF}_k$.
\end{definition}
}
\subsection{Theoretical Foundation: Graph Diffusion Models}
\label{subsec:GDE}

Graph diffusion models provide a generative framework that transforms samples from a simple prior distribution $\mathbf{G}_T \sim p_T$ (e.g., Gaussian, uniform, or marginal distributions~\cite{vignac2023digress}) into the target distribution $\mathbf{G}_0 \sim p_0$ via a learned reverse process.
For any graph instance $G_0$, its generative probability can be estimated as the expectation over a latent noising trajectory $\{\mathbf{G}_t\}_{t \in \mathcal{T}}$:
\begin{equation}
  \label{eq:pG0}
  p_{0}(G_0) \!=\! \mathbb{E}_{\substack{\mathbf{G}_{\Delta t}, \dots, \mathbf{G}_T \\ \sim p_{\Delta t, \dots, T|0}}} \!\left[ p_T(\mathbf{G}_T) \!\prod_{t \in \mathcal{T}}\! \frac{ p_{t-\Delta t|t}(\mathbf{G}_{t-\Delta t}|\mathbf{G}_{t})}{p_{t|t-\Delta t}(\mathbf{G}_{t}|\mathbf{G}_{t-\Delta t})} \!\right]\!,
\end{equation}
where $\mathcal{T}=\{\Delta t, \dots, T\}$ is the time steps.
Specifically, $\Delta t = 1$ for discrete time and $\Delta t \in \mathbb{R}^+$ for continuous time.
This equation provides a formal way to estimate $p_0(G_0)$.

%% file: sections/Method.tex
\section{Method}
\label{sec:Method}

For a graph set $\mathcal{G}$ and its corresponding collection of connected $k$-subgraphs $\mathcal{D}_k$, finding subgraphs with high relative frequency requires evaluating $\text{RF}_k(G_0, \mathcal{G})$ for candidates $G_0\in\mathcal{D}_k$, \footnote{{In this section, $G_0$ denotes a $k$-subgraph instance from $\mathcal{D}_k$, not a full graph from $\mathcal{G}$.}} which is computationally NP-hard.
GraDE circumvents this challenge based on the following observation:
\begin{observation}[Key Insight]
    The generative probability $p_0(G_0)$ of a graph diffusion model trained on $\mathcal{D}_k$ serves as a direct surrogate for $\text{RF}_k(G_0, \mathcal{G})$.
\end{observation}
Accordingly, GraDE constructs a subgraph frequency estimator based on graph diffusion models.

The remainder of this section is organized as follows.
Section \ref{subsec:subgraph_sampling} describes how to prepare the training set for the estimator by subgraph sampling.
Section \ref{subsec:GDM_estimator} details the graph diffusion estimator, providing the theoretical foundation for GraDE.
Section \ref{subsec:realization} presents a realization of the estimator by factorization and parameterization.
Section \ref{subsec:mining_framework} introduces a beam search framework that leverages this estimator for frequent subgraph discovery.

\subsection{Subgraph Sampling}
\label{subsec:subgraph_sampling}
Training the graph diffusion estimator requires the dataset $\mathcal{D}_k$. 
However, as the subgraph size $k$ increases, the scale of $\mathcal{D}_k$ grows exponentially, making it impossible to obtain the complete $\mathcal{D}_k$.
Fortunately, the estimator can fit the global distribution using only a representative subset of $\mathcal{D}_k$, thereby bypassing the requirement for exhaustive enumeration.

GraDE employs subgraph sampling algorithms to construct the representative subset of $\mathcal{D}_k$.
By ensuring each element in $\mathcal{D}_k$ is sampled with equal probability, the resulting subset reflects the distribution of subgraphs in $\mathcal{D}_k$.
Rather than designing a new sampling algorithm, GraDE treats the sampling module as a pluggable component.
This modular design enables the seamless integration of various established uniform subgraph sampling algorithms, such as Rand-ESU~\cite{wernicke_efficient_2006}, ARS, NRS~\cite{lu_sampling_2012}, and Rand-FaSE~\cite{paredes_rand-fase_2015}.
Details of these algorithms are provided in Appendix \ref{sec:im_ssm}.

\subsection{Graph Diffusion Estimator}
\label{subsec:GDM_estimator}

In principle, various graph diffusion models can be adapted to serve as estimators under the GraDE framework.
In this work, the estimator is concretely instantiated into three versions based on representative discrete diffusion models: DisCo~\cite{xu_discrete-state_2024}, DiGress~\cite{vignac2023digress}, and DeFoG~\cite{qin2025defog}.
For conciseness, the following derivation focuses on the estimator based on DisCo (denoted as DisCo-E), while the formulations for alternative versions deferred to Appendix \ref{DiGress_E} and \ref{DeFoG_E}.

\textbf{Forward Transition Probability.}
The graph diffusion process is defined over a discrete state space $\Omega = \{1, \dots, a\}^k \times \{1, \dots, b\}^{k^2}$, which is formed by the Cartesian product of the state spaces for $k$ nodes and $k^2$ edges.
In this space, the noising process is modeled as a continuous-time Markov chain (CTMC) $\{\mathbf{G}_t, t \in [0, T]\}$.

This CTMC is controlled by a forward rate matrix $\mathbf{R}_t \in \mathbb{R}^{|\Omega| \times |\Omega|}$, where each entry $\mathbf{R}_t(G, \tilde{G})$ denotes the instantaneous transition rate from state $G$ to $\tilde{G}$:
\begin{equation}
    \mathbf{R}_t(G, \tilde{G}) = \lim_{\Delta t \to 0} \cfrac{p_{t+\Delta t|t}(\tilde{G}|G) - \delta(G, \tilde{G})}{\Delta t},
\end{equation}
where $p_{t+\Delta t|t}(\tilde{G}|G)$ is the conditional probability that $\mathbf{G}_{t+\Delta t} = \tilde{G}$ given $\mathbf{G}_t = G$, and $\delta(G, \tilde{G})$ is the Kronecker delta.
For $s < t$, the transition probability $p_{t|s}(G_t|G_s)$ from $G_s$ to $G_t$ follows the Kolmogorov forward equation:
\begin{equation}
    \label{eq:KF}
    \frac{d}{dt}p_{t|s}(G_t|G_s) = \sum_{G \in \Omega} p_{t|s}(G|G_s) \mathbf{R}_t(G, G_t),
\end{equation}

Assuming the forward rate matrices commute at different times (i.e., $\mathbf{R}_t \mathbf{R}_s = \mathbf{R}_s \mathbf{R}_t$ for any $t, s \geq 0$), Eq. \eqref{eq:KF} has a closed-form solution\footnote{For any matrix $\mathbf{M}$, the notation $(\mathbf{M})_{ij}$ denotes the entry at the $i$-th row and $j$-th column, where indices $i$ and $j$ correspond to the states in the discrete space $\Omega$.} (see proofs in Appendix \ref{proof_fp}):
\begin{equation}
    \label{eq:FP}
    p_{t|s}(G_t|G_s)=\Big(\exp({\int_{s}^{t}\mathbf{R}_u du})\Big)_{G_sG_t}, 
\end{equation}
which provides the explicit form for the forward transition probability $p_{t|t-\Delta t}$ required in Eq. \eqref{eq:pG0}.

\textbf{Reverse Transition  Probability.}
Formulating $p_{t-\Delta t|t}$ in Eq. \eqref{eq:pG0} requires reversing the noising process. 
This reverse process is also a CTMC, governed by a reverse rate matrix $\hat{\mathbf{R}}_t \in \mathbb{R}^{|\Omega| \times |\Omega|}$.
Under the detailed balance condition, the entries of this matrix are given by:
\begin{equation}
    \label{eq:DB}
    \hat{\mathbf{R}}_t(\tilde{G}, G) = \cfrac{p_t(G)\mathbf{R}_t(G, \tilde{G})}{p_t(\tilde{G})}, 
\end{equation}
where $p_t(G)$ denotes the marginal probability $P(\mathbf{G}_t = G)$.

As $p_t$ is typically intractable, $\hat{\mathbf{R}}_t$ lacks a closed-form expression, making $p_{s|t}$ inexpressible as a matrix exponential like Eq. \eqref{eq:FP}.
Therefore, in this work, by leveraging the connection between the reverse CTMC and a Poisson process, $p_{s|t}$ is derived for a sufficiently small interval $\Delta t = t - s$ as follows (see proofs in Appendix \ref{proof_rp}):
\begin{equation}
    \label{eq:RP}
    p_{s|t}(G_s|G_t)\!\!=\!\!\begin{cases}\!\! 
    \Big(1 \!\!-\! \exp\!\big(\Delta t \hat{\mathbf{R}}_t(G_t,\!G_t)\big)\!\Big)\! \cfrac{\hat{\mathbf{R}}_t(G_t,\! G_s)}{-\hat{\mathbf{R}}_t(G_t,\! G_t)} \\
    \hspace{4em} \text{if } G_s \neq G_t \\
    \exp\Big({\Delta t \hat{\mathbf{R}}_t(G_t, G_t)}\Big) \\
    \hspace{4em} \text{if } G_s = G_t
    \end{cases}\!\!\!.    
\end{equation}
Intuitively, this approximation models the evolution over $\Delta t$ as a Poisson process.
The case $G_s = G_t$ represents no state transition.
For $G_s \neq G_t$, the probability is the product of at least one transition occurring, $1 - \exp(\Delta t \hat{\mathbf{R}}_t(G_t, G_t))$, and the conditional probability of jumping specifically to state $G_s$, i.e., the ratio of the transition rate to the total exit rate.

\textbf{Monte Carlo Estimation.}
Given the transition probabilities derived above, $p_0(G_0)$ in \eqref{eq:pG0} can be estimated via Monte Carlo trials.
Each Monte Carlo trial stochastically samples a noising path $\{\mathbf{G}_t\}_{t \in \mathcal{T}}$ from $G_0$, calculates its joint probability using Eq. \eqref{eq:FP} and \eqref{eq:RP}.
By averaging these path-wise joint probabilities over multiple trials, an estimate of $p_0(G_0)$ is obtained.
Alg. \ref{alg:Monte_Carlo_Estimation} in Appendix \ref{sec:Monte_Carlo} describes this procedure.

\subsection{Factorized Realization and Parameterization}
\label{subsec:realization}
While Section \ref{subsec:GDM_estimator} provides a formulation for estimating generative probabilities via Monte Carlo trials, it lacks concrete forms for the forward rate matrix $\mathbf{R}_t$ and its reverse counterpart $\hat{\mathbf{R}}_t$.
This section employs factorization and neural parameterization to provide a concrete implementation.

\textbf{Factorized Forward Process.}
Following existing practices~\cite{xu_discrete-state_2024}, this work applies noise independently to each node and edge.
All nodes and edges share the same forward rate matrices: $\mathbf{R}_t^{X} = \beta(t) \mathbf{R}_x$ and $\mathbf{R}_t^{E} = \beta(t) \mathbf{R}_e$.
Specifically, $\mathbf{R}_x = \mathbf{1}\mathbf{1}^{\top} - a\mathbf{I}$ and $\mathbf{R}_e = \mathbf{1}\mathbf{1}^{\top} - b\mathbf{I}$ are constant matrices, with the noise schedule given by $\beta(t)=\alpha \gamma^t \log (\gamma)$.
Such independence allows the forward transition probability to factorize into the product of individual node and edge transitions.
Solving the component-wise Kolmogorov forward equation(cf. Eq.~\eqref{eq:KF}), Theorem \ref{thm:factorized_forward} specifies the concrete form of the forward transition probability (see proofs in Appendix \ref{proof_ffp}).

\begin{theorem}[Forward Transition Probability]
\label{thm:factorized_forward}
    For any $0 \leq s < t \leq T$, the forward transition probability from $G_s$ to $G_t$ is given by:
    \begin{equation}
        p_{t|s}(G_t|G_s)=\prod_{i=1}^{k} p_{t|s}(x_t^i|x_s^i) \prod_{i, j=1}^{k} p_{t|s}(e_t^{ij}|e_s^{ij}),
    \end{equation}
    where the transition probability for the $i$-th node from state $x_s^i$ to $x_t^i$ is given by:
    \begin{equation}
        p_{t|s}(x_t^i|x_s^i)=\Big(\exp (\int_{s}^t \beta(u)\mathbf{R}_x du)\Big)_{x_s^ix_t^i}.
    \end{equation}
    The transition probability $p_{t|s}(e_t^{ij}|e_s^{ij})$ for the edge between nodes $i$ and $j$ is given analogously.
\end{theorem}

\textbf{Factorized Reverse Process.}
Similar to the forward process, the reverse graph transition decomposes into independent node and edge components.
Applying Bayes' theorem to component-wise forward rates yields the reverse rate matrix for node $i$ (proofs in Appendix \ref{proof_frp}):
\begin{equation}
    \label{eq:rRtx}
    \hat{\mathbf{R}}_t^i(x_t^i, \tilde{x}_t^i)=\beta(t)\mathbf{R}_x(\tilde{x}_t^i, x_t^i)\sum_{x_0^i}\cfrac{p_{t|0}(\tilde{x}_t^i|x_0^i)}{p_{t|0}(x_t^i|x_0^i)}p_{0|t}(x_0^i|G_t).
\end{equation}
The edge-specific reverse rate matrix $\hat{\mathbf{R}}_t^{ij}$ is given similarly. 
Notably, both reverse rate matrices depend on the entire graph $G_t$ via the posterior $p_{0|t}(\cdot|G_t)$, capturing structural context during denoising.
Theorem \ref{thm:factorized_reverse} specifies the reverse transition probability (proofs in Appendix \ref{proof_frp}).
\begin{theorem}[Reverse Transition Probability]
\label{thm:factorized_reverse}
    For a small time step $\Delta t = t - s$, the reverse transition probability from $G_t$ to $G_s$ is given by:
    \begin{equation}
        p_{s|t}(G_s|G_t)=\prod_{i=1}^{k} p_{s|t}(x_s^i|G_t) \prod_{i, j=1}^{k} p_{s|t}(e_s^{ij}|G_t)
    \end{equation}
    where the transition probability for the $i$-th node from state $x_t^i$ to $x_s^i$ is given by:
    \begin{equation}
        p_{s|t}(x_s^i|G_t)\!\! = \!\!
        \begin{cases} 
        \!\!\Big(\!\!\exp\!\Big(\!\Delta t \hat{\mathbf{R}}_t^i(x_t^i,\! x_t^i)\!\Big)\!\! -\!\! 1\!\Big)\! \cfrac{\hat{\mathbf{R}}_t^i(x_t^i, \!x_s^i)}{\hat{\mathbf{R}}_t^i(x_t^i, \!x_t^i)}\! \\
        \hspace{4em} \text{if } x_s^i \neq x_t^i \\[1ex]
        \exp\Big(\Delta t \hat{\mathbf{R}}_t^i(x_t^i, x_t^i)\Big) \\
        \hspace{4em} \text{if } x_s^i = x_t^i
        \end{cases}\!\!\!.
    \end{equation}
    The transition probability $p_{s|t}(e_s^{ij}|G_t)$ for the edge between nodes $i$ and $j$ is given analogously.
\end{theorem}

\textbf{Model Parameterization.}
Thus far, all derivations for the graph diffusion estimator are complete except for the posterior distributions $p_{0|t}(x_0^i|G_t)$ in Eq. \eqref{eq:rRtx} and $p_{0|t}(e_0^{ij}|G_t)$ in the edge reverse rate matrix. 
A neural network $\phi_{\theta}$ is employed to parameterize these distributions, denoted as $p_{0|t}^{\theta}(G_0|G_t)$.
By taking the noisy graph $G_t$ at time $t$ as input, the model predicts the categorical probabilities for each node and edge attribute of the original clean graph: \begin{equation} 
    \hat{x}_0^i = p_{0|t}^{\theta}(x_0^i|G_t), \quad \hat{e}_0^{ij} = p_{0|t}^{\theta}(e_0^{ij}|G_t) .
\end{equation}
The optimization of the network parameters $\theta$ is achieved by minimizing the following cross-entropy (CE) loss function:
\begin{equation} 
    \mathcal{L} = \mathbb{E}_{
        \begin{subarray}{l} 
        t \sim \mathcal{U}(0,T) \\ 
        \mathbf{G}_0 \sim p_0 \\ 
        \mathbf{G}_t \sim p_{t|0} 
        \end{subarray}
        } \left[ \sum_{i=1}^{k} \text{CE}(x_0^i, \hat{x}_0^i) + \lambda\sum_{i,j=1}^{k} \text{CE}(e_0^{ij}, \hat{e}_0^{ij}) \right] .
\end{equation}
This work adopts the Graph Transformer architecture as the backbone for $\phi_{\theta}$; further implementation details regarding the model architecture are provided in Appendix \ref{ma}.

\begin{algorithm}
\caption{Beam Search}
\label{alg:mining}
    \begin{algorithmic}
    \STATE {\bfseries Input:} Graph set $\mathcal{G}$, target size $k_{max}$, top-$N$ threshold $N$, graph diffusion estimator $p^{\mathbf{\theta}}$.
    \STATE {\bfseries Output:} Top-$N$ frequent subgraphs $\mathcal{F}_{k_{max}}$.
    \STATE $\mathcal{C}_3 \gets$ Enumerate all 3-subgraphs in $\mathcal{G}$
    \STATE $\mathcal{F}_3 \gets$ select top-$N$ from $\mathcal{C}_3$ 
    \FOR{$k = 3 \to k_{max}-1$}
        \STATE $\mathcal{C}_{k+1} \gets$ ExpandOneNode($\mathcal{G}$, $\mathcal{F}_k$)
        \STATE Compute scores $s_{g} = p^{\theta}(g)$ for all $g \in \mathcal{C}_{k+1}$
        \STATE $\mathcal{F}_{k+1} \gets$ select top-$N$ from $\mathcal{C}_{k+1}$
    \ENDFOR
    \STATE {\bfseries return} $\mathcal{F}_{k_{max}}$
    \end{algorithmic}
\end{algorithm}

\subsection{Beam Search}
\label{subsec:mining_framework}
Leveraging the graph diffusion estimator, GraDE adopts a beam search framework to discover frequent subgraphs.
The process follows a node-incremental expansion strategy.
Initially, all connected subgraphs of size $k=3$ are enumerated, with their exact frequencies and instance locations recorded.
In each iteration, the top-$N$ frequent $k$-subgraphs are selected as seeds.
These seeds, with their locations, are used to generate candidate $(k+1)$-subgraphs by adding an adjacent node, which are then scored by the estimator.
Only candidates with high predicted probabilities are retained for the next iteration.
The complete procedure and further details are provided in Alg. \ref{alg:mining} and Appendix \ref{appendix:beam_search}.

%% file: sections/Experiments.tex
\section{Experiments}
\label{sec:Experiments}
In this section, comprehensive experiments are conducted to evaluate GraDE.
Section \ref{subsec:es} details the experimental setup.
Section \ref{subsec:gdme} compares the GraDE estimator with baselines.
Section \ref{subsec:as} provides an ablation study to investigate how key hyperparameters influence the performance of the GraDE estimator.
Section \ref{subsec:fsm} evaluates the overall effectiveness of the GraDE framework in finding frequent subgraph patterns.

\subsection{Experimental Setup}
\begin{table}[hbp]
    \centering
    \caption{Statistical information of different datasets (NB is the abbreviation for NAS-Bench). Statistics include the maximum number of nodes in a single graph, the total number of nodes across all graphs, and the number of operator types.}
    \label{tab:datasets}
    \small
    \begin{tabular}{cccc}
    \toprule
     & \textbf{Max\#Nodes} & \textbf{Sum\#Nodes} & \textbf{\#Operators} \\
    \midrule
    NB-101 & 7 & 2898k & 5  \\
    NB-201 & 8 & 125k & 7  \\
    NB-301 & 11 & 220k & 11 \\
    Younger & 45670 & 28780k & 314 \\
    \bottomrule
    \end{tabular}
\end{table}
\label{subsec:es}
\textbf{Dataset.} Datasets utilized in the experiments are divided into synthesized and real-world datasets.
The synthesized datasets include NAS-Bench-101~\cite{Ying2019NASBench101TR}, NAS-Bench-201~\cite{Dong020} and NAS-Bench-301~\cite{9762315} datasets.
For the real-world dataset, this paper employs Younger, which is the first real-world neural architecture dataset~\cite{Yang2024YoungerTF}.
Their information is shown in Table \ref{tab:datasets}.
See \ref{subsec:dataset_description} for more details.

\textbf{Key Hyperarameters.}
The estimator performance depends on the training set, which is determined by three parameters: \textit{subgraph size} $k$, \textit{sampling method}, and \textit{sampling density} \footnote{The ratio of non-isomorphic $k$-subgraph types in the training set to that in the dataset. For a dataset with 5 types, 0.4 means the training set contains instances from 2 types (e.g., $\{A, A, A, B\}$).}.
Another parameter is \textit{estimation rounds}, which is the number of Monte Carlo trials for estimating $p_0(G_0)$.

\textbf{Evaluation Protocol.}
The evaluation consists of two parts. 
First, the estimator performance is evaluated by Spearman's rank correlation coefficient (Spearman's $\rho$) and Kendall's rank correlation coefficient (Kendall's $\tau$) between predicted scores and ground-truth frequencies obtained from exhaustive enumeration of $k$-subgraphs.
Second, the overall framework is evaluated for larger $k$ where enumeration is infeasible.
This part compares the mean and median frequencies of the discovered top-$N$ subgraphs, verified by VF2~\cite{VF2} algorithm with a cutoff time.
Details are in \ref{subsec:evaluation_protocol}.

\subsection{The Performance of Graph Diffusion Estimator}
\label{subsec:gdme}

\begin{table*}[htbp]
    \centering
    \caption{Performance of different methods on synthesized and real-world datasets. Results are reported as mean$\pm$std for $k\in\{4, 5\}$ at a sampling density of 0.1. Evaluation metrics are Spearman's $\rho$ and Kendall's $\tau$. \textbf{Bold} and \underline{underlined} values denote the best and second-best performance, respectively. DisCo-E, DeFoG-E, and DiGress-E denote estimators built upon corresponding graph diffusion models, using Rand-ESU for training set construction. Estimation rounds are fixed to 20 for DisCo-E, DiGress-E, and GraphVAE, and 60 for DeFoG-E. The ARS sampling method timed out on the Younger dataset, indicated by ``X''. Due to the prohibitive cost of exact enumeration to obtain ground truth for size $k=5$, results on Younger are evaluated on a random 10\% subset of graphs.}
    \label{tab:estimator}
    \small
    \begin{tabular}{ll cc cc cc cc}
        \toprule
        \multirow{2}{*}{Model} & \multirow{2}{*}{Class} & \multicolumn{2}{c}{NAS-Bench-101} & \multicolumn{2}{c}{NAS-Bench-201} & \multicolumn{2}{c}{NAS-Bench-301} & \multicolumn{2}{c}{Younger}\\
        \cmidrule(lr){3-4} \cmidrule(lr){5-6} \cmidrule(lr){7-8} \cmidrule(lr){9-10}
        & & $\rho$ $\uparrow$ & $\tau$ $\uparrow$ & $\rho$ $\uparrow$ & $\tau$ $\uparrow$ & $\rho$ $\uparrow$ & $\tau$ $\uparrow$ & $\rho$ $\uparrow$ & $\tau$ $\uparrow$ \\
        \midrule
        \multicolumn{10}{c}{Subgraph size $k=4$} \\
        \midrule
        ARS & Sampling & 0.33\tiny{$\pm$ 0.008} & 0.27\tiny{$\pm$ 0.007} & 0.41\tiny{$\pm$ 0.005} & 0.38\tiny{$\pm$ 0.004} & 0.42\tiny{$\pm$ 0.005} & 0.35\tiny{$\pm$ 0.004} & X & X \\
        NRS & Sampling & 0.34\tiny{$\pm$ 0.005} & 0.28\tiny{$\pm$ 0.004} & 0.39\tiny{$\pm$ 0.009} & 0.36\tiny{$\pm$ 0.009} & 0.41\tiny{$\pm$ 0.002} & 0.34\tiny{$\pm$ 0.002} & 0.45\tiny{$\pm$ 0.000} & \underline{0.38}\tiny{$\pm$ 0.000} \\
        Rand-ESU & Sampling & 0.36\tiny{$\pm$ 0.022} & 0.29\tiny{$\pm$ 0.018} & 0.39\tiny{$\pm$ 0.012} & 0.36\tiny{$\pm$ 0.011} & 0.42\tiny{$\pm$ 0.007} & 0.35\tiny{$\pm$ 0.006} & \underline{0.46}\tiny{$\pm$ 0.001} & 0.38\tiny{$\pm$ 0.001}  \\
        Rand-FaSE & Sampling & 0.34\tiny{$\pm$ 0.011} & 0.28\tiny{$\pm$ 0.009} & 0.40\tiny{$\pm$ 0.014} & 0.37\tiny{$\pm$ 0.013} & 0.43\tiny{$\pm$ 0.004} & 0.36\tiny{$\pm$ 0.004} & 0.35\tiny{$\pm$ 0.001} & 0.29\tiny{$\pm$ 0.001} \\
        GraphVAE & VAE & 0.40\tiny{$\pm$ 0.021} & 0.27\tiny{$\pm$ 0.017} & 0.34\tiny{$\pm$ 0.058} & 0.26\tiny{$\pm$ 0.048} & 0.22\tiny{$\pm$ 0.011} & 0.14\tiny{$\pm$ 0.008} & 0.38\tiny{$\pm$ 0.002} & 0.26\tiny{$\pm$ 0.002} \\
        \rowcolor[gray]{0.9}
        DiGress-E & Diffusion & 0.60\tiny{$\pm$ 0.077} & 0.43\tiny{$\pm$ 0.063} & 0.72\tiny{$\pm$ 0.065} & 0.57\tiny{$\pm$ 0.042} & 0.51\tiny{$\pm$ 0.117} & 0.35\tiny{$\pm$ 0.101} & 0.38\tiny{$\pm$ 0.008} & 0.27\tiny{$\pm$ 0.006} \\
        \rowcolor[gray]{0.9}
        DeFoG-E & Flow & \underline{0.84}\tiny{$\pm$ 0.010} & \underline{0.65}\tiny{$\pm$ 0.006} & \underline{0.75}\tiny{$\pm$ 0.060} & \underline{0.60}\tiny{$\pm$ 0.038} & \underline{0.70}\tiny{$\pm$ 0.110} & \textbf{0.52}\tiny{$\pm$ 0.090} & 0.41\tiny{$\pm$ 0.013} & 0.29\tiny{$\pm$ 0.009} \\
        \rowcolor[gray]{0.9}
        DisCo-E & Diffusion & \textbf{0.88}\tiny{$\pm$ 0.023} & \textbf{0.71}\tiny{$\pm$ 0.032} & \textbf{0.79}\tiny{$\pm$ 0.030} & \textbf{0.66}\tiny{$\pm$ 0.029} & \textbf{0.70}\tiny{$\pm$ 0.032} & \underline{0.51}\tiny{$\pm$ 0.029} & \textbf{0.55}\tiny{$\pm$ 0.007} & \textbf{0.40}\tiny{$\pm$ 0.006} \\
        \midrule
        \multicolumn{10}{c}{Subgraph size $k=5$} \\
        \midrule
        ARS & Sampling & 0.43\tiny{$\pm$ 0.001} & 0.37\tiny{$\pm$ 0.001} & 0.31\tiny{$\pm$ 0.004} & 0.28\tiny{$\pm$ 0.003} & 0.42\tiny{$\pm$ 0.002} & 0.36\tiny{$\pm$ 0.002} & X & X \\
        NRS & Sampling & 0.42\tiny{$\pm$ 0.001} & 0.36\tiny{$\pm$ 0.001} & 0.32\tiny{$\pm$ 0.006} & 0.29\tiny{$\pm$ 0.006} & 0.41\tiny{$\pm$ 0.002} & 0.35\tiny{$\pm$ 0.002} & 0.41\tiny{$\pm$ 0.000} & \underline{0.35}\tiny{$\pm$ 0.000} \\
        Rand-ESU & Sampling & 0.44\tiny{$\pm$ 0.002} & 0.38\tiny{$\pm$ 0.002} & 0.33\tiny{$\pm$ 0.011} & 0.30\tiny{$\pm$ 0.010} & 0.42\tiny{$\pm$ 0.001} & 0.36\tiny{$\pm$ 0.001} & 0.42\tiny{$\pm$ 0.001} & \underline{0.35}\tiny{$\pm$ 0.000} \\
        Rand-FaSE & Sampling & 0.42\tiny{$\pm$ 0.002} & 0.36\tiny{$\pm$ 0.002} & 0.30\tiny{$\pm$ 0.007} & 0.27\tiny{$\pm$ 0.007} & 0.43\tiny{$\pm$ 0.001} & 0.37\tiny{$\pm$ 0.001} & 0.33\tiny{$\pm$ 0.007} & 0.28\tiny{$\pm$ 0.006} \\
        GraphVAE & VAE & 0.42\tiny{$\pm$ 0.011} & 0.30\tiny{$\pm$ 0.009} & 0.70\tiny{$\pm$ 0.009} & 0.56\tiny{$\pm$ 0.008} & 0.46\tiny{$\pm$ 0.004} & 0.35\tiny{$\pm$ 0.004} & 0.32\tiny{$\pm$ 0.003} & 0.23\tiny{$\pm$ 0.002} \\
        \rowcolor[gray]{0.9}
        DiGress-E & Diffusion & 0.88\tiny{$\pm$ 0.003} & 0.73\tiny{$\pm$ 0.004} & \underline{0.85}\tiny{$\pm$ 0.001} & \underline{0.70}\tiny{$\pm$ 0.001} & \textbf{0.81}\tiny{$\pm$ 0.007} & \textbf{0.64}\tiny{$\pm$ 0.009} & 0.40\tiny{$\pm$ 0.008} & 0.28\tiny{$\pm$ 0.006} \\
        \rowcolor[gray]{0.9}
        DeFoG-E & Flow & \underline{0.91}\tiny{$\pm$ 0.007} & \underline{0.75}\tiny{$\pm$ 0.009} & 0.61\tiny{$\pm$ 0.004} & 0.47\tiny{$\pm$ 0.003} & 0.74\tiny{$\pm$ 0.018} & 0.56\tiny{$\pm$ 0.016} & \underline{0.47}\tiny{$\pm$ 0.002} & 0.33\tiny{$\pm$ 0.002} \\
        \rowcolor[gray]{0.9}
        DisCo-E & Diffusion & \textbf{0.94}\tiny{$\pm$ 0.002} & \textbf{0.81}\tiny{$\pm$ 0.002} & \textbf{0.87}\tiny{$\pm$ 0.016} & \textbf{0.73}\tiny{$\pm$ 0.020} & \underline{0.79}\tiny{$\pm$ 0.006} & \underline{0.60}\tiny{$\pm$ 0.008} & \textbf{0.56}\tiny{$\pm$ 0.002} & \textbf{0.40}\tiny{$\pm$ 0.001} \\
        \bottomrule
    \end{tabular}
\end{table*}

To the best of our knowledge, this work is the first to apply graph diffusion models for subgraph frequency estimation.
Besides DisCo-E in Section \ref{subsec:GDM_estimator}, this section further adapts and implements several representative graph generative models, including DiGress~\cite{vignac2023digress}, DeFoG~\cite{qin2025defog}, and GraphVAE~\cite{Simonovsky2018GraphVAETG}, as subgraph frequency estimators for comparison.
These models are evaluated alongside sampling-based methods, including ARS, NRS~\cite{lu_sampling_2012}, Rand-ESU~\cite{wernicke_efficient_2006}, and Rand-FaSE~\cite{paredes_rand-fase_2015}.
For a fair comparison, the sampling density for all methods is fixed to 0.1.
For generative methods, Rand-ESU serves as the sampling method for constructing the training set.
The performance of all methods is evaluated across subgraph sizes $k\in\{4,5\}$, as obtaining ground truth for larger $k$ on Younger is computationally prohibitive.

As shown in Table \ref{tab:estimator}, DisCo-E achieves the best or second-best performance across all datasets.
Specifically, DisCo-E significantly improves Spearman's $\rho$ by up to 114\% on synthesized NAS-Bench-101 and 33\% on real-world Younger (both vs. Rand-ESU, $k=5$).
Furthermore, diffusion-based and flow-based methods, including DiGress-E and DeFoG, significantly outperform sampling and VAE-based methods on synthesized datasets, reflecting their superior capability in modeling underlying frequency distributions.
These results demonstrate the overall superiority of leveraging graph diffusion models as subgraph frequency estimators, with DisCo-E consistently achieving the overall best results.

\subsection{Ablation Study}
\label{subsec:as}
This section investigates the impact of key hyperparameters on the estimators' performance, including the estimation rounds, the sampling density, and the sampling method.

\begin{figure}[h!]
    \centering
    \includegraphics[width=0.48\textwidth]{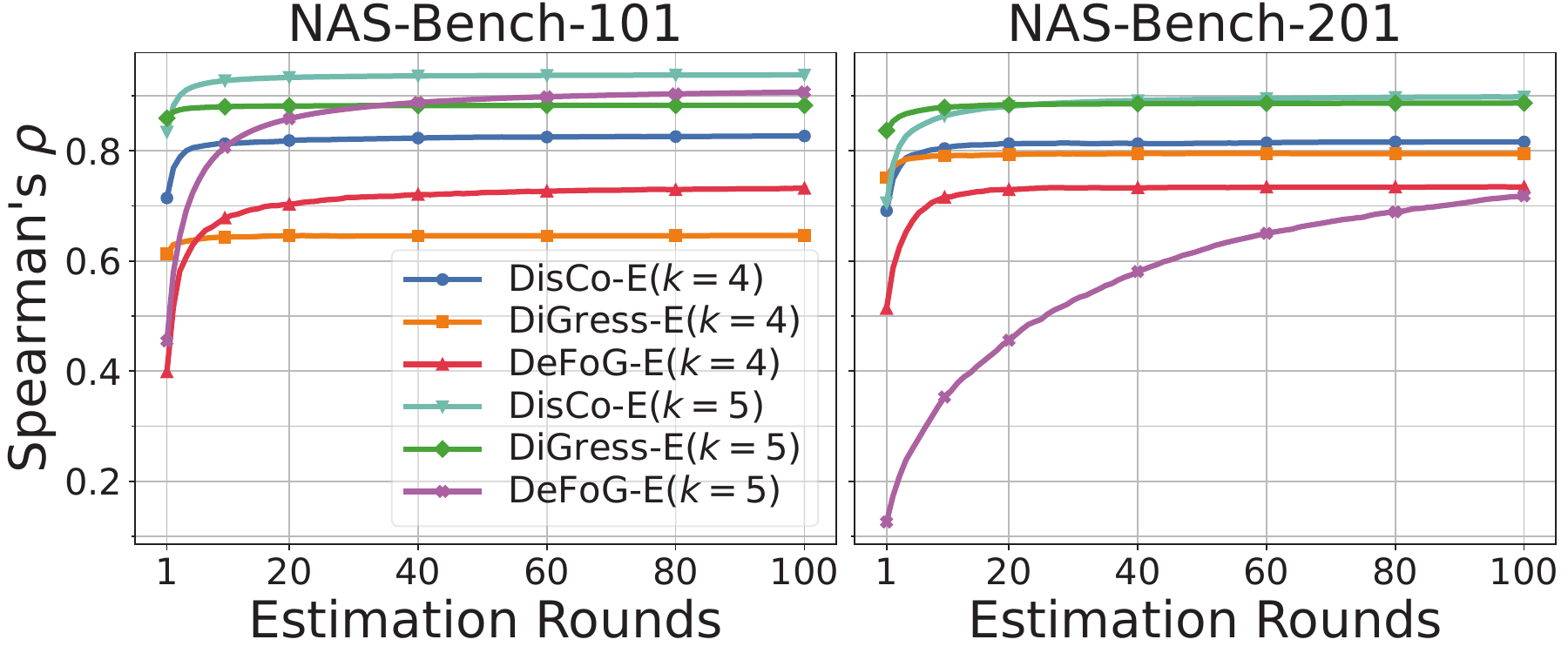}
    \caption{Spearman's $\rho$ across different estimation rounds. All estimators (DisCo-E, DeFoG-E, and DiGress-E) are paired with the Rand-ESU sampling method at a fixed sampling density of 0.1. The parameter $k$ in the legend denotes the subgraph size.}
    \label{fig:rounds_vs_metric}
\end{figure}

\textbf{Impact of Estimation Rounds.}
The estimation rounds influence both performance and computational cost.
Figure \ref{fig:rounds_vs_metric} presents Spearman's $\rho$ across different methods and subgraph sizes ($k \in \{4, 5\}$) as estimation rounds increase.
For all estimators, Spearman's $\rho$ initially rises and then stabilizes.
DisCo-E and DiGress-E converge at approximately 20 rounds, while DeFoG-E stabilizes around 60 rounds.
These elbow points represent an optimal trade-off between performance and efficiency, indicating that GraDE's Monte Carlo process converges rapidly with limited iterations.

\begin{figure}[htbp]
    \centering
    \includegraphics[width=0.48\textwidth]{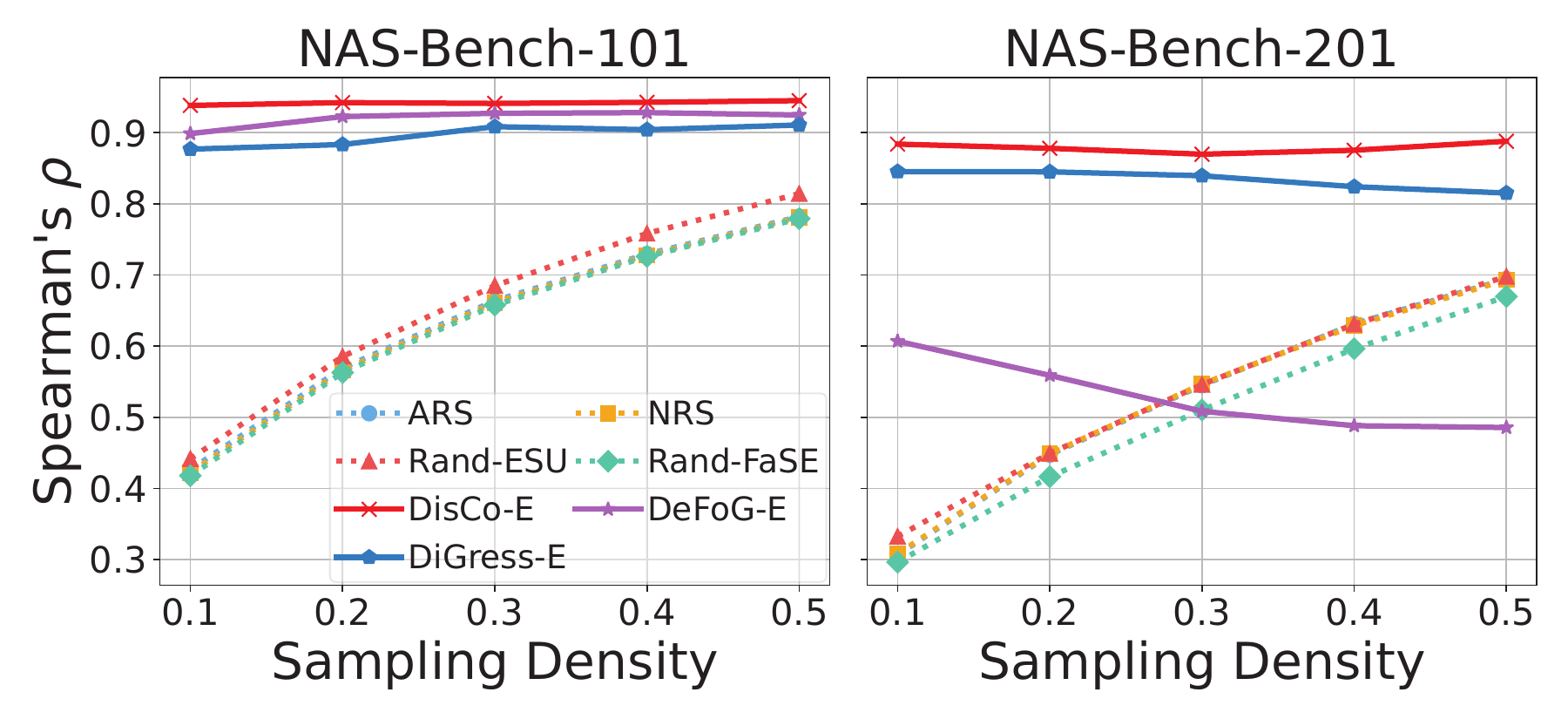}
    \caption{Spearman's $\rho$ across different sampling densities. All estimators (DisCo-E, DeFoG-E, and DiGress-E) are paired with Rand-ESU. The subgraph size is fixed to 5. The estimation rounds are fixed to 20 for DisCo-E and DiGress-E, and 60 for DeFoG-E.}
    \label{fig:sd}
\end{figure}

\begin{figure}[htbp]
    \centering
    \includegraphics[width=0.47\textwidth]{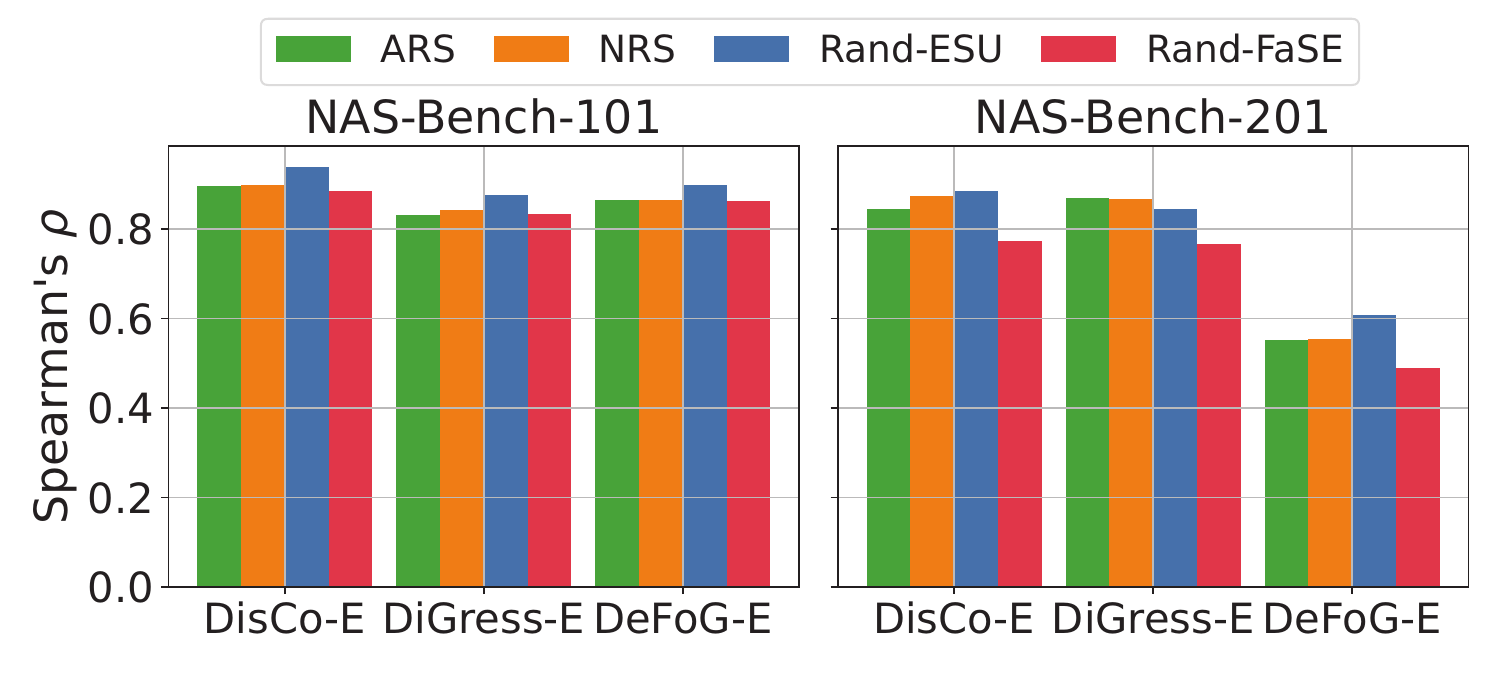}
    \caption{Spearman's $\rho$ across different sampling methods. DisCo-E, DiGress-E, and DeFoG-E are estimators. The subgraph size is fixed to 5. The sampling density is fixed to 0.1. Estimation rounds are fixed to 20 for DisCo-E and DiGress-E, and 60 for DeFoG-E.}
    \label{fig:ssm}
\end{figure}

\textbf{Impact of Sampling Density.}
The sampling density is directly related to the total number of sampled subgraphs.
Since the sampling cost grows exponentially with the sampling density (see Appendix \ref{subsec:impact_of_sd}), a high sampling density leads to prohibitive computational costs.
Figure \ref{fig:sd} shows the performance of various methods as the sampling density increases from 0.1 to 0.5.
For sampling-based methods, their performance improves with higher sampling densities.
In contrast, the performance of graph diffusion estimators is stable, approaching their optimum at a sampling density of only 0.1.
Notably, DisCo-E achieves higher performance at 0.1 density than all sampling-based methods at 0.5 density.
These results demonstrate that GraDE can achieve superior performance even under sparse sampling conditions, which is crucial for mitigating the exponentially increasing costs of subgraph sampling.

\textbf{Impact of Sampling Method.}
Figure \ref{fig:ssm} compares the performance of various graph diffusion estimators when paired with different sampling methods.
Results on NAS-Bench-101 and NAS-Bench-201 show that the performance for all estimators remains stable across different sampling methods, verifying the robustness of GraDE.
Nevertheless, Rand-ESU consistently achieves the best or near-best results in most cases, which is likely due to its unbiased sampling nature.

\subsection{Discovery Capability of the GraDE Framework}
\label{subsec:fsm}

To evaluate GraDE's capability to find frequent subgraph patterns, experiments are conducted on the real-world Younger dataset for $k \in \{4, \dots, 15\}$.
The GraDE framework, specifically using DisCo-E trained on samples from NRS, Rand-ESU, or Rand-FaSE, is compared against these sampling-based methods.
For each $k$, the sampling baselines draw 40,000 instances, while GraDE utilizes the same samples for training before initiating the search.
Following the beam search strategy in Section~\ref{subsec:mining_framework}, GraDE starts from 3-subgraphs and retains the top-1000 candidates per iteration via DisCo-E scores.
Finally, the true frequencies of the top-50 discovered subgraphs are verified via VF2 algorithm.

\begin{figure}[tbp]
    \centering
    \includegraphics[width=0.45\textwidth]{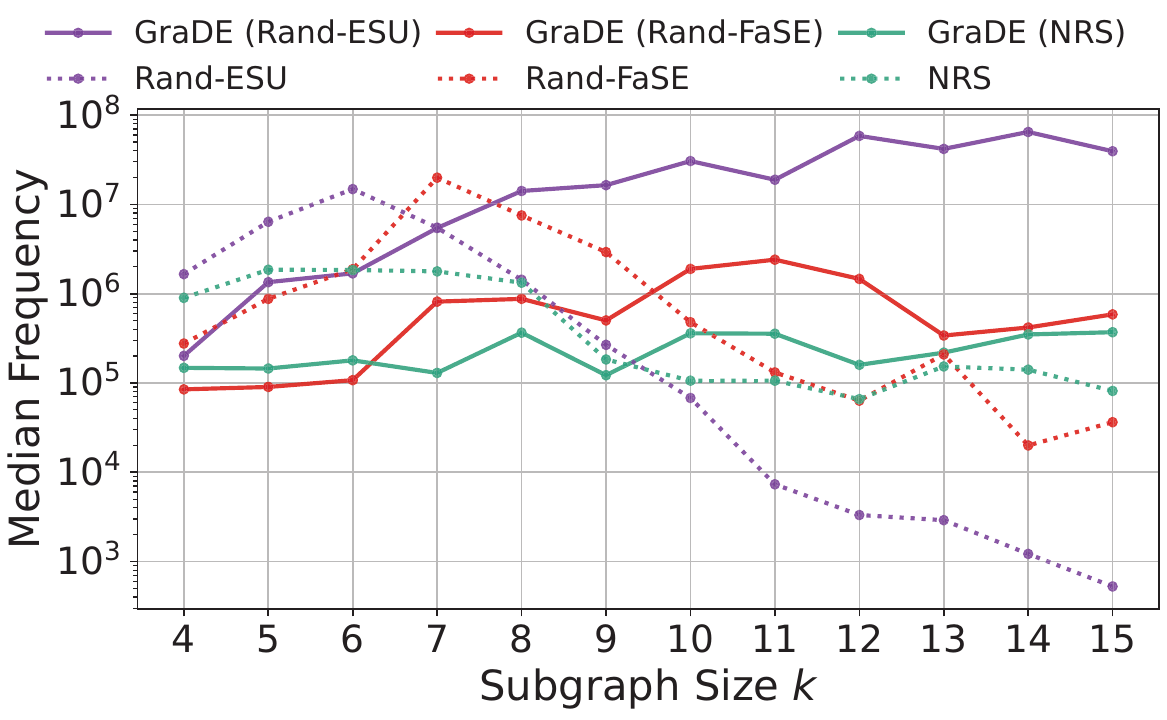}
    \caption{Median frequencies of top-50 discovered subgraphs. GraDE(Rand-ESU), GraDE(Rand-FaSE), and GraDE(NRS) denote the GraDE framework with corresponding sampling methods.}
    \label{fig:FSM_on_Younger}
\end{figure}

Figure \ref{fig:FSM_on_Younger} presents the median frequencies of the top-50 discovered subgraphs as quantitative measures of discovery quality.
As shown, a performance crossover occurs as $k$ increases: while sampling-based methods are better for smaller $k$, GraDE gains a substantial advantage for $k \ge 10$.
In this range, GraDE (Rand-ESU) outperforms Rand-ESU by over $100\times$, while GraDE (Rand-FaSE) and GraDE (NRS) achieve $10\times$ and $2.5\times$ average improvements over their respective sampling baselines.
Notably, GraDE (Rand-ESU) emerges as the best overall combination starting from $k=8$. This advantage over the best sampling-based method exceeds $30\times$ in large-scale scenarios ($k \ge 10$), demonstrating GraDE's superior capability in discovering large-scale frequent subgraph patterns.

\begin{figure}[htbp]
    \centering
    \includegraphics[width=0.45\textwidth]{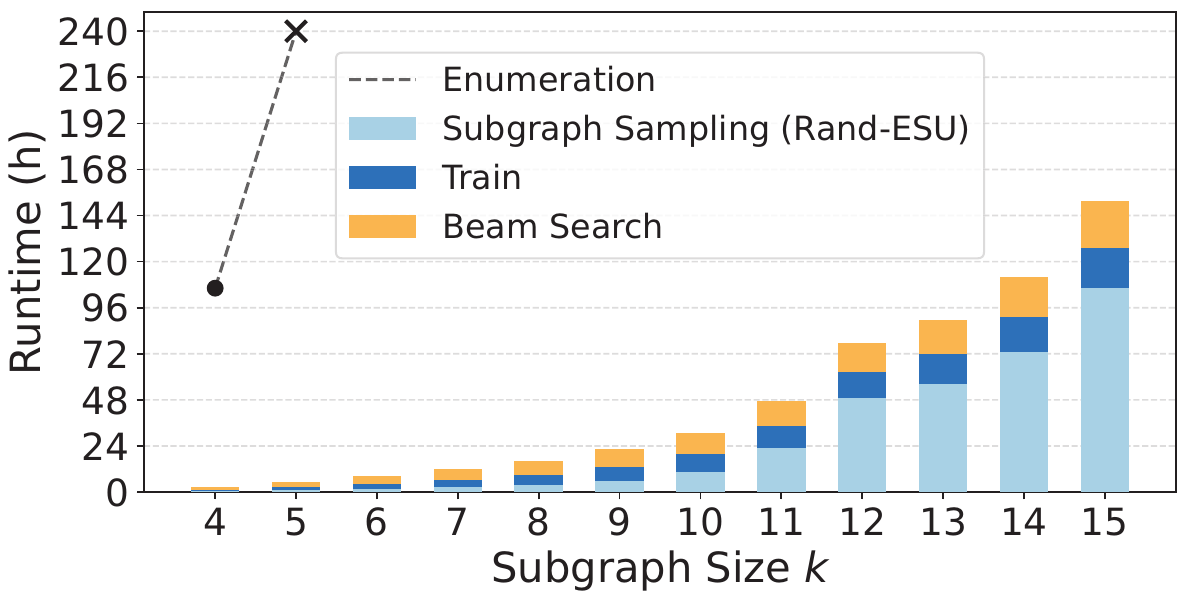}
    \caption{Runtime combination of GraDE (Rand-ESU).}
    \label{fig:GraDE_RESU_runtime}
\end{figure}

Figure \ref{fig:GraDE_RESU_runtime} decomposes the cumulative runtime of GraDE (Rand-ESU) from $k=4$ into sampling, training, and beam search.
As $k$ increases, sampling occupies an expanding proportion of the total time, whereas the additional overhead from model training and beam search becomes relatively marginal.
Notably, while the sampling time grows rapidly with $k$, the training and search costs exhibit modest growth.
This efficiency contrasts with exact enumeration, which becomes computationally prohibitive as $k$ increases. 
These results demonstrate the computational feasibility of GraDE.

%% file: sections/Conclusion.tex
\section{Conclusion}
\label{sec:Conclusion}
This paper proposes GraDE, a novel framework for frequent subgraph discovery in neural architectures.
It is the first to apply a graph diffusion model to score subgraph typicality as a frequency estimator.
This framework overcomes the exponential complexity of enumeration-based methods and the limited scalability of sampling-based methods.
Comprehensive experiments validate the effectiveness of the GraDE estimator and the overall GraDE framework.
Future work will extend this method to more diverse graph domains.

%% file: sections/Appendix.tex
\section*{Appendix Overview}
Appendix provides additional details organized as follows:
\begin{enumerate}
    \item Appendix \ref{sec:Monte_Carlo}: Monte Carlo Estimation.
    \item Appendix \ref{appendix:beam_search}: Details of the Beam Search.
    \item Appendix \ref{sec:appendix_disco}: Proofs of DisCo Estimator (DisCo-E).
    \item Appendix \ref{DiGress_E}: DiGress Estimator (DiGress-E).
    \item Appendix \ref{DeFoG_E}: DeFoG Estimator (DeFoG-E).
    \item Appendix \ref{sec:im_ssm}: Subgraph Sampling Methods.
    \item Appendix \ref{sec:appendix_further_details}: Further Implementation Details.
    \item Appendix \ref{sec:appendix_additional_exp}: Additional Experimental Results.
    \item Appendix \ref{sec:appendix_similar}: Similar but Distinct Work.
\end{enumerate}

\section{Monte Carlo Estimation}
\label{sec:Monte_Carlo}

By applying Jensen's inequality~\cite{jensen1906fonctions}, $\log p_0(G_0)$ is bounded as follows:
\begin{align}
  \label{eq:logpG0}
  \log p_{0}(G_0) & \ge \mathbb{E}_{\substack{\mathbf{G}_{\Delta t}, \dots, \mathbf{G}_T \\ \sim p_{\Delta t, \dots, T|0}}} \left[ \log \Big(p_T(\mathbf{G}_T) \prod_{t \in \mathcal{T}} \frac{ p_{t-\Delta t|t}(\mathbf{G}_{t-\Delta t}|\mathbf{G}_{t})}{p_{t|t-\Delta t}(\mathbf{G}_{t}|\mathbf{G}_{t-\Delta t})}\Big) \right] \\
    & =\mathbb{E}_{\substack{\mathbf{G}_{\Delta t}, \dots, \mathbf{G}_T \\ \sim p_{\Delta t, \dots, T|0}}} \left[ \log p_T(\mathbf{G}_T) + \sum_{t \in \mathcal{T}} \Big(\log p_{t-\Delta t|t}(\mathbf{G}_{t-\Delta t}|\mathbf{G}_{t}) - \log p_{t|t-\Delta t}(\mathbf{G}_{t}|\mathbf{G}_{t-\Delta t})\Big) \right],
\end{align}
where $\mathcal{T}=\{\Delta t, \dots, T\}$.
This work utilizes this lower bound to estimate $\log p_0(G_0)$ due to its superior numerical stability. 
Theoretically, the gap in this inequality vanishes when the reverse process perfectly recovers the forward trajectory—a core objective in diffusion model training.
Thus, the lower bound serves as a principled proxy for the exact probability.
Based on this derivation, the estimation of $p_0(G_0)$ is performed by averaging log-probabilities across trials and exponentiating the final mean, as detailed in Alg. \ref{alg:Monte_Carlo_Estimation}.

\begin{algorithm}[htbp]
\caption{Monte Carlo Estimation}
\label{alg:Monte_Carlo_Estimation}
    \begin{algorithmic}
    \STATE {\bfseries Input:} Target subgraph $G_0$, forward and reverse rate matrices $\{\mathbf{R}_t, \hat{\mathbf{R}}_t\}$, trials $M$, time step $\Delta t$, max time $T$, prior noising distribution $p_T$
    \STATE {\bfseries Output:} Estimated probability $\hat{p}_0(G_0)$
    
    \FOR{$i = 1 \to M$}
        \STATE $t \gets \Delta t$
        \STATE $L_i \gets 0$ \COMMENT{Initialize log-probability for current path}
        \WHILE{$t \leq T$}
            \STATE Sample $G_t \sim p_{t|t-\Delta t}(\cdot | G_{t-\Delta t})$ \COMMENT{Forward transition via Eq. \eqref{eq:FP}}
            \STATE $f \gets p_{t|t-\Delta t}(G_t | G_{t-\Delta t})$ \COMMENT{Forward probability via Eq. \eqref{eq:FP}}
            \STATE $r \gets p_{t-\Delta t|t}(G_{t-\Delta t} | G_t)$ \COMMENT{Reverse probability via Eq. \eqref{eq:RP}}
            \STATE $L_i \gets L_i + (\log r - \log f)$ \COMMENT{Accumulate log-ratio}
            \STATE $t \gets t + \Delta t$
        \ENDWHILE
        \STATE $L_i \gets L_i + \log p_T(G_T)$ \COMMENT{Path joint probability}
    \ENDFOR
    \STATE \textbf{return} $\hat{p}_0(G_0) = \exp (\frac{1}{M} \sum_{i=1}^M L_i)$
\end{algorithmic}
\end{algorithm}

\section{Details of the Beam Search}
\label{appendix:beam_search}
This section describes the node-incremental expansion strategy in GraDE, providing additional technical details to supplement Section \ref{subsec:mining_framework}.
Alg. \ref{alg:Expand_One_Node} provides the complete procedure.

The expansion process begins by using the pre-stored occurrences $\mathcal{I}_g$ associated with each frequent pattern $g$ in $\mathcal{F}_k$.
The algorithm performs a localized expansion by identifying nodes $v$ that are adjacent to each specific instance $I_g$.
By adding an adjacent node to each $I_g$, the algorithm forms a set of $(k+1)$-node candidate instances $\mathcal{R}_{k+1}$ that naturally reside within $\mathcal{G}$.

To manage redundancy, the algorithm deduplicates these candidates by grouping isomorphic instances into unique structural patterns.
Specifically, for each instance in $\mathcal{R}_{k+1}$, the algorithm maps it to a unique structural fingerprint via graph hashing.
This hash-based approach allows the algorithm to construct a dictionary $\mathcal{C}_{k+1}$, where all isomorphic instances are aggregated under their corresponding hash key. 
This organization ensures that each unique subgraph pattern is evaluated by the graph diffusion estimator only once, while the associated occurrence lists are preserved to facilitate frequency verification and further expansion in the subsequent iteration.

\begin{algorithm}[htbp]
\caption{Expand One Node in Beam Search}
\label{alg:Expand_One_Node}
    \begin{algorithmic}
    \STATE {\bfseries Input:} Graph set $\mathcal{G}$, frequent $k$-subgraph patterns $\mathcal{F}_k$.
    \STATE {\bfseries Output:} Candidate $(k+1)$-subgraph patterns $\mathcal{C}_{k+1}$.
    
    \FUNCTION{ExpandOneNode($\mathcal{G}$, $\mathcal{F}_k$)}

    \STATE $\mathcal{R}_{k+1} \gets \emptyset$  // Set of raw candidate instances
    \FOR{$g \in \mathcal{F}_k$}
        \STATE $\mathcal{I}_g \gets $ all occurrences of $g$ in $\mathcal{G}$ // Pre-stored occurrences
        \FOR{$I_g \in \mathcal{I}_g$}
            \STATE $N(I_g) \gets \{v \in V(\mathcal{G}) \setminus I_g \mid \exists u \in I_g, (u, v) \in E(\mathcal{G}) \lor (v, u) \in E(\mathcal{G})\}$
            \FOR{$v \in N(I_g)$}
                \STATE $\mathcal{R}_{k+1} \gets \mathcal{R}_{k+1} \cup \{I_g \cup \{v\}\}$
            \ENDFOR
        \ENDFOR
    \ENDFOR
    \STATE $\mathcal{C}_{k+1} \gets \{\}$ // Dictionary mapping \{GraphHash: Instances\}
    \FOR{$I_g' \in \mathcal{R}_{k+1}$}
        \STATE $h \gets \text{GraphHash}(I_g')$
        \STATE $\mathcal{C}_{k+1}[h] \gets \mathcal{C}_{k+1}[h] \cup \{I_g'\}$
    \ENDFOR
    \STATE {\bfseries return} $\mathcal{C}_{k+1}$
    
    \ENDFUNCTION
\end{algorithmic}
\end{algorithm}

\section{Proofs of DisCo Estimator (DisCo-E)}
\label{sec:appendix_disco}
\subsection{Forward Probability}
\label{proof_fp}

As defined in Eq. \eqref{eq:KF}, the evolution of the forward transition probability $p_{t|s}(y|x)$ for $y, x \in \Omega$ is governed by the Kolmogorov forward equation:
\begin{equation}
    \label{eq:KF_app}
    \frac{d}{dt}p_{t|s}(y|x) = \sum_{z \in \Omega} p_{t|s}(z|x) \mathbf{R}_t(z, y).
\end{equation}
Let $\mathbf{P}_{t|s} \in \mathbb{R}^{|\Omega| \times |\Omega|}$ be the forward transition probability matrix where $(\mathbf{P}_{t|s})_{xy} = p_{t|s}(y|x)$.
Eq. \eqref{eq:KF_app} is then expressed in matrix form as:
\begin{equation}
    \label{eq:matrix_diff}
    \frac{d}{dt}\mathbf{P}_{t|s} = \mathbf{P}_{t|s} \mathbf{R}_t,
\end{equation}
subject to the initial condition $\mathbf{P}_{s|s} = \mathbf{I}$.
The proposed solution $\mathbf{P}_{t|s} = \exp \left( \int_{s}^{t} \mathbf{R}_u du \right)$ can be verified by substituting it into Eq. \eqref{eq:matrix_diff}.
Under the commutativity assumption $\mathbf{R}_t \mathbf{R}_s = \mathbf{R}_s \mathbf{R}_t$, the integral commutes with its derivative such that $\left( \int_{s}^{t} \mathbf{R}_u du \right) \mathbf{R}_t = \mathbf{R}_t \left( \int_{s}^{t} \mathbf{R}_u du \right)$.
Expanding the matrix exponential as a power series yields: 
\begin{equation} 
    \mathbf{P}_{t|s} = \sum_{i=0}^{+\infty} \frac{1}{i!}\left( \int_{s}^{t} \mathbf{R}_u du \right)^i .
\end{equation}
Substituting this expression into the left-hand side (LHS) of Eq. \eqref{eq:matrix_diff}, the derivative is computed as: 
\begin{align} 
    \text{LHS} 
    &= \frac{d}{dt} \sum_{i=0}^{+\infty} \frac{1}{i!} \left( \int_{s}^{t} \mathbf{R}_u du \right)^i \\
    &= \sum_{i=1}^{+\infty} \frac{1}{i!} \cdot i \left( \int_{s}^{t} \mathbf{R}_u du \right)^{i-1} \frac{d}{dt} \left( \int_{s}^{t} \mathbf{R}_u du \right) \\
    &= \sum_{i=1}^{+\infty} \frac{1}{(i-1)!} \left( \int_{s}^{t} \mathbf{R}_u du \right)^{i-1} \mathbf{R}_t \\
    &= \left( \sum_{j=0}^{+\infty} \frac{1}{j!} \left( \int_{s}^{t} \mathbf{R}_u du \right)^j \right) \mathbf{R}_t \\
    &= \mathbf{P}_{t|s} \mathbf{R}_t \\
    &= \text{RHS} \label{eq:RHS}
\end{align}
By taking the entry at row $x$ and column $y$ of the matrix $\mathbf{P}_{t|s}$, the forward transition probability is obtained as:
\begin{equation}
    \label{eq:fp2}
    p_{t|s}(y|x) = \left[ \exp \left( \int_{s}^{t} \mathbf{R}_u du \right) \right]_{xy}
\end{equation}
This result coincides with the formulation in Eq. \eqref{eq:FP}.

\subsection{Reverse Probability}
\label{proof_rp}

The reverse transition rate matrix $\hat{\mathbf{R}}_t$ is defined through the detailed balance condition with respect to the forward process:
\begin{equation}
    p_t(y) \hat{\mathbf{R}}_t(y, x) = p_t(x) \mathbf{R}_t(x, y).
\end{equation}
For any two states $x, y \in \Omega$, the entry at row $x$ and column $y$ of the reverse rate matrix is given by:
\begin{equation}
    \hat{\mathbf{R}}_t(y, x) = \mathbf{R}_t(x, y) \cfrac{p_t(x)}{p_t(y)}.
\end{equation}

The number of transitions occurring within a small time interval $\Delta t=t-s$ in a continuous-time Markov process follows a Poisson distribution.
Let $\mathbf{K}$ be the random variable representing the number of transitions from state $y$ during $\Delta t$, with the rate parameter $\lambda = -\Delta t \hat{\mathbf{R}}_t(y, y)$.
The probability of observing exactly $k$ transitions is given by:
\begin{equation}
    P(\mathbf{K}=k) = \cfrac{(\lambda)^k e^{-\lambda}}{k!}.
\end{equation}
For a sufficiently small step $\Delta t$, the probability of more than one transition occurring can be ignored, allowing for the approximation $P(\mathbf{K}=1) \approx 1 - P(\mathbf{K}=0)$.
The reverse transition probability $p_{s|t}(x|y)$ is then derived based on whether a state transition occurs.

If no transition occurs ($x=y$), the probability is simply the zero-event case of the Poisson distribution:
\begin{equation}
    p_{s|t}(x|y) = P(\mathbf{K}=0) = \exp \left( \Delta t \hat{\mathbf{R}}_t(y, y) \right).
\end{equation}

If a transition occurs ($x \neq y$), the probability is the product of the probability that a transition is initiated and the conditional probability of jumping to state $x$:
\begin{align}
    p_{s|t}(x|y) &= (1 - P(\mathbf{K}=0)) \cdot \cfrac{\hat{\mathbf{R}}_t(y, x)}{\sum_{z \neq y} \hat{\mathbf{R}}_t(y, z)} \\
    &= \left(1 - \exp \left( \Delta t \hat{\mathbf{R}}_t(y, y) \right)\right) \cfrac{\hat{\mathbf{R}}_t(y, x)}{-\hat{\mathbf{R}}_t(y, y)} \\
    &= \left(\exp \left( \Delta t \hat{\mathbf{R}}_t(y, y) \right) - 1\right) \cfrac{\hat{\mathbf{R}}_t(y, x)}{\hat{\mathbf{R}}_t(y, y)}.
\end{align}
Combining the two cases above, the reverse transition probability can be summarized in a unified form: 
\begin{equation} 
    p_{s|t}(x|y) = 
    \begin{cases} 
        \exp \left( \Delta t \hat{\mathbf{R}}_t(y, y) \right) & x = y \\
        \left(\exp \left( \Delta t \hat{\mathbf{R}}_t(y, y) \right) - 1\right) \cfrac{\hat{\mathbf{R}}_t(y, x)}{\hat{\mathbf{R}}_t(y, y)} & x \neq y 
    \end{cases} .
\end{equation} 
This result coincides with the formulation in Eq. \eqref{eq:RP}.

\subsection{Factorized Forward Probability}
\label{proof_ffp}
Under the factorization assumption, each node $i$ and edge $(i, j)$ evolves independently according to its respective forward transition rate matrix.
The forward rate matrices for a node and an edge are defined as $\mathbf{R}_t^{X} = \beta(t) \mathbf{R}_x$ and $\mathbf{R}_t^{E} = \beta(t) \mathbf{R}_e$, where $\Omega_x = \{1, 2, \cdots, a\}$ and $\Omega_e = \{1, 2, \cdots, b\}$ denote their state spaces.
The evolution of the transition probabilities $p_{t|s}(x_t^i|x_s^i)$ and $p_{t|s}(e_t^{ij}|e_s^{ij})$ is governed by the following component-wise Kolmogorov forward equations:

\begin{equation}
    \frac{d}{dt}p_{t|s}(x_t^i|x_s^i) = \sum_{x \in \Omega_x} p_{t|s}(x|x_s^i) \beta(t)\mathbf{R}_x(x, x_t^i),
\end{equation}
\begin{equation}
    \frac{d}{dt}p_{t|s}(e_t^{ij}|e_s^{ij}) = \sum_{e \in \Omega_e} p_{t|s}(e|e_s^{ij}) \beta(t)\mathbf{R}_e(e, e_t^{ij}).
\end{equation}
Since these equations share the same mathematical structure as Eq. \eqref{eq:KF_app} in Appendix \ref{proof_fp}, the analytical solutions can be obtained by following an identical derivation from Eq. \eqref{eq:KF_app} to Eq. \eqref{eq:RHS}.
Applying the matrix exponential form to each component yields:
\begin{equation}
    \label{eq:ffp_x}
    p_{t|s}(x_t^i|x_s^i)=\left(\exp (\int_{s}^t \beta(u)\mathbf{R}_x du)\right)_{x_s^ix_t^i},
\end{equation}
\begin{equation}
    \label{eq:ffp_e}
    p_{t|s}(e_t^{ij}|e_s^{ij})=\left(\exp (\int_{s}^t \beta(u)\mathbf{R}_e du)\right)_{e_s^{ij}e_t^{ij}}.
\end{equation}
Due to the independence of nodes and edges in the forward process, the joint transition probability $p_{t|s}(G_t|G_s)$ is the product of these individual transition probabilities.
This completes the proof of Theorem \ref{thm:factorized_forward}.

Based on Eq. \eqref{eq:ffp_x} and Eq. \eqref{eq:ffp_e}, the final expression of the forward transition probability is:
\begin{equation}
    p_{t|s}(G_t|G_s)=\prod_{i=1}^{k} \left(\exp (\int_{s}^t \beta(u)\mathbf{R}_x du)\right)_{x_s^ix_t^i} \prod_{i, j=1}^{k} \left(\exp (\int_{s}^t \beta(u)\mathbf{R}_e du)\right)_{e_s^{ij}e_t^{ij}}.
\end{equation}

\subsection{Factorized Reverse Probability}
\label{proof_frp}

Under the factorization assumption, the reverse transition rate matrix for each node $i$ at time $t$, denoted as $\hat{\mathbf{R}}_t^i$, is derived from the detailed balance condition:
\begin{equation}
    \hat{\mathbf{R}}_t^i(x_t^i, \tilde{x}_t^i)=\mathbf{R}_t^{X}(\tilde{x}_t^i, x_t^i)\cfrac{p_t(\tilde{x}_t^i)}{p_t(x_t^i)}.
\end{equation}
By marginalizing over the initial states $x_0^i \in \Omega_x=\{1, 2, \cdots, a\}$ and applying Bayes' rule, the ratio of marginal probabilities is expanded as follows:
\begin{align}
    \cfrac{p_t(\tilde{x}_t^i)}{p_t(x_t^i)}&=\cfrac{\sum_{x_0^i \in \Omega_x}p_{t, 0}(\tilde{x}_t^i, x_0^i)}{p_t(x_t^i)} \\
    &= \cfrac{\sum_{x_0^i \in \Omega_x}p_{t|0}(\tilde{x}_t^i|x_0^i)p_0(x_0^i)}{p_t(x_t^i)} \\
    &= \sum_{x_0^i \in \Omega_x}\cfrac{p_{t|0}(\tilde{x}_t^i|x_0^i)p_0(x_0^i)p_{0|t}(x_0^i|x_t^i)}{p_t(x_t^i)p_{0|t}(x_0^i|x_t^i)} \\
    &=\sum_{x_0^i \in \Omega_x}\cfrac{p_{t|0}(\tilde{x}_t^i|x_0^i)p_0(x_0^i)p_{0|t}(x_0^i|x_t^i)}{p_0(x_0^i)p_{t|0}(x_t^i|x_0^i)} \\
    &= \sum_{x_0^i \in \Omega_x} p_{0|t}(x_0^i|x_t^i) \cfrac{p_{t|0}(\tilde{x}_t^i|x_0^i)}{p_{t|0}(x_t^i|x_0^i)}.
\end{align}
In practice, since the state of a single node is conditioned on the entire graph structure, the term $p_{0|t}(x_0^i|x_t^i)$ is replaced by the posterior $p_{0|t}(x_0^i|G_t)$, which is approximated by a neural network:
\begin{equation}
    \cfrac{p_t(\tilde{x}_t^i)}{p_t(x_t^i)}=\sum_{x_0^i \in \Omega_x} p_{0|t}(x_0^i|G_t) \cfrac{p_{t|0}(\tilde{x}_t^i|x_0^i)}{p_{t|0}(x_t^i|x_0^i)}.
\end{equation}
Substituting the analytical solution of the forward transition probability from Theorem \ref{thm:factorized_forward} into the rate matrix expression yields:
\begin{align}
    \hat{\mathbf{R}}_t^i(x_t^i, \tilde{x}_t^i)&=\beta(t)\mathbf{R}_x(\tilde{x}_t^i, x_t^i)\sum_{x_0^i \in \Omega_x} p_{0|t}(x_0^i|G_t) \cfrac{\left[\exp (\int_{0}^t \beta(u)\mathbf{R}_x du)\right]_{x_0^i\tilde{x}_t^i}}{\left[\exp (\int_{0}^t \beta(u)\mathbf{R}_x du)\right]_{x_0^ix_t^i}}.
\end{align}
By replacing the dummy variable $\tilde{x}_t^i$ with the specific states $x_s^i$ and $x_t^i$, the entries of the reverse rate matrix for nodes are obtained:
\begin{equation}
    \hat{\mathbf{R}}_t^i(x_t^i, x_s^i)=\beta(t)\mathbf{R}_x(x_s^i, x_t^i)\sum_{x_0^i \in \Omega_x} p_{0|t}(x_0^i|G_t) \cfrac{\left[\exp (\int_{0}^t \beta(u)\mathbf{R}_x du)\right]_{x_0^ix_s^i}}{\left[\exp (\int_{0}^t \beta(u)\mathbf{R}_x du)\right]_{x_0^ix_t^i}}.
\end{equation}
The same derivation applies to the edge transition rate matrix $\hat{\mathbf{R}}_t^{ij}$ for any edge $(i, j)$, where $\Omega_e=\{1,2,\cdots,b\}$:
\begin{equation}
    \hat{\mathbf{R}}_t^{ij}(e_t^{ij}, e_s^{ij})=\beta(t)\mathbf{R}_e(e_s^{ij}, e_t^{ij})\sum_{e_0^{ij} \in \Omega_e} p_{0|t}(e_0^{ij}|G_t) \cfrac{\left[\exp (\int_{0}^t \beta(u)\mathbf{R}_e du)\right]_{e_0^{ij}e_s^{ij}}}{\left[\exp (\int_{0}^t \beta(u)\mathbf{R}_e du)\right]_{e_0^{ij}e_t^{ij}}}.
\end{equation}
Following the same logic as the derivation of reverse transition probability in Appendix \ref{proof_rp}, the factorized reverse probabilities are expressed in a unified form:
\begin{equation}
    p_{s|t}(x_s^i|G_t) = 
    \begin{cases} 
        \left(\exp\left(\Delta t \hat{\mathbf{R}}_t^i(x_t^i, x_t^i)\right) - 1\right) \cfrac{\hat{\mathbf{R}}_t^i(x_t^i, x_s^i)}{\hat{\mathbf{R}}_t^i(x_t^i, x_t^i)} & x_s^i\neq x_t^i \\
        \exp\left(\Delta t \hat{\mathbf{R}}_t^i(x_t^i, x_t^i)\right) & x_s^i= x_t^i
    \end{cases},
\end{equation}
\begin{equation}
    p_{s|t}(e_s^{ij}|G_t) = 
    \begin{cases} 
        \left(\exp\left(\Delta t \hat{\mathbf{R}}_t^{ij}(e_t^{ij}, e_t^{ij})\right) - 1\right) \cfrac{\hat{\mathbf{R}}_t^{ij}(e_t^{ij}, e_s^{ij})}{\hat{\mathbf{R}}_t^{ij}(e_t^{ij}, e_t^{ij})} & e_s^{ij}\neq e_t^{ij} \\
        \exp\left(\Delta t \hat{\mathbf{R}}_t^{ij}(e_t^{ij}, e_t^{ij})\right) & e_s^{ij}= e_t^{ij}
    \end{cases}.
\end{equation}
These results coincide with the formulations in Theorem \ref{thm:factorized_reverse}.

\section{DiGress Estimator (DiGress-E)}
\label{DiGress_E}

The estimator for $p_0(G_0)$ in discrete-time diffusion models can be derived as a specific case of Eq. \eqref{eq:pG0} with a constant step size $\Delta t = 1$.
This leads to the following formulation:
\begin{equation}
    p_{0}(G_0) = \mathbb{E}_{\mathbf{G}_1, \mathbf{G}_2, \cdots, \mathbf{G}_T \sim p_{1,2,\cdots,T|0}} \left[ p_T(\mathbf{G}_T) \prod_{t=1}^{T} \frac{ p_{t-1|t}(\mathbf{G}_{t-1}|\mathbf{G}_{t})}{p_{t|t-1}(\mathbf{G}_{t}|\mathbf{G}_{t-1})} \right] .
\end{equation}

Similar to DisCo, DiGress~\cite{vignac2023digress} adopts a factorized transition kernel for the forward process.
The transition probability from step $t-1$ to $t$ is decomposed into independent components for nodes and edges:
\begin{equation}
    p_{t|t-1}(G_t|G_{t-1})=\prod_{i=1}^{k} p_{t|t-1}(x_t^i|x_{t-1}^i) \prod_{i, j=1}^{k} p_{t|t-1}(e_t^{ij}|e_{t-1}^{ij}).
\end{equation}

In DiGress, the noise is injected via Markov transition matrices. The node-wise and edge-wise one-step transition probabilities are given by:
\begin{equation}
    p_{t|t-1}(x_t^i|x_{t-1}^i)=[\mathbf{Q}_t^X]_{x_{t-1}^ix_t^i},
\end{equation}
\begin{equation}
    p_{t|t-1}(e_t^{ij}|e_{t-1}^{ij})=[\mathbf{Q}_t^E]_{e_{t-1}^{ij}e_t^{ij}}.
\end{equation}
The transition matrices are structured to pull the distributions towards a prior distribution $\mathbf{m}$ (marginal distribution of the dataset):
\begin{equation}
    \mathbf{Q}_t^X=\alpha_t \mathbf{I} + (1-\alpha_t)\mathbf{1}_a(\mathbf{m}^X)^{\top} \quad \text{and} \quad \mathbf{Q}_t^E=\alpha_t \mathbf{I} + (1-\alpha_t)\mathbf{1}_b(\mathbf{m}^E)^{\top}.
\end{equation}
In these expressions, $\mathbf{1}_a \in \mathbb{R}^a$ and $\mathbf{1}_b \in \mathbb{R}^b$ are all-ones column vectors, while $\mathbf{m}^X \in \mathbb{R}^a$ and $\mathbf{m}^E \in \mathbb{R}^b$ are the marginal distributions (column vectors) of node types and edge types in the training set, respectively.

Due to the property of Markov Chain, the multi-step transition probability $p_{t|0}$ follows a similar form:
\begin{equation}
    p_{t|0}(G_t|G_0)=\prod_{i=1}^{k} p_{t|0}(x_t^i|x_0^i) \prod_{i, j=1}^{k} p_{t|0}(e_t^{ij}|e_0^{ij}),
\end{equation}
where the cumulative transition matrices $\bar{\mathbf{Q}}_t^X=\mathbf{Q}_1^X\mathbf{Q}_2^X\cdots\mathbf{Q}_t^X$ and $\bar{\mathbf{Q}}_t^E=\mathbf{Q}_1^E\mathbf{Q}_2^E\cdots\mathbf{Q}_t^E$ are as follows:
\begin{equation}
    \bar{\mathbf{Q}}_t^X=\bar{\alpha}_t \mathbf{I} + (1-\bar{\alpha}_t)\mathbf{1}_a(\mathbf{m}^X)^{\top} \quad \text{and} \quad \bar{\mathbf{Q}}_t^E=\bar{\alpha}_t \mathbf{I} + (1-\bar{\alpha}_t)\mathbf{1}_b(\mathbf{m}^E)^{\top}.
\end{equation}

The cumulative noise schedule follows a cosine schedule $\bar{\alpha}_t = \cos(0.5\pi(t/T+s)/(1+s))^2$.
As $t \to T$, the parameter $\bar{\alpha}_t$ approaches zero, which implies that the cumulative transition matrices $\bar{\mathbf{Q}}_t^X$ and $\bar{\mathbf{Q}}_t^E$ converge to $\mathbf{1}_a(\mathbf{m}^X)^{\top}$ and $\mathbf{1}_b(\mathbf{m}^E)^{\top}$, respectively.
Consequently, the transition probability $p_{t|0}$ becomes independent of the initial state $G_0$, leading to the stationary distribution:
\begin{equation}
    p_T(G_T)=\prod_{i=1}^k (\mathbf{m}^X)_{x_T^i} \prod_{i, j=1}^k (\mathbf{m}^E)_{e_T^{ij}}.
\end{equation}

To obtain the reverse transition $p_{t-1|t}(G_{t-1}|G_t)$, the following Bayes' expansion is utilized:
\begin{align}
    p_{t-1|t}(G_{t-1}|G_t)&=\sum_{G_0} p_{t-1,0|t}(G_{t-1}, G_0|G_t) \\
    &=\sum_{G_0} p_{t-1|0,t}(G_{t-1}|G_0, G_t) \cdot p_{0|t}(G_0|G_t) \\
    &=\sum_{G_0} \cfrac{p_{t-1,t|0}(G_{t-1},G_t|G_0)}{p_{t|0}(G_t|G_0)} \cdot p_{0|t}(G_0|G_t) \\
    &=\sum_{G_0} \cfrac{p_{t|t-1,0}(G_t|G_{t-1},G_0)p_{t-1|0}(G_{t-1}|G_0)}{p_{t|0}(G_t|G_0)} \cdot p_{0|t}(G_0|G_t) \\
    &=\sum_{G_0} \cfrac{p_{t|t-1}(G_t|G_{t-1})p_{t-1|0}(G_{t-1}|G_0)}{p_{t|0}(G_t|G_0)} \cdot p_{0|t}(G_0|G_t).
\end{align}
The unknown posterior $p_{0|t}(G_0|G_t)$ is parameterized by a Graph Transformer $\phi_{\theta}(G_t)$, which predicts the clean graph $\hat{G}_0 = \{\hat{x}_0^i, \hat{e}_0^{ij} \}_{i,j=1}^k$, where the components are given by the corresponding outputs of the neural network: $\hat{x}_0^i = \phi_{\theta}(G_t)^X_i \in \mathbb{R}^a$ and $\hat{e}_0^{ij} = \phi_{\theta}(G_t)^E_{ij} \in \mathbb{R}^b$.
Each component represents a categorical distribution (probability vector) over the possible node or edge types.
The model is optimized using the cross-entropy loss between the true $G_0$ and the predicted $\hat{G}_0$:
\begin{equation}
    \mathcal{L}=\mathbb{E}_{t \sim \mathcal{U}(1,T), \mathbf{G}_0 \sim p_0, \mathbf{G}_t \sim p_{t|0} } \left[ \sum_{i=1}^{k} \text{CE}(x_0^i, \hat{x}_0^i) + \lambda\sum_{i,j=1}^{k} \text{CE}(e_0^{ij}, \hat{e}_0^{ij}) \right].
\end{equation}
Once the model is trained, the estimated posterior probability for each node and each edge can be extracted from the corresponding entries of the predicted vectors:
\begin{equation}
    p_{0|t}^{\theta}(x_0^i|G_t)=[\phi_{\theta}(G_t)^X_i]_{x_0^i} \quad \text{and} \quad p_{0|t}^{\theta}(e_0^{ij}|G_t)=[\phi_{\theta}(G_t)^E_i]_{e_0^{ij}}.
\end{equation}
Consequently, the joint posterior $p_{0|t}^{\theta}(G_0|G_t)$ used in the reverse step is obtained through the product of individual probabilities:
\begin{equation}
    p_{0|t}(G_0|G_t)^{\theta}=\prod_{i=1}^{k} p_{0|t}^{\theta}(x_0^i|G_t) \prod_{i, j=1}^{k} p_{0|t}^{\theta}(e_0^{ij}|G_t).
\end{equation}
By substituting $p_{0|t}^{\theta}(G_0|G_t)$ back into the Bayes' expansion, the reverse transition $p_{t-1|t}(G_{t-1}|G_t)$ becomes computable, completing the DiGress estimator.

\section{DeFoG Estimator (DeFoG-E)}
\label{DeFoG_E}

Unlike standard diffusion models, DeFoG is built upon the discrete flow matching framework~\cite{qin2025defog,DFM}.
In this context, $p_0(G_0)$ represents the prior (noise) distribution, and $p_1(G_1)$ represents the data distribution.
Notably, the direction of time is defined such that $t=0$ denotes pure noise and $t=1$ denotes data.
Consequently, the forward (nosing) process corresponds to decreasing time $t$ ($1 \to 0$), while the reverse (denoising) process corresponds to increasing time $t$ ($0 \to 1$). 
The objective in this section is to estimate the data density $p_1$ using a reverse-time trajectory:
\begin{equation}
    \label{eq:defog_pG1}
      p_{1}(G_1) = \mathbb{E}_{\mathbf{G}_{1-\Delta t}, \mathbf{G}_{1-2\Delta t}, \dots, \mathbf{G}_0 \sim p_{1-\Delta t, 1-2\Delta t, \dots, 0|1}} \left[ p_0(\mathbf{G}_0) \prod_{t \in \mathcal{T}} \frac{ p_{t+\Delta t|t}(\mathbf{G}_{t+\Delta t}|\mathbf{G}_{t})}{p_{t|t+\Delta t}(\mathbf{G}_{t}|\mathbf{G}_{t+\Delta t})} \right], 
\end{equation}
where $\mathcal{T}=\{1-\Delta t, 1-2\Delta t, \cdots, \Delta t, 0\}$.
In this formulation, $p_{t|t+\Delta t}$ denotes the forward transition probability (noising) and $p_{t+\Delta t|t}$ denotes the reverse transition probability (denoising).

\subsection{Forward Trajectory and Transition Probabilities}

The noising trajectory in DeFoG is constructed by defining a probability path that linearly interpolates between the prior distribution $p_0$ and the data distribution $p_1$.
For any $t \in [0, 1]$, the marginal distribution $p_t$ is defined as:
\begin{equation}
    p_t(z_t)=tp_1(z_t)+(1-t)p_0(z_t).
\end{equation}
To achieve this marginal distribution, the transition probability from the data $z_1$ to an intermediate state $z_t$ is defined as:
\begin{equation}
    p_{t|1}(z_t|z_1)=t\delta(z_t, z_1)+(1-t)p_0(z_t).
\end{equation}
To implement the estimator in Eq. \eqref{eq:defog_pG1}, the forward transition probability $p_{t|s}$ for any $t<s$ must be determined.
A general form for the transition probability $p_{t|s,1}$ is assumed as follows:
\begin{equation}
    p_{t|s,1}(z_t|z_s, z_1)=\alpha(t,s)\delta(z_t,z_s)+\beta(t,s)p_0(z_t).
\end{equation}
According to the law of total probability formula, the following identity must be satisfied:
\begin{equation}
    \sum_{z_s}p_{t|s,1}(z_t|z_s,z_1)p_{s|1}(z_s|z_1)=p_{t|1}(z_t|z_1).
\end{equation}
Substituting the assumed form of $p_{t|s,1}$ and the known expression of $p_{s|1}$ into the left-hand side (LHS) of the identity yields:
\begin{align}
    \text{LHS}&=\left[ \alpha(t,s)\delta(z_t, z_s) + \beta(t,s) p_0(z_t) \right] \left[ s\delta(z_s, z_1) + (1-s)p_0(z_s) \right] \\
    &=\sum_{z_s} \alpha(t, s) s \delta(z_t, z_s) \delta(z_s, z_1) \\ 
    &+ \sum_{z_s} \alpha(t,s) (1-s) \delta(z_t, z_s) p_0(z_s) \\ 
    &+ \sum_{z_s} \beta(t,s) s p_0(z_t) \delta(z_s, z_1) \\ 
    &+ \sum_{z_s} \beta(t,s) (1-s) p_0(z_t) p_0(z_s) \\ 
    &= \alpha(t,s) s\delta(z_t, z_1) + [\alpha(t,s)(1-s) + \beta(t,s)]p_0(z_t).
\end{align}
Comparing the coefficients of $\delta(z_t, z_1)$ and $p_0(z_t)$ with the right-hand side (RHS), $t\delta(z_t, z_1) + (1-t)p_0(z_t)$, results in the following system of equations:
\begin{align}
    \begin{cases}
        \alpha(t,s) \cdot s = t \\
        \alpha(t,s)(1-s)+\beta(t,s) = 1-t
    \end{cases}    .
\end{align}
Solving this system gives $\alpha(t,s)=\cfrac{t}{s}$ and $\beta(t,s)=1-\cfrac{t}{s}$.
The resulting expression for $p_{t|s,1}$ is thus:
\begin{equation}
    p_{t|s,1}(z_t|z_s,z_1)=\cfrac{t}{s}\delta(z_t,z_s)+(1-\cfrac{t}{s})p_0(z_t).
\end{equation}
Since this expression is independent of $z_1$, the transition probability $p_{t|s}$ is equivalent to $p_{t|s,1}$:
\begin{align}
    p_{t|s}(z_t|z_s)&=\sum_{z_1}p_{t|s,1}(z_t|z_s,z_1)p_{1|s}(z_1|z_s) \\
    &=\sum_{z_1}[\cfrac{t}{s}\delta(z_t,z_s)+(1-\cfrac{t}{s})p_0(z_t)]p_{1|s}(z_1|z_s) \\
    &=[\cfrac{t}{s}\delta(z_t,z_s)+(1-\cfrac{t}{s})p_0(z_t)] \cdot \sum_{z_1}p_{1|s}(z_1|z_s) \\
    &=\cfrac{t}{s}\delta(z_t,z_s)+(1-\cfrac{t}{s})p_0(z_t).
\end{align}
This confirms the Markovian property of the noising process.
For nodes and edges, the transition probabilities within the noising trajectory are given by:
\begin{equation}
    p_{t|t+\Delta t}(x_t^i|x_{t+\Delta t}^i)=\cfrac{t}{t+\Delta t}\delta(x_{t+\Delta t}^i, x_{t}^i) + (1 - \cfrac{t}{t+\Delta t})p_0(x_t^i),
\end{equation}
\begin{equation}
    p_{t|t+\Delta t}(e_t^{ij}|e_{t+\Delta t}^{ij})=\cfrac{t}{t+\Delta t}\delta(e_{t+\Delta t}^{ij}, e_{t}^{ij}) + (1 - \cfrac{t}{t+\Delta t})p_0(e_t^{ij}).
\end{equation}
The joint transition probability $p_{t|t+\Delta t}(G_t|G_{t+\Delta t})$ is then reconstructed as the product of these individual component-wise probabilities:
\begin{equation}
    p_{t|t+\Delta t}(G_t|G_{t+\Delta t})=\prod_{i}^{k}p_{t|t+\Delta t}(x_t^i|x_{t+\Delta t}^i) \prod_{i,j=1}^{k}p_{t|t+\Delta t}(e_t^{ij}|e_{t+\Delta t}^{ij}).
\end{equation}

\subsection{Reverse Process and Rate Matrix Construction}
The reverse process $p_{t+\Delta t|t}(G_{t+\Delta t}|G_t)$ is assumed to be factorized across nodes and edges, implying that each component evolves according to its own transition rate:
\begin{equation}
    \label{eq:defog_reversep}
    p_{t+\Delta t|t}(G_{t+\Delta t}|G_t)=\prod_{i=1}^{k}p_{t+\Delta t}(x_{t+\Delta t}^i|G_t) \prod_{i,j=1}^{k}p_{t+\Delta t|t}(e_{t+\Delta t}^{ij}|G_t).
\end{equation}
For a sufficiently small time step $\Delta t$, the transition probabilities can be approximated using the transition rates:
\begin{equation}
    p_{t+\Delta t|t}(z_{t+\Delta t}|G_t) = \delta(z_t, z_{t+\Delta t}) + \mathbf{R}_t(z_t, z_{t+\Delta t}) \Delta t,
\end{equation}
where $\mathbf{R}_t$ denotes the time-dependent rate matrix.
In the discrete flow matching framework, the rate matrix $\mathbf{R}_t$ is defined as the expectation of a conditional rate matrix $\mathbf{R}_{t|1}$ over the predicted data posterior $p_{1|t}(z_1|G_t)$:
\begin{align}
    \label{eq:defog_Rt}
    \mathbf{R}_t(z_t, z_{t+\Delta t}) &= \mathbb{E}_{p_{1|t}(z_1|G_t)} \left[ \mathbf{R}_{t|1}(z_t, z_{t+\Delta t} | z_1) \right] \\
    &= \sum_{z_1} p_{1|t}(z_1|G_t) \mathbf{R}_{t|1}(z_t, z_{t+\Delta t}|z_1).
\end{align}
The conditional rate matrix $\mathbf{R}_{t|1}$ must satisfy the Kolmogorov forward equation:
\begin{equation}
    \partial_t p_{t|1} = p_{t|1} \mathbf{R}_{t|1}   . 
\end{equation}
To achieve this, DeFoG constructs $\mathbf{R}_{t|1}$ by combining a particular solution $\mathbf{R}_{t|1}^{*}$ with two additional terms, $\mathbf{R}^{\eta}_{t|1}$ and $\mathbf{R}^{\omega}_{t|1}$, both of which satisfy the detailed balance (DB) condition:
\begin{equation}
    p_{t|1}(z_t|z_1)\mathbf{R}_{t|1}^{\text{DB}}(z_t, z_{t+\Delta t}|z_1) = p_{t|1}(z_{t+\Delta t}|z_1)\mathbf{R}_{t|1}^{\text{DB}}(z_{t+\Delta t}, z_t|z_1).
\end{equation}
Specifically, $\mathbf{R}_{t|1}^{*}$ is defined to capture the primary probability flow:
\begin{equation}
    \mathbf{R}_{t|1}^{*}(z_t, z_{t+\Delta t}|z_1)=\cfrac{\text{ReLU}[\partial_t p_{t|1}(z_{t+\Delta t}|z_1) - \partial_t p_{t|1}(z_t|z_1)]}{Z_t^{>0}p_{t|1}(z_t|z_1)},
\end{equation}
where $Z_t^{>0}=|\{z_t:p_{t|1}(z_t|z_1)>0\}|$, $\delta(\cdot, \cdot)$ is the Kronecker delta function.
Then, the terms $\mathbf{R}^{\eta}_{t|1}$ and $\mathbf{R}^{\omega}_{t|1}$ are formulated as specific instances of $\mathbf{R}_{t|1}^{\text{DB}}$ to ensure stability and exploration:
\begin{equation}
    \mathbf{R}^{\eta}_{t|1}(z_t, z_{t+\Delta t} | z_1)= \eta \cdot p_{t|1}(z_{t+\Delta t}|z_1),
\end{equation}
\begin{equation}
    \mathbf{R}^{\omega}_{t|1}(z_t, z_{t+\Delta t} | z_1) = \omega \cdot \cfrac{\delta(z_{t+\Delta t}, z_1)}{Z_t^{>0}p_{t|1}(z_t|z_1)}.
\end{equation}
The total conditional rate matrix is thus the sum of these three components:
\begin{equation}
    \mathbf{R}_{t|1}(z_t, z_{t+\Delta t} | z_1) = \mathbf{R}_{t|1}^{*}(z_t, z_{t+\Delta t} | z_1)  + \mathbf{R}^{\eta}_{t|1}(z_t, z_{t+\Delta t} | z_1) + \mathbf{R}^{\omega}_{t|1}(z_t, z_{t+\Delta t} | z_1).
\end{equation}
By substituting the probability path derivative $\partial_t p_{t|1}(z_t|z_1) = \delta(z_t, z_1) - p_0(z)$ and the conditional path $p_{t|1}(z|z_1) = t\delta(z_t, z_1) + (1-t)p_0(z_t)$ into the sum, the explicit form is obtained:
\begin{align}
    \label{eq:defog_Rt1}
    \mathbf{R}_{t|1}(z_t, z_{t+\Delta t} | z_1)&=\cfrac{\text{ReLU}[\partial_t p_{t|1}(z_{t+\Delta t}|z_1)-\partial_t p_{t|1}(z_{t}|z_1)]}{Z_t^{>0}p_{t|1}(z_{t}|z_1)} + \eta p_{t|1}(z_{t+\Delta t}|z_1) + \omega \cfrac{\delta(z_{t+\Delta t}, z_1)}{Z_t^{>0}p_{t|1}(z_{t}|z_1)} \\
    &= \cfrac{\text{ReLU}[\delta(z_{t+\Delta t}, z_1)-\delta(z_t, z_1)-p_0(z_{t+\Delta t})+p_0(z_{t})]+\omega \delta(z_{t+\Delta t}, z_1)}{Z_t^{>0}[t\delta(z_{t}, z_1)+(1-t)p_0(z_t)]} \\ &+ \eta \cdot [t \delta(z_{t+\Delta}, z_1) + (1-t)p_0(z_{t+\Delta t})],
\end{align}
where $p_0$ represents the prior distribution (taken as the marginal distribution of the dataset in DeFoG). 
The terms $\eta$ and $\omega$ are two adjustable hyperparameters.

By applying the definition of the expected rate matrix $\mathbf{R}_t$ (Eq. \eqref{eq:defog_Rt}) and the specific form of $\mathbf{R}_{t|1}$ (Eq. \eqref{eq:defog_Rt1}) to each component of the graph, the final transition rates for nodes $x^i$ and edges $e^{ij}$ are expressed as:
\begin{equation}
    \label{eq:defog_Rtx}
    \mathbf{R}_t^i(x_t^i, x_{t+\Delta t}^i)=\sum_{x_1^i}p_{1|t}(x_1^i|G_t)\left[\mathbf{R}_{t|1}^i(x_t^i,x_{t+\Delta t}^i|x_1^i)\right],
\end{equation}
\begin{equation}
    \label{eq:defog_Rte}
    \mathbf{R}_t^{ij}(e_t^{ij}, e_{t+\Delta t}^{ij})=\sum_{e_1^{ij}} p_{1|t}(e_1^i|G_t)\left[\mathbf{R}_{t|1}^{ij}(e_t^{ij},e_{t+\Delta t}^{ij}|e_1^{ij})\right],
\end{equation}
\begin{align}
    \mathbf{R}_{t|1}^i(x_{t}^{i}, x_{t+\Delta t}^{i}|x_1^{i}) &= \cfrac{\text{ReLU}[\partial_t p_{t|1}(x_{t+\Delta t}^i|x_1^i)-\partial_t p_{t|1}(x_{t}^i|x_1^i)]}{Z_t^{>0}p_{t|1}(x_{t}^i|x_1^i)} + \eta p_{t|1}(x_{t+\Delta t}^i|x_1^i) + \omega \cfrac{\delta(x_{t+\Delta t}^i, x_1^i)}{Z_t^{>0}p_{t|1}(x_{t}^i|x_1^i)} \\
    &= \cfrac{\text{ReLU}[\delta(x_{t+\Delta t}^i, x_1^i)-\delta(x_t^i, x_1^i)-p_0(x_{t+\Delta t}^i)+p_0(x_{t}^i)]+\omega \delta(x_{t+\Delta t}^i, x_1^i)}{Z_t^{>0}[t\delta(x_{t}^i, x_1^i)+(1-t)p_0(x_t^i)]} \\ &+ \eta \cdot [t  \delta(x_{t+\Delta}^i, x_1^i) + (1-t)p_0(x_{t+\Delta t}^i)],
\end{align}
\begin{align}
    \mathbf{R}_{t|1}^{ij}(e_{t}^{ij}, e_{t+\Delta t}^{ij}|e_1^{ij}) &= \cfrac{\text{ReLU}[\partial_t p_{t|1}(e_{t+\Delta t}^{ij}|e_1^{ij})-\partial_t p_{t|1}(e_{t}^{ij}|e_1^{ij})]}{Z_t^{>0}p_{t|1}(e_{t}^{ij}|e_1^{ij})} + \eta p_{t|1}(e_{t+\Delta t}^{ij}|e_1^{ij}) + \omega \cfrac{\delta(e_{t+\Delta t}^{ij}, e_1^{ij})}{Z_t^{>0}p_{t|1}(e_{t}^{ij}|e_1^{ij})} \\
    &= \cfrac{\text{ReLU}[\delta(e_{t+\Delta t}^{ij}, e_1^{ij})-\delta(e_t^{ij}, e_1^{ij})-p_0(e_{t+\Delta t}^{ij})+p_0(e_{t}^{ij})]+\omega \delta(e_{t+\Delta t}^{ij}, e_1^{ij})}{Z_t^{>0}[t\delta(e_{t}^{ij}, e_1^{ij})+(1-t)p_0(e_t^{ij})]} \\ &+ \eta \cdot [t \delta(e_{t+\Delta}^{ij}, e_1^{ij}) + (1-t)p_0(e_{t+\Delta t}^{ij})].
\end{align}

While the above expressions provide the exact form of the transition rates, they remain dependent on the unknown data posterior $p_{1|t}(G_1|G_t)$. 
In practice, this posterior cannot be computed analytically for complex graph distributions. 
To address this, DeFoG employs a Graph Transformer $\phi_\theta(G_t)$ to approximate the categorical distributions of the clean graph. 
The network is trained to predict the probabilities for each node and edge, denoted as $\hat{x}_1^i = \phi_{\theta}(G_t)^X_i$ and $\hat{e}_1^{ij} = \phi_{\theta}(G_t)^E_{ij}$, respectively. 
The optimization is guided by a cross-entropy loss:
\begin{equation}
    \mathcal{L}_{\text{DeFoG}}=\mathbb{E}_{t \sim \mathcal{U}(0,1), \mathbf{G}_1 \sim p_1, \mathbf{G}_t \sim p_{t|1}} \left[ \sum_{i=1}^{k} \text{CE}(x_1^i, \hat{x}_1^i) + \lambda\sum_{i,j=1}^{k} \text{CE}(e_1^{ij}, \hat{e}_1^{ij}) \right].
\end{equation}

The sampling process $\mathbf{G}_t \sim p_{t|1}$ in the loss function is governed by the factorized probability path. 
Specifically, $p_{t|1}(G_t|G_1)$ defines how a clean graph is progressively diffused into the prior distribution, and is expressed as the product of independent component-wise transitions:
\begin{equation}
    p_{t|1}(G_t|G_1)=\prod_{i}^{k}p_{t|1}(x_t^i|x_1^i) \prod_{i,j=1}^{k}p_{t|1}(e_t^{ij}|e_1^{ij}).
\end{equation}
The component-wise probability paths are defined as:
\begin{equation}
    p_{t|1}(x_t^i|x_1^i)=t\delta(x_t^i, x_1^i)+(1-t)p_0(x_t^i),
\end{equation}
\begin{equation}
    p_{t|1}(e_t^{ij}|e_1^{ij})=t\delta(e_t^{ij}, e_1^{ij})+(1-t)p_0(e_t^{ij}).
\end{equation}

Once the model $\phi_{\theta}$ is trained, the posterior distributions $p_{1|t}(x_1^i|G_t)$ and $p_{1|t}(e_1^{ij}|G_t)$ are replaced by the network's predicted probabilities $p_{1|t}^{\theta}(x_1^i|G_t)$ and $p_{1|t}^{\theta}(e_1^{ij}|G_t)$, respectively. 
Consequently, the expected rate matrices $\mathbf{R}_t^i$ in Eq. \eqref{eq:defog_Rtx} and $\mathbf{R}_t^{ij}$ in Eq. \eqref{eq:defog_Rte} can be computed using the explicit forms derived above. 
These rates in turn determine the transition probabilities $p_{t+\Delta t|t}(x_{t+\Delta t}^i|G_t)$ and $p_{t+\Delta t|t}(e_{t+\Delta t}^{ij}|G_t)$ in Eq. \eqref{eq:defog_reversep}, thereby completing the estimation of $p_{G}(G_1)$ in Eq. \eqref{eq:defog_pG1}.

\section{Subgraph Sampling Methods}
\label{sec:im_ssm}
\subsection{Acceptance-Rejection Sampling (ARS)}

The Acceptance-Rejection Sampling (ARS) is the most straightforward method for subgraph sampling from a graph set $\mathcal{G}$. 
As shown in Alg. \ref{alg:ARS}, this algorithm first samples a graph $G^{(i)}$ from $\mathcal{G}$ with probability proportional to its edge count. 
Then, $k$ nodes are selected uniformly at random from $G^{(i)}$ to form an induced subgraph. 
If the induced subgraph is connected, it is accepted; otherwise, it is rejected.

While theoretically unbiased, ARS suffers from extreme inefficiency due to its high rejection rate. 
As $k$ increases, the probability of forming a connected subgraph decreases rapidly, making the algorithm computationally impractical for large subgraph sizes.

\begin{algorithm}[htbp]
\caption{Acceptance-Rejection Sampling (ARS)}
\label{alg:ARS}
    \begin{algorithmic}

    \STATE {\bfseries Input:} Raw graph set $\mathcal{G}=\{G^{(1)}, G^{(2)}, \cdots, G^{(|\mathcal{G}|)}\}$, subgraph size $k$, target count $tc$.
    \STATE {\bfseries Output:} Subgraph dataset $\mathcal{D}_k$.
    
    \FUNCTION{ARS($\mathcal{G}$, $k$, $tc$)}
    \STATE Initialize empty set $\mathcal{D}_k=\emptyset$
    \STATE Calculate weights $w_i \gets \cfrac{|E(G^{(i)})|}{\sum_{j=1}^{|\mathcal{G}|}|E(G^{(j)})|}$ \quad // $|E(G^{(i)})|$ is the number of edges for $G^{(i)}$
    \STATE $\mathbf{w} \gets (w_1, w_2, ..., w_{|\mathcal{G}|})$
    \WHILE{$|\mathcal{D}_k| < tc$}
        \STATE Sample $i \sim \text{Categorical}(\mathbf{w})$ 
        \STATE $G \gets G^{(i)}$
        \STATE Select $k$ nodes from $G$ at random
        \STATE Generate the induced subgraph $g$
        \IF{$g$ is connected}
            \STATE $\mathcal{D}_k \gets \mathcal{D}_k \cup \{g\}$
        \ENDIF
    \ENDWHILE
    \STATE {\bfseries return} $\mathcal{D}_k$
    \ENDFUNCTION
    
    \end{algorithmic}

\end{algorithm}

\subsection{Neighbour Reservoir Sampling (NRS)}

The Neighbor Reservoir Sampling (NRS) algorithm \cite{lu_sampling_2012} is designed to enhance sampling efficiency by directly exploring the neighborhood of the current node set. 
As shown in Alg. \ref{alg:NRS}, the algorithm proceeds in two main stages. 
It first expands from a randomly sampled edge until the node set reaches size $k$, and then adopts a Reservoir Sampling mechanism to iteratively update the candidate nodes.

Although NRS is efficient, it should be noted that the sampling process is theoretically biased. 
Despite this bias, its high efficiency and ability to maintain connectivity make it a practical and robust choice for large-scale subgraph sampling tasks.

\begin{algorithm}[htbp]
\caption{Neighbour Reservoir Sampling (NRS)}
\label{alg:NRS}
    \begin{algorithmic}
    \STATE {\bfseries Input:} Raw graph set $\mathcal{G}=\{G^{(1)}, G^{(2)}, \cdots, G^{(|\mathcal{G}|)}\}$, subgraph size $k$, target count $tc$.
    \STATE {\bfseries Output:} Subgraph dataset $\mathcal{D}_k$.
    \FUNCTION{NRS($\mathcal{G}$, $k$, $tc$)}
        \STATE Initialize empty set $\mathcal{D}_k=\emptyset$
        \STATE Calculate weights $w_i \gets \cfrac{|E(G^{(i)})|}{\sum_{j=1}^{|\mathcal{G}|}|E(G^{(j)})|}$ \quad // $|E(G^{(i)})|$ is the number of edges for $G^{(i)}$
        \STATE $\mathbf{w} \gets (w_1, w_2, ..., w_{|\mathcal{G}|})$
        \WHILE{$|\mathcal{D}_k| < tc$}
            \STATE Sample $i \sim \text{Categorical}(\mathbf{w})$ 
            \STATE $G \gets G^{(i)}$
            \STATE $g \gets $ NRSonSingleGraph($G$, $k$)
            \STATE $\mathcal{D}_k \gets \mathcal{D}_k \cup \{g\}$
        \ENDWHILE
        \STATE {\bfseries return} $\mathcal{D}_k$
    \ENDFUNCTION
    \STATE
    \FUNCTION{NRSonSingleGraph($G$, $k$)}
        \STATE Sample an edge $e \in E(G)$ at random
    \STATE $V \gets \{x \in e\}$
    \WHILE{$|V| < k$}
        \STATE $EL \gets \{ (x, y) \in E(G) : (x \in V \land y \notin V) \lor (x \notin V \land y \in V) \}$
        \STATE Sample $e \in EL$ at random
        \STATE $V \gets V \cup \{x \in e \mid x \notin V\}$
    \ENDWHILE

    \STATE $V_u \gets V(G) - V$  \quad // the unprocessed nodes of $G$
    \STATE $EL \gets \{(x, y): (x \in V \land y \in V_u) \lor (x \in V_u \land y \in V)\}$
    
    \STATE $i \gets k$
    \WHILE{$|EL| > 0$}
        \STATE $i \gets i + 1$
        \STATE Sample $e \in EL$ at random
        \STATE $v \gets \{x \in e \mid x \in V_u\}$
        \STATE $V_u \gets V_u - \{v\}$
        \STATE $\alpha \sim \text{Uniform}(0,1)$
        \IF{$\alpha < \cfrac{k}{i}$}
            \STATE Sample $u \in V$ at random
            \STATE $V' \gets (V - \{u\}) \cup \{v\}$
            \STATE $g' \gets $ the induced subgraph of $V'$
            \IF{ $g'$ is connected}
                \STATE $V \gets V'$
            \ENDIF
        \ENDIF
        \STATE $EL \gets \{(x, y): (x \in V \land y \in V_u) \lor (x \in V_u \land y \in V)\}$
    \ENDWHILE
    \STATE $g \gets $ the induced subgraph of $V$ 
    \STATE {\bfseries return} $g$
    \ENDFUNCTION
    \end{algorithmic}
\end{algorithm}

\subsection{Rand-ESU}

The Rand-ESU algorithm \cite{wernicke_efficient_2006} is a randomized version of the Exact Subgraph Enumeration (ESU) method.
As shown in Alg. \ref{alg:Rand_ESU}, it systematically explores a graph by constructing an enumeration tree of depth $k$.
At each level $d$ of the tree, the algorithm decides whether to continue the traversal with probability $p_d \in (0, 1]$.
The mechanism allows for controlling the number of samples through the probability list $p_{1:k}$, where a lower $p_d$ significantly reduces the computational overhead in exchange for a smaller sample set.
The algorithm is theoretically unbiased because each subgraph of size $k$ is associated with a unique path in the enumeration tree, ensuring an equal sampled probability for all subgraphs at the same depth.

\begin{algorithm}[htbp]
\caption{Rand-ESU}
\label{alg:Rand_ESU}
    \begin{algorithmic}
    \STATE {\bfseries Input:} Raw graph set $\mathcal{G}=\{G^{(1)}, G^{(2)}, \cdots, G^{(|\mathcal{G}|)}\}$, subgraph size $k$, target count $tc$, probability list $p_{1:k}$.
    \STATE {\bfseries Output:} Subgraph dataset $\mathcal{D}_k$.

    \STATE
    \STATE \textbf{Notation:}
    \STATE $N(V', G) := \{ u \in V(G) - V' \mid \exists v \in V', (u, v) \in E(G) \lor (v, u) \in E(G) \}$
    \STATE $N_{excl}(w, V_{Sub}, G) := \{ u \in N(\{w\}, G) \mid u \notin V_{Sub} \cup N(V_{Sub}, G) \}$
    \STATE

    \FUNCTION{Rand-ESU($\mathcal{G}$, $k$, $tc$)}
        \STATE Initialize empty set $\mathcal{D}_k=\emptyset$
        \WHILE{$|\mathcal{D}_k| < tc$}
            \FOR{$i = 1 \to |\mathcal{G}|$}
                \STATE $\mathcal{D}_k \gets $ $\mathcal{D}_k \cup$ EnumerateSubgraphs($G$, $k$)
            \ENDFOR
        \ENDWHILE
        \STATE $\mathcal{D}_k \gets $ Select $tc$ elements from $\mathcal{D}_k$ at random
        \STATE {\bfseries return} $\mathcal{D}_k$    
    \ENDFUNCTION
    
    \STATE

    \FUNCTION{EnumerateSubgraphs($G$, $k$)}
        \STATE Initialize $\mathcal{D}_{single} = \emptyset$
        \FOR{each $v \in V(G)$}
            \STATE $V_{Ext} \gets \{u \in N(\{v\}, G): u > v\}$
            \STATE With probability $p_1$, \textbf{call} ExtendSubgraph($\{v\}, V_{Ext}, v, G, k, \mathcal{D}_{single}$)
        \ENDFOR
        \STATE {\bfseries return} $\mathcal{D}_{single}$
    \ENDFUNCTION

    \STATE

    \FUNCTION{ExtendSubgraph($V_{Sub}, V_{Ext}, v, G, k, \mathcal{D}_{single}$)}
        \IF{$|V_{Sub}| = k$}
            \STATE $\mathcal{D}_{single} \gets \mathcal{D}_{single} \cup \{\text{the induced subgraph of } V_{Sub}\}$
            \STATE {\bfseries return}
        \ENDIF
        \WHILE{$V_{Ext} \neq \emptyset$}
            \STATE Remove an arbitrarily chosen node $w$ from $V_{Ext}$
            \STATE $V_{Ext}' \gets V_{Ext} \cup \{u \in N_{excl}(w, V_{Sub}, G): u > v\}$
            \STATE $d \gets |V_{Sub}| + 1$ \quad // The depth of the next potential call
            \STATE With probability $p_d$, \textbf{call} ExtendSubgraph($V_{Sub} \cup \{w\}, V_{Ext}', v, G, k, \mathcal{D}_{single}$)
        \ENDWHILE
    \ENDFUNCTION

    \end{algorithmic}
\end{algorithm}

\subsection{Rand-FaSE}

\begin{algorithm}[htbp]
\caption{Rand-FaSE}
\label{alg:Rand_FaSE}
    \begin{algorithmic}

    \STATE {\bfseries Input:} Graph set $\mathcal{G}=\{G^{(1)},\dots,G^{(|\mathcal{G}|)}\}$, subgraph size $k$, target count $tc$, probability list $p_{1:k}$.
    \STATE {\bfseries Output:} Subgraph dataset $\mathcal{D}_k$ .
    
    \STATE
    \STATE \textbf{Notation:}
    \STATE $N(V', G):=\{ u \in V(G) - V' \mid \exists v \in V', (u, v) \in E(G) \lor (v, u) \in E(G) \}$
    \STATE $N_{excl}(w, V_{Sub}, G):= \{ u \in N(\{w\}, G) \mid u \notin V_{Sub} \cup N(V_{Sub}, G) \}$
    \STATE $\mathcal{T} = (V_{\mathcal{T}}, E_{\mathcal{T}}, \ell)$ (FaSE trie, which is labeld by LS-label)
    
    \STATE
    
    \FUNCTION{Rand-FaSE($\mathcal{G}$, $k$, $tc$)}
        \STATE Initialize empty set $\mathcal{D}_k=\emptyset$, trie $\mathcal{T}$ with root, probability list $p_{1:(k-1)}$
    
        \WHILE{$|\mathcal{D}_k| < tc$}
            \FOR{$i = 1$$ \to |\mathcal{G}|$}
                \STATE EnumerateSubgraphs($G^{(i)}$, $k$, $p_{1:(k-1)}$, $\mathcal{T}$)
            \ENDFOR
            \STATE $\mathcal{D}_k \gets$ Aggregate from leaves of $ \mathcal{T}$
        \ENDWHILE
        \STATE $\mathcal{D}_k \gets $ Select $tc$ elements from $\mathcal{D}_k$ at random
        \STATE {\bfseries return} $\mathcal{D}_k$
    \ENDFUNCTION
    
    \STATE
    
    \FUNCTION{EnumerateSubgraphs($G$, $k$, $p_{1:(k-1)}$, $\mathcal{T}$)}
        \FOR{each $v \in V(G)$}
            \STATE $V_{Sub} \gets \{v\}$; $V_{Ext} \gets \{u \in N(\{v\},G): u > v\}$
            \STATE $d \gets 1$; $q \gets 1$
            \STATE Insert LS-label of $v$ into trie $\mathcal{T}$
            \STATE \textbf{call} ExtendSubgraph($G$, $k$, $V_{Sub}$, $V_{Ext}$, $v$, $\mathcal{T}$, $d$, $q$)
        \ENDFOR
    \ENDFUNCTION
    
    \STATE
    
    \FUNCTION{ExtendSubgraph($G$, $k$, $V_{Sub}$, $V_{Ext}$, $v$, $\mathcal{T}$, $d$, $q$)}
        \STATE $q \gets q \cdot p_d$
        \IF{$|V_{Sub}| = k$}
            \STATE $weight \gets 1 / q$ \quad // Cumulative sampling probability
            \STATE $\sigma \gets \mathrm{Canon}\!\left(G[V_{Sub}]\right)$ \quad //$\sigma$: Canonical signature
            \STATE $\mathcal{T}[\sigma].\text{count} \;\mathrel{+}= weight$
            \STATE {\bfseries return}
        \ENDIF
    
        \WHILE{$V_{Ext} \neq \emptyset$}
            \STATE Remove a node $w$ from $V_{Ext}$
            \STATE With probability $p_{d}$:
                \STATE \quad $\ell \gets \mathrm{LS}(w, V_{Sub})$ \quad // LS-label
                \STATE \quad $u \gets \mathcal{T}_{\mathrm{child}} (u, \ell)$
                \STATE \quad $V_{Ext}' \gets V_{Ext} \cup \{u \in N_{excl}(w, V_{Sub}, G): u > v\}$
                \STATE \quad \textbf{call} ExtendSubgraph($G$, $k$, $V_{Sub} \cup \{w\}$, $V_{Ext}'$, $v$, $\mathcal{T}$, $d+1$, $q$)
        \ENDWHILE
    \ENDFUNCTION

    \end{algorithmic}

\end{algorithm}

The Rand-FaSE algorithm~\cite{paredes_rand-fase_2015} is a randomized subgraph sampling method based on an ESU-style enumeration framework.

As shown in Alg. \ref{alg:Rand_FaSE}, it explores each input graph by constructing an enumeration tree of depth $k$, where nodes correspond to partial subgraphs and edges represent incremental extensions. At each depth $d$ of the tree, the traversal is continued with probability $p_d \in (0,1]$, which enables explicit control over the sampling rate through the probability list $p_{1:(k-1)}$. Smaller values of $p_d$ reduce the computational cost by probabilistically pruning the enumeration tree, at the expense of a smaller set of sampled subgraphs.

During traversal, Rand-FaSE organizes expansion paths using a trie structure indexed by LS-labels, allowing subgraphs with identical structural extensions to be aggregated online. For each sampled subgraph of size $k$, the algorithm records the cumulative sampling probability along its enumeration path and applies inverse-probability weighting at the leaf level. As a result, Rand-FaSE produces unbiased frequency estimates for subgraphs while maintaining efficient exploration of the enumeration tree.

\section{Further Implementation Details}
\label{sec:appendix_further_details}
This section presents more details about the datasets, evaluation protocol, and hyperparameters.

\subsection{Datasets Description}
\label{subsec:dataset_description}

\textbf{NAS-Bench-101.} 
The NAS-Bench-101 dataset~\cite{Ying2019NASBench101TR} is a classic open-source benchmark built by the Google Brain team to provide a reproducible and comparable standard for Neural Architecture Search (NAS). 
The dataset contains a total of 423,624 unique cell-based architectures, each with a maximum of 7 nodes.
These architectures are composed of 5 different operator types. 

\textbf{NAS-Bench-201.}
The NAS-Bench-201 dataset~\cite{Dong020} is another cell-based benchmark with 15,625 unique architectures. 
In this paper, the original edge-based operator representation is converted to a node-based representation, following the method in DiNAS~\cite{asthana2024multiconditioned}. 
After this conversion, each cell contains 8 nodes, and the architectures are composed of 7 distinct operator types. 

\textbf{NAS-Bench-301.}
NAS-Bench-301~\cite{zela2022surrogatenasbenchmarksgoing} is a surrogate-based benchmark for NAS, designed to address the high computational cost of evaluating architectures in a huge search space.
Following the setting in DiNAS~\cite{asthana2024multiconditioned}, this paper collected 10,000 architectures from the DARTS search space. 
In this benchmark, each architecture contains 11 nodes. 
The architectures are composed of 11 different operator types.

\textbf{NAS-Bench-NLP.}
The NAS-Bench-NLP dataset~\cite{9762315} is a benchmark focused on Recurrent Neural Network (RNN) cells for natural language processing tasks.
Following the setting in DiNAS~\cite{asthana2024multiconditioned}, this paper filters this dataset to contain a total of 7258 architectures.
In this benchmark, each architecture contains at most 13 nodes.
The architectures are composed of 9 different operator types.

\textbf{Younger.}
Unlike the previously discussed synthetic benchmarks, the Younger~\cite{Yang2024YoungerTF} dataset is a large-scale, diverse collection of real-world neural architectures. 
The maximum number of nodes for one architecture in the dataset is about 50000, with a total of 14,230 unique architectures, composed of 314 operator types.

\subsection{Evaluation Protocol}
\label{subsec:evaluation_protocol}
The evaluation protocol consists of two primary components: the assessment of the Graph Diffusion Estimator's predictive performance and the evaluation of the search framework's overall effectiveness in identifying frequent subgraphs.

\textbf{Evaluation of the Graph Diffusion Estimator.}
The performance of the estimator is quantified using Spearman's $\rho$ and Kendall's $\tau$.
These metrics evaluate the monotonic relationship between the predicted probabilities and the ground-truth frequencies.

For a specific subgraph size $k$ on a given dataset, the ground-truth frequency $F_i$, for each non-isomorphic subgraph class $i$, is obtained via exhaustive enumeration.
The graph diffusion estimator provides a predicted probability $P_i$ for each class.
\begin{itemize}
    \item For the graph diffusion estimator, $P_i$ represents the direct output of the model.
    \item For sampling-based baselines, $P_i$ is defined as the frequency of class $i$ in the sampled set. If a class is not sampled, $P_i$ is set to 0.
\end{itemize}
Given two sequences of predicted $\mathbf{P}$ and ground-truth frequencies $\mathbf{F}$, both of length $n$, the coefficients are defined as follows:
\begin{itemize}
    \item Spearman's $\rho$: Let $R(P_i)$ and $R(F_i)$ denote the ranks of the predicted probability and the ground-truth frequency within their respective sequences. The coefficient is defined as:
    \begin{equation}
        \rho=1-\cfrac{6\sum_{i=1}^n (R(P_i)-R(F_i))^2}{n(n^2-1)}
    \end{equation}
    \item Kendall's $\tau$: Let $n_c$ and $n_d$ denote the number of concordant and discordant pairs, respectively. For any pair of observations $(P_i, F_i)$ and $(P_j, F_j)$, the pair is concordant if their ranks follow the same order (i.e., $(R(P_i) - R(P_j))(R(F_i) - R(F_j)) > 0$), and discordant otherwise. The coefficient is defined as:
    \begin{equation}
        \tau = \frac{2(n_c - n_d)}{n(n - 1)}
    \end{equation}
\end{itemize}

\textbf{Evaluation of the Overall Framework.}
The effectiveness of the overall framework is assessed by its capacity to discover frequent subgraphs.
The VF2 algorithm, as implemented in the \textit{igraph} library, is employed to perform exact subgraph isomorphism counting to provide the ground-truth counts.
The execution time of VF2 increases exponentially as the subgraph size grows, becoming computationally expensive for larger patterns.
To ensure computational feasibility within a reasonable timeframe, a cutoff time is established for the VF2 algorithm.
Subgraph counts are only recorded if their occurrences are found before the cutoff time.
The performance of the framework is evaluated using the mean and median of the actual counts of the top-$N$ subgraphs identified by the framework.

\begin{itemize}
    \item For the search framework, the top-$N$ subgraphs correspond to the $K$ subgraphs with the highest predicted probabilities from the graph diffusion estimator.
    \item For the sampling-based methods, the top-$N$ are the $N$ most frequent subgraphs observed in the sampled dataset.
\end{itemize}

\subsection{Baselines}
The experiments incorporate five baselines: four sampling-based methods (ARS, NRS, Rand-ESU, and Rand-FaSE) and one VAE-based model, GraphVAE.

For sampling-based methods, performance is measured according to the evaluation protocol in Section ~\ref{subsec:evaluation_protocol}.
For GraphVAE, the encoder produces the mean and variance of the latent distribution for each subgraph $G_0$.
By sampling from the latent space to reconstruct the graph, the Evidence Lower Bound (ELBO) is computed as a surrogate for the log-generative probability.
By averaging ELBOs over multiple latent space samplings, a generative probability is obtained from the perspective of GraphVAE.
This value corresponds to $p_0(G_0)$ in graph diffusion estimators, allowing GraphVAE to be assessed using the same evaluation protocol.

\subsection{Model Architecture}
\label{ma}

The graph diffusion estimator employs a graph-to-graph neural network $\phi_{\theta}$ as its backbone to predict the distribution of nodes and edges.
Given an input graph characterized by node attributes $X \in \mathbb{R}^{n \times a}$ and edge attributes $E \in \mathbb{R}^{n \times n \times b}$, the model outputs transformed features in the same dimensions.

The architecture is based on the GraphTransformer introduced in DiGress~\cite{vignac2023digress}.
The input attributes first pass through an MLP to map them into a latent space. 
Then, several XEyTransformerLayers iteratively update node and edge attributes.
Finally, the output MLP projects these latent representations to the required output dimensions.

To adapt to Directed Acyclic Graphs (DAGs), this study introduces a positional encoding mechanism that is not present in the original DiGress backbone.
Specifically, node levels are calculated based on the topological structure as:
\begin{equation}
    \text{level}(v) = \max_{u \in \mathcal{N}_{in}(v)} (\text{level}(u) + 1)
\end{equation}
Nodes are then sorted according to these levels.
After the initial node attributes $X$ are projected into the latent space via the first MLP layer, absolute positional encodings (PE) derived from this topological ordering are added to the latent representations. 
This modification incorporates the hierarchical information of DAGs into the model.

\subsection{Hardware Setting}
Experiments were run with Intel Xeon E5-2620 v3 CPU, 63GB RAM, and NVIDIA RTX-2080Ti GPUs.

\subsection{Hyperparameters}

For the generative models DisCo~\cite{xu_discrete-state_2024}, DeFoG~\cite{qin2025defog}, and DiGress~\cite{vignac2023digress}, the hyperparameters are primarily kept consistent with the original configurations reported in their respective papers.
However, specific adjustments are made to meet the requirements of this study.
For DisCo, the number of sampling steps is set to 100; $\alpha$ and $\gamma$ are set to 0.8 and 2, respectively.
For DeFoG, the time distortion for the reverse process is configured as \textit{polydec}.
The number of estimation rounds is set to 20 for DisCo and DiGress, and 60 for DeFoG.

For the sampling methods, ARS and NRS are parameter-free.
For Rand-ESU and Rand-FaSE, the sampling probability at depth $d$ for a target subgraph size $k$ is determined by $p_d = (1 - d/(k+1))^r$.
The parameter $r$ controls the sampling density and is assigned based on the dataset and subgraph size.
For NAS-Bench-101, $r$ is set to 1.
For NAS-Bench-201, NAS-Bench-301, and NAS-Bench-NLP, $r$ is set to 0.
For the Younger dataset, if $k\leq 12$, $r$ is set to 3, if $13 \leq k \leq 15$, $r$ is set to 2.5. 
The cutoff time for VF2 algorithm is set to 20 minutes, with the counting process parallelized across 10 CPU cores.

\section{Additional Experimental Results}
\label{sec:appendix_additional_exp}

\subsection{The Number of Sampled Subgraphs}

Table \ref{tab:dataset_size} reports the number of sampled subgraphs required to reach a sampling density of 0.1 across different datasets and sampling methods.

\begin{table*}[t]
    \centering
    \caption{The number of sampled subgraphs required to reach a sampling density of 0.1 across different datasets and sampling methods.}
    \label{tab:dataset_size}
    \begin{tabular}{l cccc cccc}
        \toprule
        \multirow{2}{*}{Dataset} & \multicolumn{4}{c}{Subgraph Size $k=4$} & \multicolumn{4}{c}{Subgraph Size $k=5$} \\ 
        \cmidrule(lr){2-5} \cmidrule(lr){6-9}
        & ARS & NRS & Rand-ESU & Rand-FaSE & ARS & NRS & Rand-ESU & Rand-FaSE \\
        \midrule
        NAS-Bench-101 & 399 & 404 & 568 & 422 & 14642 & 14293 & 16165 & 14807 \\
        NAS-Bench-201 & 286 & 279 & 273 & 286 & 1316 & 1320 & 1323 & 1335 \\
        NAS-Bench-301 & 3126 & 2892 & 3013 & 2797 & 27588 & 25032 & 28335 & 26912 \\
        NAS-Bench-NLP & 1697 & 2076 & 1768 & 1778 & 8511 & 9218 & 8212 & 7475 \\
        Younger & X & 140316 & 171903 & 96692 & X & 108010 & 152780 & 87000 \\
        \bottomrule
    \end{tabular}
\end{table*}

\subsection{Runtime Analysis}

\begin{table*}[t]
    \centering
    \caption{Training time (hours) of different estimators. The sampling method is set to Rand-ESU, and the sampling density is set to 0.1.}
    \label{tab:nas_runtime}
    \begin{tabular}{c cc cc cc}
        \toprule
        \multirow{2}{*}{Dataset} & \multicolumn{3}{c}{Subgraph size $k=4$} & \multicolumn{3}{c}{Subgraph size $k=5$} \\
        \cmidrule(lr){2-4} \cmidrule(lr){5-7}
        & DisCo-E & DeFoG-E & DiGress-E & DisCo-E & DeFoG-E & DiGress-E \\ 
        \midrule
        NAS-Bench-101 & 0.10 & 0.09 & 0.08 & 0.35 & 0.20 & 0.18 \\ 
        NAS-Bench-201 & 0.10 & 0.06 & 0.07 & 0.12 & 0.07 & 0.07 \\
        NAS-Bench-301 & 0.19 & 0.08 & 0.13 & 0.55 & 0.58 & 0.42 \\
        NAS-Bench-NLP & 0.14 & 0.11 & 0.07 & 0.27 & 0.22 & 0.24 \\
        Younger & 5.97 & 2.03 & 1.81 & 5.26 & 2.16 & 2.11 \\
        \bottomrule
    \end{tabular}
\end{table*}

\begin{figure}
    \centering
    \begin{subfigure}{0.45\textwidth}
        \centering
        \includegraphics[width=\linewidth]{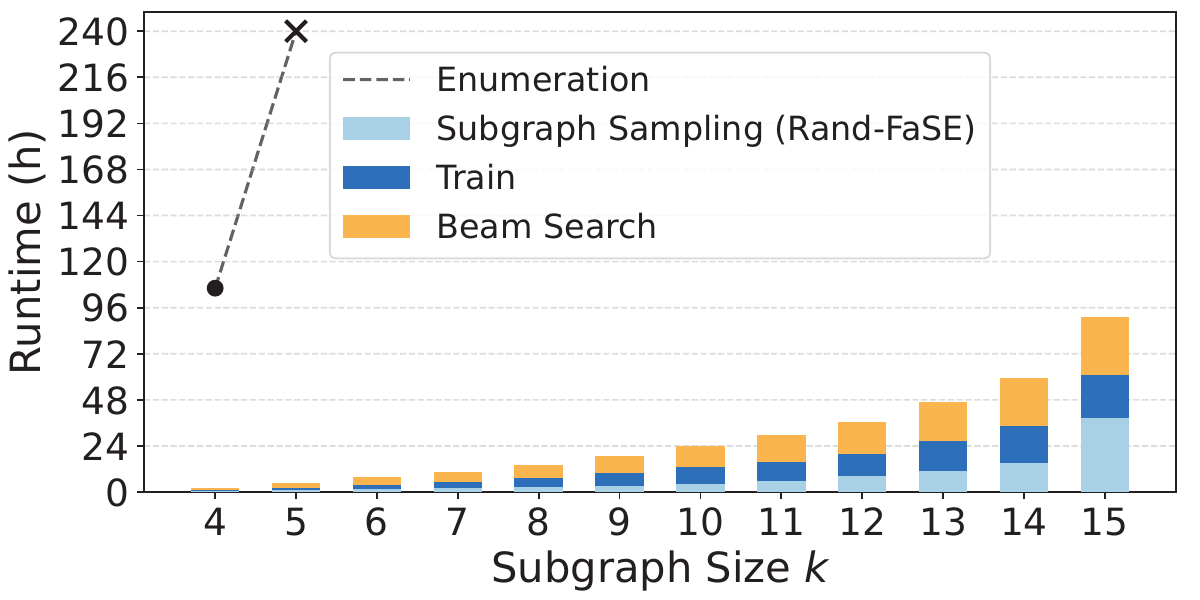}
        \caption{Runtime combination of GraDE (Rand-FaSE).}
        \label{fig:runtime_RFASE}
    \end{subfigure}
    \hfill
    \begin{subfigure}{0.45\textwidth}
        \centering
        \includegraphics[width=\linewidth]{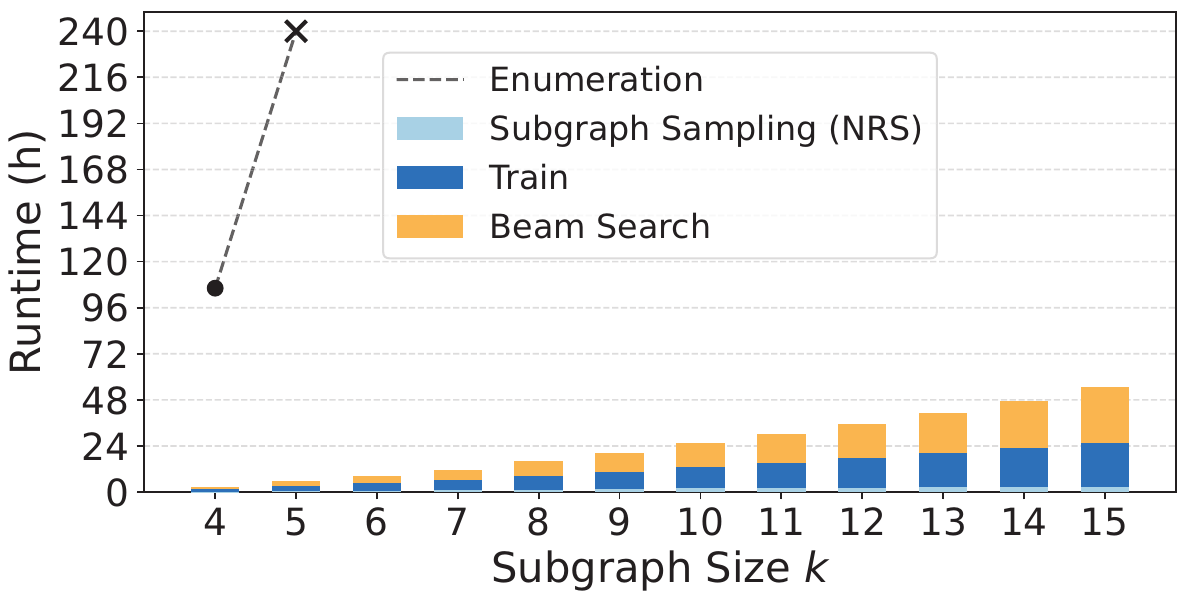}
        \caption{Runtime combination of GraDE (NRS).}
        \label{fig:runtime_NRS}
    \end{subfigure}
    \caption{Runtime combination of GraDE.}
    \label{fig:runtime_GraDE_RFASE_NRS}
\end{figure}

This section evaluates the computational efficiency of the GraDE framework, specifically focusing on the time required for model training and the frequent subgraph discovery process.
Table \ref{tab:nas_runtime} reports the training time for DisCo-E, DeFoG-E, and DiGress-E across various datasets, with the number of training epochs fixed at 200, the sampling method set to Rand-ESU, and the sampling density set to 0.1.

The results indicate that training on NAS-Bench-101, NAS-Bench-201, NAS-Bench-301, and NAS-Bench-NLP is highly efficient, with total training times consistently under one hour.
On the Younger dataset, training requires more time.
This is primarily due to the larger variety of operator types (314) in the Younger dataset architecture space and the increased number of samples needed to maintain a 0.1 sampling density.

The runtime characteristics of GraDE using Rand-FaSE and NRS are illustrated in Figure \ref{fig:runtime_RFASE} and Figure \ref{fig:runtime_NRS}, respectively.
For GraDE(Rand-FaSE), the runtime trend is similar to Rand-ESU: the sampling part takes a larger share as $k$ increases, nearly equaling the training and search time at $k=15$. 
In terms of absolute runtime, Rand-FaSE is generally faster than Rand-ESU.
In contrast, the sampling time for GraDE(NRS) is so short that it is almost negligible compared to the training and search phases.
Since the time for training and searching in GraDE is relatively stable, GraDE(NRS) provides a faster trade-off for users who prioritize speed over discovery capability.
These results further demonstrate the computational feasibility of the GraDE framework.

\subsection{Impact of Sampling Density on Sampling Cost}
\label{subsec:impact_of_sd}

\begin{figure}
    \centering
    \includegraphics[width=0.8\textwidth]{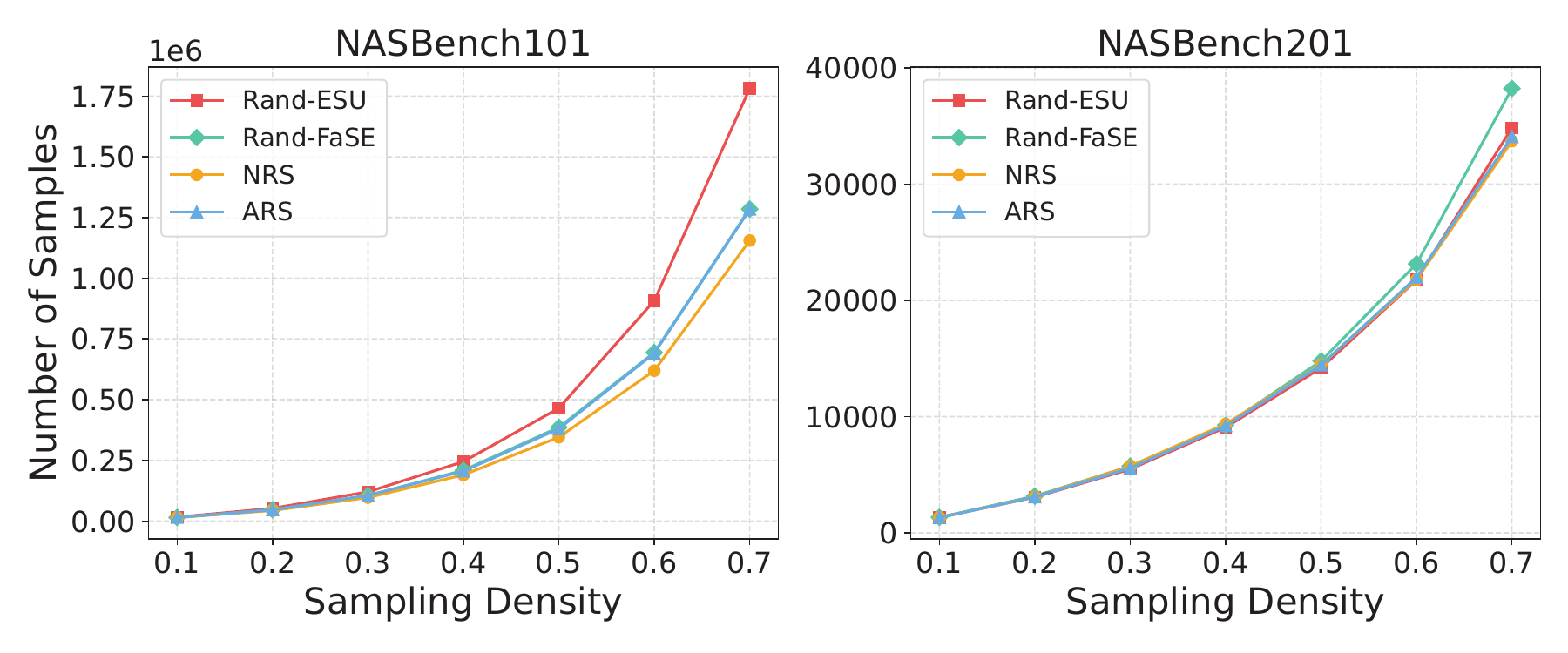}
    \caption{The number of sampled subgraphs vs. sampling density. These results are obtained using Rand-ESU, Rand-FaSE, NRS, and ARS on the NAS-Bench-101 and NAS-Bench-201 datasets.}
    \label{fig:number_of_samples}
\end{figure}

To provide empirical evidence for the relationship between sampling density and sampling cost, an analysis of the required sample sizes was conducted.
This evaluation employs Rand-ESU, Rand-FaSE, NRS, and ARS across the NAS-Bench-101 and NAS-Bench-201 datasets. 
The sampling density in these experiments ranges from 0.1 to 0.7.

Figure \ref{fig:number_of_samples} illustrates the number of sampled subgraphs as a function of sampling density across both datasets.
The results consistently demonstrate that, regardless of the sampling method employed, the total number of required samples grows at an exponential rate as the density increases.
This observation confirms that maintaining a high sampling density leads to prohibitive computational costs.

\subsection{Comparison of Different Methods on Various Datasets}

This section provides a comprehensive evaluation of various frequency estimators across a diverse range of datasets and sampling methods.
In addition to the NAS-Bench-101, NAS-Bench-201, NAS-Bench-301, and Younger datasets discussed in the main text, the experiments here further include results on NAS-Bench-NLP.

Moreover, while Table \ref{tab:estimator} in the main text only presents the performance of different generative models using a single sampling strategy (Rand-ESU), the tables in this section provide a more detailed analysis by showcasing the effectiveness of these methods across all four sampling strategies.
For all evaluations, Spearman's $\rho$ and Kendall's $\tau$ are reported for subgraph sizes $k \in \{4, 5\}$ with a consistent sampling density of 0.1.

Note that due to the large scale of the Younger dataset, the ARS method is computationally infeasible.
Therefore, all corresponding entries for ARS in Table \ref{tab:Younger} are marked with an  ``X'' to indicate the absence of results.

The full results are presented in Tables \ref{tab:101}, \ref{tab:201}, \ref{tab:301}, \ref{tab:NLP}, and \ref{tab:Younger}.
Regarding general performance, DisCo-E achieves the best results under most conditions across the NAS-Bench-101, NAS-Bench-201, NAS-Bench-NLP, and Younger datasets, significantly outperforming sampling-based methods.
On NAS-Bench-301, the performance rankings change depending on the subgraph size $k$.
For $k=4$, DeFoG-E gets the best results in most cases, while DisCo-E and DiGress-E often take the second-best spots.
When the subgraph size increases to $k=5$, DiGress-E becomes the top performer, and DisCo-E usually keeps the second-best position.

The experimental results show that DisCo-E has good generality, as it works well with different types of datasets and different sampling methods.
Specifically, for the DisCo-E estimator, the Rand-ESU sampling strategy consistently gives the best or near-best performance across all NAS datasets compared to other sampling strategies.
This is also true for the Younger dataset, where the combination of DisCo-E and Rand-ESU achieves the best results.
Consequently, taking all conditions into account, DisCo-E paired with Rand-ESU provides the most stable and reliable performance overall.

\begin{table*}[hthp]
    \centering
    \caption{Comparison of different methods with various sampling strategies on NAS-Bench-101.}
    \label{tab:101}
    \begin{tabular}{cll cc cc cc cc}
        \toprule
        \multirow{2}{*}{Size} & \multirow{2}{*}{Estimator} & \multicolumn{2}{c}{ARS} & \multicolumn{2}{c}{NRS} & \multicolumn{2}{c}{Rand-ESU} & \multicolumn{2}{c}{Rand-FaSE}\\
        \cmidrule(lr){3-4} \cmidrule(lr){5-6} \cmidrule(lr){7-8} \cmidrule(lr){9-10}
        & & $\rho$ $\uparrow$ & $\tau$ $\uparrow$ & $\rho$ $\uparrow$ & $\tau$ $\uparrow$ & $\rho$ $\uparrow$ & $\tau$ $\uparrow$ & $\rho$ $\uparrow$ & $\tau$ $\uparrow$ \\
        \midrule
        & Sampling-only & 0.332 & 0.273 & 0.349 & 0.287 & 0.388 & 0.319 & 0.331 & 0.272 \\
        & GraphVAE & 0.379 & 0.255 & 0.412 & 0.281 & 0.378 & 0.256 & 0.383 & 0.259 \\
        \rowcolor[gray]{0.9}
        4 & DiGress-E & 0.545 & 0.376 & 0.346 & 0.232 & 0.692 & 0.501 & 0.621 & 0.436 \\
        \rowcolor[gray]{0.9}
        & DeFoG-E & \underline{0.694} & \underline{0.499} & \underline{0.783} & \underline{0.575} & \underline{0.827} & \underline{0.653} & \underline{0.836} & \underline{0.654} \\
        \rowcolor[gray]{0.9}
        & DisCo-E & \textbf{0.869} & \textbf{0.686} & \textbf{0.872} & \textbf{0.688} & \textbf{0.860} & \textbf{0.681} & \textbf{0.877} & \textbf{0.699} \\
        \midrule  
        & Sampling-only & 0.426 & 0.366 & 0.422 & 0.362 & 0.442 & 0.379 & 0.418 & 0.358 \\
        & GraphVAE & 0.391 & 0.276 & 0.399 & 0.282 & 0.424 & 0.303 & 0.426 & 0.305 \\
        \rowcolor[gray]{0.9}
        5 & DiGress-E & 0.831 & 0.671 & 0.842 & 0.682 & 0.877 & 0.725 & 0.833 & 0.666  \\
        \rowcolor[gray]{0.9}
        & DeFoG-E & \underline{0.864} & \underline{0.698} & \underline{0.865} & \underline{0.699} & \underline{0.899} & \underline{0.741} & \underline{0.862} & \underline{0.694} \\
        \rowcolor[gray]{0.9}
        & DisCo-E & \textbf{0.896} & \textbf{0.749} & \textbf{0.898} & \textbf{0.750} & \textbf{0.938} & \textbf{0.809} & \textbf{0.885} & \textbf{0.726} \\
        \bottomrule
    \end{tabular}
\end{table*}

\begin{table*}[htbp]
    \centering
    \caption{Comparison of different methods with various sampling strategies on NAS-Bench-201.}
    \label{tab:201}
    \begin{tabular}{cll cc cc cc cc}
        \toprule
        \multirow{2}{*}{Size} & \multirow{2}{*}{Estimator} & \multicolumn{2}{c}{ARS} & \multicolumn{2}{c}{NRS} & \multicolumn{2}{c}{Rand-ESU} & \multicolumn{2}{c}{Rand-FaSE}\\
        \cmidrule(lr){3-4} \cmidrule(lr){5-6} \cmidrule(lr){7-8} \cmidrule(lr){9-10}
        & & $\rho$ $\uparrow$ & $\tau$ $\uparrow$ & $\rho$ $\uparrow$ & $\tau$ $\uparrow$ & $\rho$ $\uparrow$ & $\tau$ $\uparrow$ & $\rho$ $\uparrow$ & $\tau$ $\uparrow$ \\
        \midrule
        & Sampling-only & 0.400 & 0.373 & 0.386 & 0.360 & 0.402 & 0.375 & 0.417 & 0.389 \\
        & GraphVAE & 0.581 & 0.460 & 0.449 & 0.351 & 0.255 & 0.195 & 0.402 & 0.314 \\
        \rowcolor[gray]{0.9}
        4 & DiGress-E & 0.676 & 0.540 & 0.742 & 0.561 & \underline{0.791} & 0.592 & \underline{0.812} & \underline{0.654} \\
        \rowcolor[gray]{0.9}
        & DeFoG-E & \underline{0.795} & \underline{0.641} & \textbf{0.820} & \underline{0.662} & \textbf{0.815} & \underline{0.620} & 0.748 & 0.604  \\
        \rowcolor[gray]{0.9}
        & DisCo-E & \textbf{0.859} & \textbf{0.717} & \underline{0.819} & \textbf{0.677} & 0.778 & \textbf{0.641} & \textbf{0.851} & \textbf{0.709} \\
        \midrule  
        & Sampling-only & 0.306 & 0.279 & 0.308 & 0.279 & 0.332 & 0.302 & 0.296 & 0.269 \\
        & GraphVAE & 0.683 & 0.544 & 0.667 & 0.529 & 0.701 & 0.562& 0.714 & 0.573  \\
        \rowcolor[gray]{0.9}
        5 & DiGress-E & \textbf{0.868} & \textbf{0.723} & \underline{0.868} & \underline{0.725} & \underline{0.845} & \underline{0.698} & \underline{0.766} & \underline{0.611}  \\
        \rowcolor[gray]{0.9}
        & DeFoG-E & 0.552 & 0.423 & 0.554 & 0.427 & 0.607 & 0.471  & 0.490 & 0.372 \\
        \rowcolor[gray]{0.9}
        & DisCo-E & \underline{0.845} & \underline{0.700} & \textbf{0.873} & \textbf{0.731} & \textbf{0.884} & \textbf{0.746} & \textbf{0.772} & \textbf{0.622}  \\
        \bottomrule
    \end{tabular}
\end{table*}

\begin{table*}[htbp]
    \centering
    \caption{Comparison of different methods with various sampling strategies on NAS-Bench-301.}
    \label{tab:301}
    \begin{tabular}{cll cc cc cc cc}
        \toprule
        \multirow{2}{*}{Size} & \multirow{2}{*}{Estimator} & \multicolumn{2}{c}{ARS} & \multicolumn{2}{c}{NRS} & \multicolumn{2}{c}{Rand-ESU} & \multicolumn{2}{c}{Rand-FaSE}\\
        \cmidrule(lr){3-4} \cmidrule(lr){5-6} \cmidrule(lr){7-8} \cmidrule(lr){9-10}
        & & $\rho$ $\uparrow$ & $\tau$ $\uparrow$ & $\rho$ $\uparrow$ & $\tau$ $\uparrow$ & $\rho$ $\uparrow$ & $\tau$ $\uparrow$ & $\rho$ $\uparrow$ & $\tau$ $\uparrow$ \\
        \midrule
        & Sampling-only & 0.418 & 0.347 & 0.417 & 0.346 & 0.428 & 0.356 & 0.431 & 0.357 \\
        & GraphVAE & 0.179 & 0.117 & 0.201 & 0.134 & 0.220 & 0.146 & 0.277 & 0.179 \\
        \rowcolor[gray]{0.9}
        4 & DiGress-E & \underline{0.789} & \textbf{0.622} & 0.119 & 0.073 & \underline{0.673} & \underline{0.493} & \underline{0.802} & \underline{0.632} \\
        \rowcolor[gray]{0.9}
        & DeFoG-E & \textbf{0.797} & \underline{0.598} & \textbf{0.797} & \textbf{0.602} & 0.545 & 0.395 & \textbf{0.852} & \textbf{0.664} \\
        \rowcolor[gray]{0.9}
        & DisCo-E & 0.755 & 0.553 & \underline{0.777} & \underline{0.578} & \textbf{0.740} & \textbf{0.544} & 0.774 & 0.578 \\
        \midrule  
        & Sampling-only & 0.421 & 0.362 & 0.414 & 0.356 & 0.423 & 0.364 & 0.428 & 0.368 \\
        & GraphVAE & 0.460 & 0.349 & 0.470 & 0.356 & 0.462 & 0.347  & 0.486 & 0.367  \\
        \rowcolor[gray]{0.9}
        5 & DiGress-E & \textbf{0.805} & \textbf{0.632} & \textbf{0.803} & \textbf{0.630} & \textbf{0.816} & \textbf{0.648} & \underline{0.795} & \textbf{0.626}  \\
        \rowcolor[gray]{0.9}
        & DeFoG-E & 0.730 & 0.551 & 0.732 & 0.551 & 0.714 & 0.537 & 0.736 & 0.560 \\
        \rowcolor[gray]{0.9}
        & DisCo-E & \underline{0.793} & \underline{0.606} & \underline{0.783} & \underline{0.596} & \underline{0.789} & \underline{0.600} & \textbf{0.798} & \underline{0.616} \\
        \bottomrule
    \end{tabular}
\end{table*}

\begin{table*}[htbp]
    \centering
    \caption{Comparison of different methods with various sampling strategies on NAS-Bench-NLP.}
    \label{tab:NLP}
    \begin{tabular}{cll cc cc cc cc}
        \toprule
        \multirow{2}{*}{Size} & \multirow{2}{*}{Estimator} & \multicolumn{2}{c}{ARS} & \multicolumn{2}{c}{NRS} & \multicolumn{2}{c}{Rand-ESU} & \multicolumn{2}{c}{Rand-FaSE}\\
        \cmidrule(lr){3-4} \cmidrule(lr){5-6} \cmidrule(lr){7-8} \cmidrule(lr){9-10}
        & & $\rho$ $\uparrow$ & $\tau$ $\uparrow$ & $\rho$ $\uparrow$ & $\tau$ $\uparrow$ & $\rho$ $\uparrow$ & $\tau$ $\uparrow$ & $\rho$ $\uparrow$ & $\tau$ $\uparrow$ \\
        \midrule
        & Sampling-only & 0.429 & 0.368 & 0.434 & 0.373 & 0.433 & 0.371 & 0.425 & 0.364 \\
        & GraphVAE & 0.493 & 0.360 & 0.474 & 0.346 & 0.466 & 0.338 & 0.459 & 0.332 \\
        \rowcolor[gray]{0.9}
        4 & DiGress-E & 0.543 & 0.403 & 0.283 & 0.204 & 0.398 & 0.290 & 0.469 & 0.342 \\
        \rowcolor[gray]{0.9}
        & DeFoG-E & \underline{0.570} & \underline{0.421} & \underline{0.610} & \underline{0.456} & \underline{0.569} & \underline{0.424} & \underline{0.590} & \underline{0.440} \\
        \rowcolor[gray]{0.9}
        & DisCo-E & \textbf{0.677} & \textbf{0.515} & \textbf{0.678} & \textbf{0.517} & \textbf{0.674} & \textbf{0.513} & \textbf{0.679} & \textbf{0.518} \\
        \midrule  
        & Sampling-only & 0.406 & 0.365 & 0.411 & 0.369 & 0.400 & 0.360 & 0.383 & 0.343 \\
        & GraphVAE & 0.433 & 0.326 & 0.438 & 0.330 & 0.433 & 0.326 & 0.433 & 0.326  \\
        \rowcolor[gray]{0.9}
        5 & DiGress-E & 0.536 & 0.410 & 0.572 & 0.442 & \underline{0.584} & \underline{0.452} & 0.519 & 0.397  \\
        \rowcolor[gray]{0.9}
        & DeFoG-E & \underline{0.564} & \underline{0.432} & \underline{0.577} & \underline{0.443} & 0.568 & 0.437 & \underline{0.557} & \underline{0.427} \\
        \rowcolor[gray]{0.9}
        & DisCo-E & \textbf{0.637} & \textbf{0.496} & \textbf{0.624} & \textbf{0.485} & \textbf{0.660} & \textbf{0.517} & \textbf{0.621} & \textbf{0.482} \\
        \bottomrule
    \end{tabular}
\end{table*}

\begin{table*}[htbp]
    \centering
    \caption{Comparison of different methods with various sampling strategies on Younger.}
    \label{tab:Younger}
    \begin{tabular}{cll cc cc cc cc}
        \toprule
        \multirow{2}{*}{Size} & \multirow{2}{*}{Estimator} & \multicolumn{2}{c}{ARS} & \multicolumn{2}{c}{NRS} & \multicolumn{2}{c}{Rand-ESU} & \multicolumn{2}{c}{Rand-FaSE}\\
        \cmidrule(lr){3-4} \cmidrule(lr){5-6} \cmidrule(lr){7-8} \cmidrule(lr){9-10}
        & & $\rho$ $\uparrow$ & $\tau$ $\uparrow$ & $\rho$ $\uparrow$ & $\tau$ $\uparrow$ & $\rho$ $\uparrow$ & $\tau$ $\uparrow$ & $\rho$ $\uparrow$ & $\tau$ $\uparrow$ \\
        \midrule 
        & Sampling-only & X & X & \underline{0.454} & \underline{0.380} & \underline{0.458} & \underline{0.384} & \underline{0.350} & \underline{0.290}  \\
        & GraphVAE & X & X & 0.375 & 0.262 & 0.378 & 0.264 & 0.301 & 0.209 \\
        \rowcolor[gray]{0.9}
        4 & DiGress-E & X & X & 0.358 & 0.251 & 0.374 & 0.262 & 0.113 & 0.077 \\
        \rowcolor[gray]{0.9}
        & DeFoG-E & X & X & 0.367 & 0.259 & 0.415 & 0.293 & 0.285 & 0.198 \\
        \rowcolor[gray]{0.9}
        & DisCo-E & X & X & \textbf{0.556} & \textbf{0.402} & \textbf{0.558} & \textbf{0.404} & \textbf{0.423} & \textbf{0.299} \\
        \midrule  
        & Sampling-only & X & X & 0.414 & \underline{0.347} & 0.416 & \underline{0.349} & 0.320 & \underline{0.267} \\
        & GraphVAE & X & X & 0.331 & 0.231 & 0.320 & 0.223 & 0.264 & 0.182  \\
        \rowcolor[gray]{0.9}
        5 & DiGress-E & X & X  & 0.314 & 0.218 & 0.392 & 0.275 & 0.051 & 0.035  \\
        \rowcolor[gray]{0.9}
        & DeFoG-E & X & X & \underline{0.453} & 0.320 & \underline{0.470} & 0.333 & \underline{0.343} & 0.239  \\
        \rowcolor[gray]{0.9}
        & DisCo-E & X & X & \textbf{0.547} & \textbf{0.394} & \textbf{0.559} & \textbf{0.404} & \textbf{0.440} & \textbf{0.311} \\
        \bottomrule
    \end{tabular}
\end{table*}

\subsection{Discovery Capability of the GraDE Framework}
\begin{figure}
    \centering
    \begin{subfigure}{0.45\textwidth}
        \centering
        \includegraphics[width=\linewidth]{figs/top50_mid.pdf}
        \caption{Median frequencies of top-50 discovered subgraphs.}
        \label{fig:app_FSM_mid}
    \end{subfigure}
    \hfill
    \begin{subfigure}{0.45\textwidth}
        \centering
        \includegraphics[width=\linewidth]{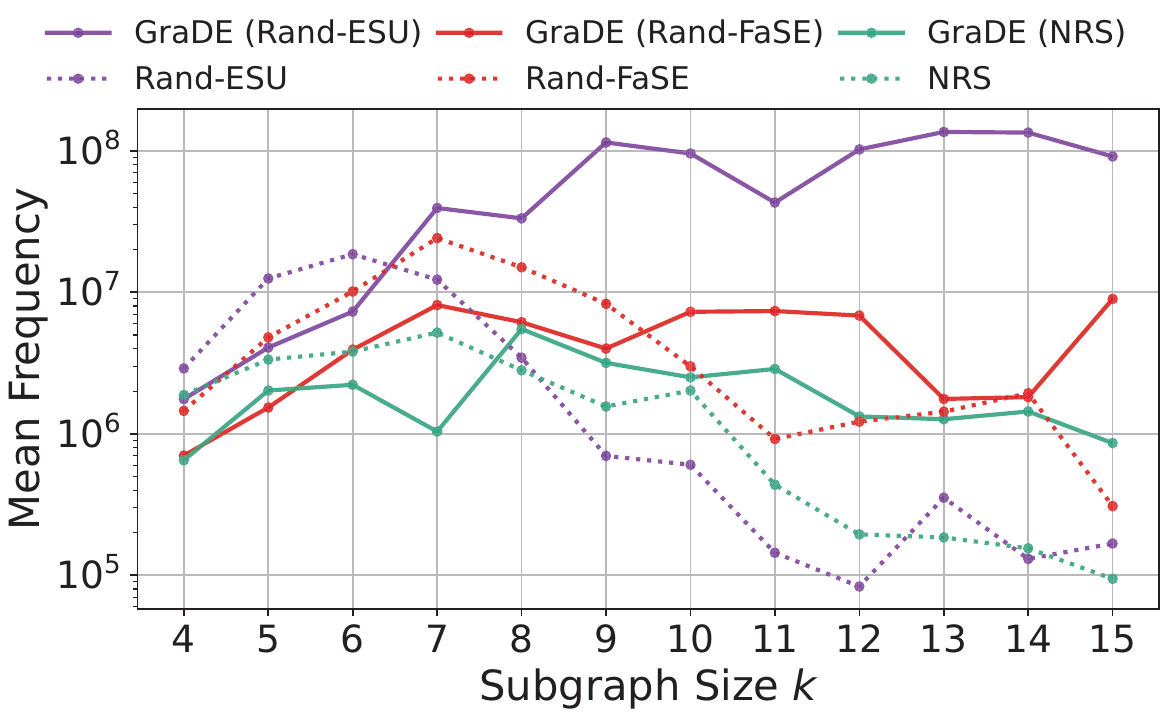}
        \caption{Mean frequencies of top-50 discovered subgraphs.}
        \label{fig:app_FSM_mean}
    \end{subfigure}
    \caption{Median and mean frequencies of top-50 discovered subgraphs. GraDE(Rand-ESU), GraDE(Rand-FaSE), and GraDE(NRS) denote the GraDE framework with corresponding sampling methods.}
    \label{fig:app_FSM}
\end{figure}

This section provides an extended analysis of the discovery performance of GraDE, supplementing the results discussed in Section \ref{subsec:fsm}.
Due to space constraints in the main text, only the median frequencies were highlighted.
Therefore, a more comprehensive statistical evaluation is presented here.
Figure~\ref{fig:app_FSM_mid} includes the previously reported median frequencies for completeness, while Figure~\ref{fig:app_FSM_mean} illustrates the mean frequencies of the top-50 discovered subgraphs.

As shown in Figure \ref{fig:app_FSM_mid} and Figure \ref{fig:app_FSM_mean}, a performance crossover is observed as $k$ increases. 
While pure sampling-based methods exhibit better performance for smaller $k$, GraDE begins to gain a substantial advantage as $k$ grows, especially for $k \ge 10$.
In this range ($k \ge 10$), GraDE (Rand-ESU) outperforms Rand-ESU by an average of over 100$\times$ across both mean and median frequencies. 
Similarly, GraDE (Rand-FaSE) and GraDE (NRS) achieve over 5$\times$ and 2.5$\times$ average improvements over their respective sampling-based methods.

Notably, GraDE(Rand-ESU) emerges as the best configuration starting from $k = 8$. As $k$ further increases, the gap between GraDE (Rand-ESU) and the best sampling-based methods widens significantly. 
Specifically, in the large-scale scenarios ($k \ge 10$), this advantage exceeds 30$\times$ across all cases. 
These results demonstrate that the GraDE framework possesses a substantial advantage in discovering large-scale frequent subgraph patterns, effectively mitigating the limitations of sampling-based methods in massive search spaces.

\subsection{Impact of Cutoff Time}

This section investigates the sensitivity of the evaluation results to the cutoff time employed in the VF2 algorithm.
Since exact frequency counting for large subgraphs is often computationally prohibitive, this analysis determines whether the relative performance between methods remains consistent under different cutoff times.

\begin{figure}[htbp]
    \centering
    \begin{subfigure}[b]{0.48\textwidth}
        \includegraphics[width=\textwidth]{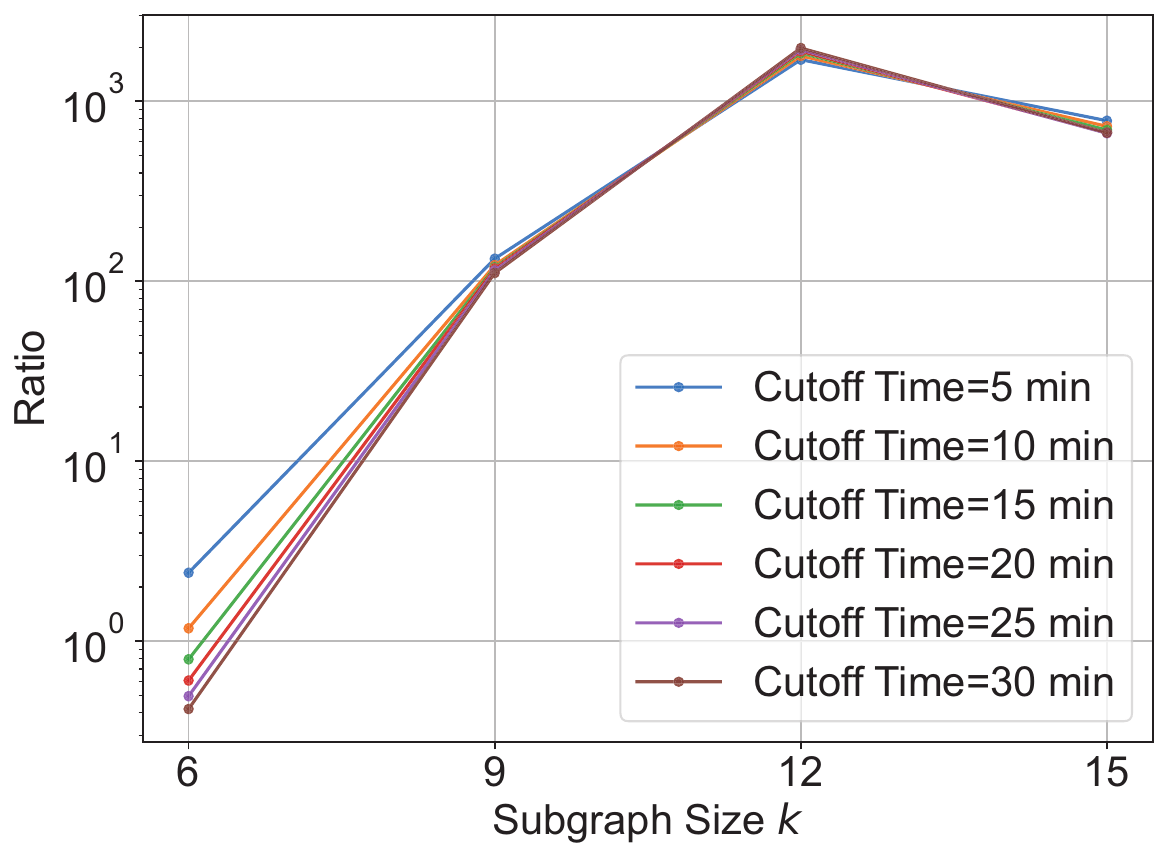}
        \caption{Mean frequency ratio.}
    \end{subfigure}
    \hfill
    \begin{subfigure}[b]{0.48\textwidth}
        \includegraphics[width=\textwidth]{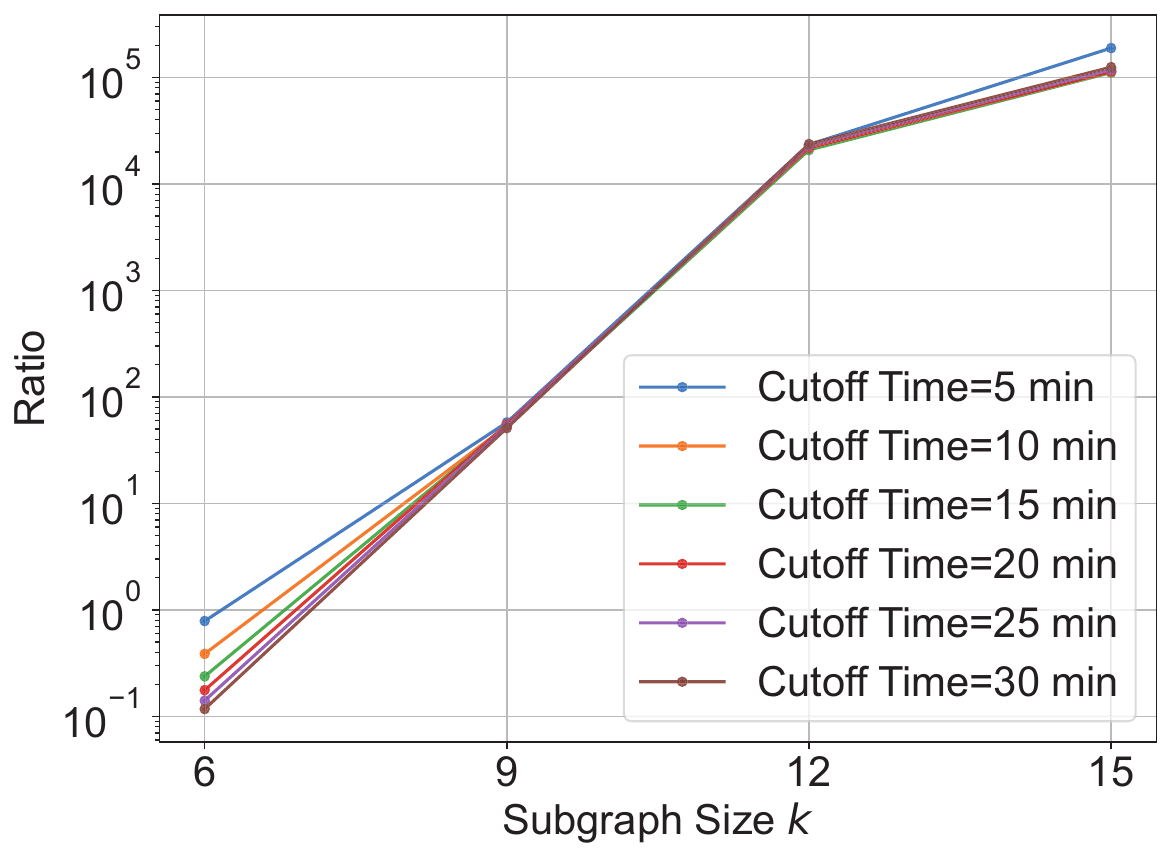}
        \caption{Median frequency ratio.}
    \end{subfigure}
    \caption{Sensitivity analysis of frequency ratios across different VF2 cutoff times (5--30\, min) on the Younger dataset. Each line represents a specific cutoff time across various subgraph sizes $k\in\{6, 9, 12, 15\}$.}
    \label{fig:sensitivity}
\end{figure}

The experiment evaluates the top 50 subgraphs discovered by GraDE (with Rand-ESU) and those found by the sampling-based method (Rand-ESU).
The mean and median frequencies of these subgraphs are calculated using the VF2 algorithm across six different cutoff times: 5, 10, 15, 20, 25, and 30 minutes.
Figure \ref{fig:sensitivity} illustrates the ratio of the frequencies (GraDE vs. sampling-based method) across different subgraph sizes $k \in \{6, 9, 12, 15\}$.

The results indicate that the frequency ratios across all six cutoff time settings are remarkably consistent, particularly as the subgraph size increases.
This stability suggests that while absolute frequency counts may be affected by the cutoff time, the relative frequency advantage of the subgraphs discovered by GraDE remains unchanged.
Therefore, a 20-minute cutoff time is considered a reasonable and reliable threshold for comparing the performance of different methods.

\subsection{Case Study: From Controllable Discovery to Practical Applications}

The incremental nature of the GraDE framework allows for the constrained search, offering significant efficiency and controllability over sampling-based methods.
In GraDE, domain-specific constraints can be integrated as pruning rules during the expansion process.
Specifically, any candidate subgraph that violates the predefined constraints is discarded immediately, preventing the search from exploring uninterested branches.
In contrast, sampling-based methods can only perform post-hoc filtering, where uninterested structures are removed only after the entire sampling process is finished.

This section considers two common structural constraints: (1) a Node-type Constraint, requiring the subgraph to contain at least one convolutional operation; and (2) an Edge Connectivity Constraint, requiring the in-degree and out-degree of every node in the subgraph to be at most 2.
Table \ref{tab:case1} reports the proportion of subgraphs that simultaneously satisfy both constraints among the raw samples collected on the Younger dataset by various sampling-based methods across different subgraph sizes $k$.

\begin{table}[htbp]
\centering
\caption{Proportion of interested samples satisfying both node-type and connectivity constraints.}
\label{tab:case1}
    \begin{tabular}{lccccccc}
    \toprule
    Size $k$ & 4 & 5 & 6 & 7 & 8 & 9 & 10 \\
    \midrule
    NRS       & 3.4150\% & 3.8375\% & 4.1500\% &  3.9800\% &  3.9950\% & 3.8500\% & 4.0075\% \\
    Rand-ESU  & 2.5800\% & 1.1950\% & 0.4175\% & 0.1275\% & 0.0350\% & 0.0200\% & 0.0025\% \\
    Rand-FaSE & 15.2550\% & 13.4950\% & 8.4250\% & 3.7575\% & 1.7525\% & 0.6900\% & 0.2900\% \\
    \bottomrule
    \end{tabular}
\end{table}

\begin{figure}[htbp]
    \centering
    \begin{subfigure}[b]{0.21\textwidth}
        \includegraphics[width=\textwidth]{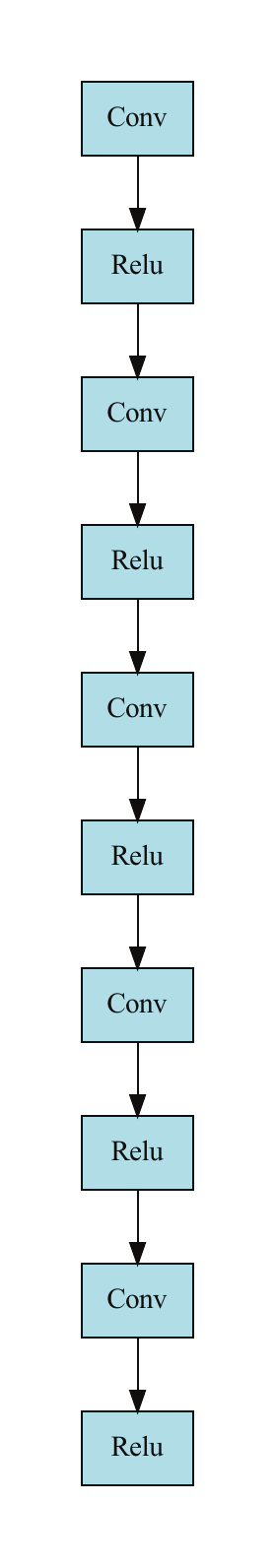}
        \caption{Case A}
        \label{fig:caseA}
    \end{subfigure}
    \hfill
    \begin{subfigure}[b]{0.32\textwidth}
        \includegraphics[width=\textwidth]{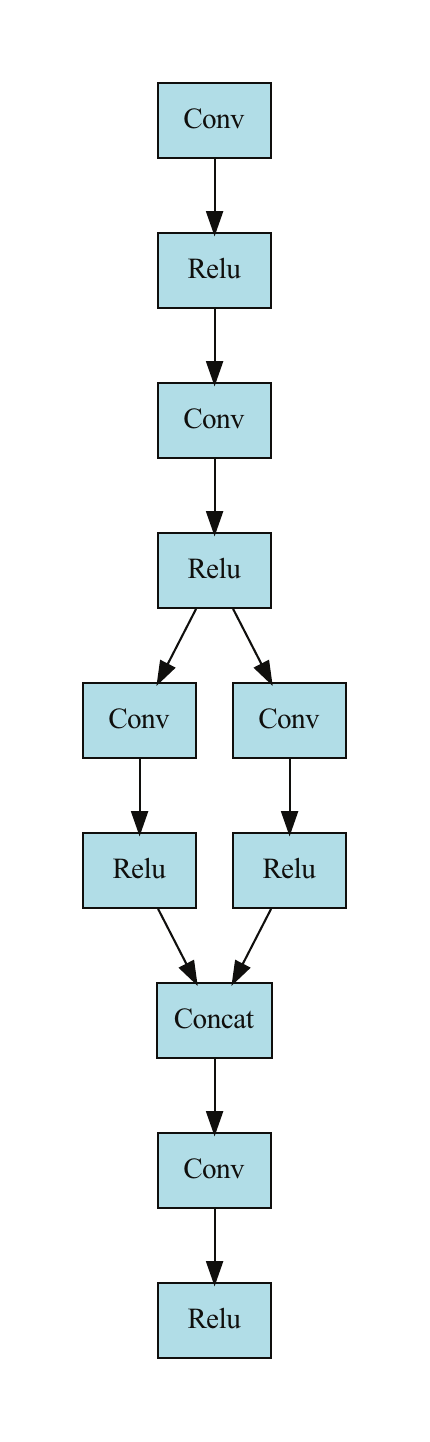}
        \caption{Case B}
        \label{fig:caseB}
    \end{subfigure}
    \hfill
    \begin{subfigure}[b]{0.21\textwidth}
        \includegraphics[width=\textwidth]{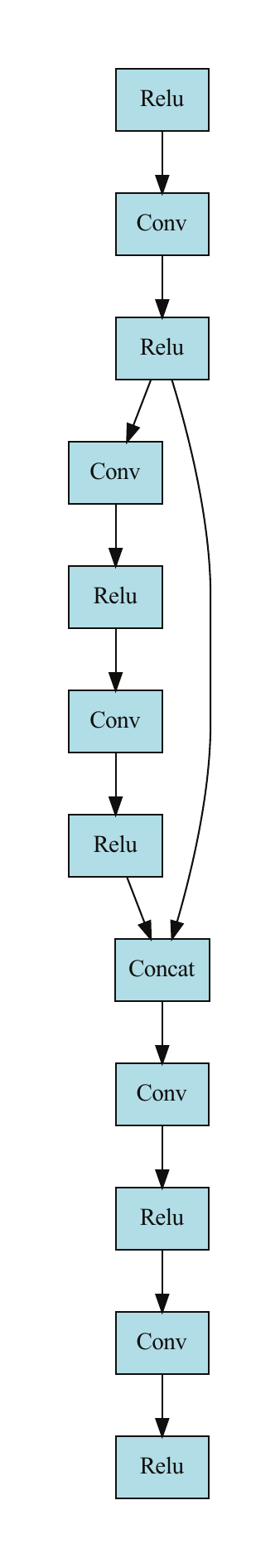}
        \caption{Case C}
        \label{fig:caseC}
    \end{subfigure}

    \caption{Discovered subgraph patterns by constrained GraDE framework on Younger.}
    \label{fig:cases}
\end{figure}

As shown in Table \ref{tab:case1}, the proportion of interested samples obtained by sampling-based methods is remarkably low, especially as the subgraph size increases.
This highlights the extreme sparsity of constrained structures in the overall architecture space, making post-hoc filtering highly inefficient.
In contrast, GraDE guarantees that 100\% of the discovered subgraphs satisfy the predefined constraints, as uninterested candidates are pruned at the earliest possible stage of the search process.
As shown in Figure \ref{fig:cases}, three typical cases are discovered by the aforementioned constrained GraDE framework.
These frequent structures can serve as building blocks for neural network construction, which helps to accelerate the overall design process.

To further evaluate the practical utility of the discovered patterns, the typical cases in Figure \ref{fig:cases} are implemented as PyTorch modules.
These implementations strictly map the structural topologies identified by GraDE to concrete functional blocks, as shown in the following code:
{
\scriptsize
\begin{verbatim}
class CaseAModule(nn.Module):
    def __init__(self):
        super().__init__()
        self.conv1 = nn.Conv2d(64, 64, kernel_size=3, padding=1, bias=False)
        self.conv2 = nn.Conv2d(64, 64, kernel_size=3, padding=1, bias=False)
        self.conv3 = nn.Conv2d(64, 64, kernel_size=3, padding=1, bias=False)
        self.conv4 = nn.Conv2d(64, 64, kernel_size=3, padding=1, bias=False)
        self.conv5 = nn.Conv2d(64, 64, kernel_size=3, padding=1, bias=False)

    def forward(self, x):
        x = F.relu(self.conv1(x))
        x = F.relu(self.conv2(x))
        x = F.relu(self.conv3(x))
        x = F.relu(self.conv4(x))
        x = F.relu(self.conv5(x))
        return x

class CaseBModule(nn.Module):
    def __init__(self):
        super().__init__()
        self.conv1 = nn.Conv2d(64, 64, kernel_size=3, padding=1, bias=False)
        self.conv2 = nn.Conv2d(64, 64, kernel_size=3, padding=1, bias=False)
        self.conv3 = nn.Conv2d(64, 64, kernel_size=3, padding=1, bias=False)
        self.conv4 = nn.Conv2d(64, 64, kernel_size=3, padding=1, bias=False)
        self.conv5 = nn.Conv2d(128, 64, kernel_size=3, padding=1, bias=False)

    def forward(self, x):
        x = F.relu(self.conv1(x))
        split_input = F.relu(self.conv2(x))
        branch_l = F.relu(self.conv3(split_input))
        branch_r = F.relu(self.conv4(split_input))
        concat = torch.cat([branch_l, branch_r], dim=1)
        out = F.relu(self.conv5(concat))
        return out      

class CaseCModule(nn.Module):
    def __init__(self):
        super().__init__()
        self.conv1 = nn.Conv2d(64, 64, kernel_size=3, padding=1, bias=False)
        self.conv2 = nn.Conv2d(64, 64, kernel_size=3, padding=1, bias=False)
        self.conv3 = nn.Conv2d(64, 64, kernel_size=3, padding=1, bias=False)
        self.conv4 = nn.Conv2d(128, 64, kernel_size=3, padding=1, bias=False)
        self.conv5 = nn.Conv2d(64, 64, kernel_size=3, padding=1, bias=False)

    def forward(self, x):
        top_relu = F.relu(x)
        re_before_split = F.relu(self.conv1(top_relu))
        bl1_r = F.relu(self.conv2(re_before_split))
        bl2_r = F.relu(self.conv3(bl1_r))
        cat = torch.cat([bl2_r, re_before_split], dim=1)
        re4 = F.relu(self.conv4(cat))
        res = F.relu(self.conv5(re4))
        return res
\end{verbatim}
}

The computational efficiency of these modules is evaluated using TVM v0.18.0~\cite{TVM} on an NVIDIA RTX 2080Ti GPU.
For each case, the input resolution is 256, the number of input channels is 64, the kernel size is 3, and the batch size is 32.
Three optimization strategies are compared: (1) Unoptimized, representing the raw graph execution without any TVM optimization; (2) Unfused, where operators are optimized independently; (3) Manual Fusion, where specific structures like Conv and ReLU are manually combined before undergoing TVM's operator-level optimization; and (4) Automated Fusion, where the entire subgraph is treated as a single block for TVM's global optimization.

The experimental results, summarized in Table \ref{tab:tvm_results}, reveal several key insights.
First, optimization is essential; compared to the unoptimized baseline, all TVM-optimized strategies achieve significant speedups across all cases.
Second, subgraph-level optimization is essential, as both fusion-based strategies consistently outperform the unfused baseline.
For instance, in Case B, manual fusion achieves a $1.27\times$ speedup over the unfused version, reducing execution latency from 151.56 ms to 119.54 ms.
Third, manual fusion often yields superior results compared to automated fusion.
In Case B, manual fusion results in 119.54 ms latency, whereas automated fusion only reaches 139.77 ms, with similar trends observed in other cases.

These findings demonstrate that while automated compilers are powerful, significant performance gains can be achieved through manual intervention in specific structural patterns.
Since manual optimization involves substantial costs, it is most practical to apply it to the most critical and frequent structures. 
Consequently, the ability of GraDE to identify these frequent subgraphs provides a vital foundation for system optimization, ensuring manual efforts are focused where they yield the highest impact.

\begin{table}[htbp]
    \centering
    \caption{Inference latency (ms) under different TVM optimization strategies.} \label{tab:tvm_results}
    \begin{tabular}{lccccc}
        \toprule Subgraph & Unoptimized & Unfused & Manual Fusion & Automated Fusion & Max Speedup \\
        \midrule 
        Case A (Fig. \ref{fig:caseA}) & 135.06 & 106.16 & 96.18 & 98.72 & 1.40$\times$ \\
        Case B (Fig. \ref{fig:caseB}) & 162.21 & 151.56 & 119.54 & 139.77 & 1.36$\times$ \\
        Case C (Fig. \ref{fig:caseC}) & 164.60 & 129.49 & 112.42 & 124.83 & 1.46$\times$ \\
        \bottomrule 
    \end{tabular} 
\end{table}

\section{Similar but Distinct Work}
\label{sec:appendix_similar}
To further clarify the positioning of GraDE, this section distinguishes the framework from recent learning-based approaches that exhibit superficial similarities but address different research goals.
SPMiner~\cite{ying_representation_2024} employs a node-anchored frequency definition and is restricted to unattributed graphs, whereas GraDE targets graph-level frequency and supports attributed graphs.
MotiFiesta~\cite{oliver_approximate_2022} considers subgraphs with minor structural variations as the same motif, which deviates from the rigorous exact isomorphism required in this work.
Micro-Graph~\cite{Micro_Graph} similarly relies on approximate motifs rather than exact structural isomorphism.
Another work~\cite{vieira_studying_2025} focuses on small subgraphs ($k \in \{3, 4\}$), lacking the scalability to handle larger subgraphs.

%% file: main.bib
@article{network_motifs,
    author = {R. Milo  and S. Shen-Orr  and S. Itzkovitz  and N. Kashtan  and D. Chklovskii  and U. Alon },
    title = {Network Motifs: Simple Building Blocks of Complex Networks},
    journal = {Science},
    volume = {298},
    number = {5594},
    pages = {824-827},
    year = {2002}
}

@article{hocevar_combinatorial_2014,
	title = {A combinatorial approach to graphlet counting},
	volume = {30},
	number = {4},
	journal = {Bioinformatics},
	author = {Hočevar, Tomaž and Demšar, Janez},
	year = {2014},
	pages = {559--565},
}

@misc{ying_representation_2024,
	title = {Representation Learning for Frequent Subgraph Mining},
	publisher = {arXiv},
	author = {Ying, Rex and Fu, Tianyu and Wang, Andrew and You, Jiaxuan and Wang, Yu and Leskovec, Jure},
	year = {2024},
	note = {arXiv:2402.14367}
}

@article{ribeiro_survey_2022,
	title = {A Survey on Subgraph Counting: Concepts, Algorithms, and Applications to Network Motifs and Graphlets},
	volume = {54},
	number = {2},
	journal = {ACM Computing Surveys},
	author = {Ribeiro, Pedro and Paredes, Pedro and Silva, Miguel E. P. and Aparicio, David and Silva, Fernando},
	year = {2022},
	pages = {1--36},
}

@inproceedings{NiuSSZGE20,
  author       = {Chenhao Niu and
                  Yang Song and
                  Jiaming Song and
                  Shengjia Zhao and
                  Aditya Grover and
                  Stefano Ermon},
  title        = {Permutation Invariant Graph Generation via Score-Based Generative Modeling},
  booktitle    = {International Conference on Artificial Intelligence and Statistics ({AISTATS})},
  year         = {2020},
}

@inproceedings{Shi2021LearningGF,
  title={Learning Gradient Fields for Molecular Conformation Generation},
  author={Chence Shi and Shitong Luo and Minkai Xu and Jian Tang},
  booktitle={International Conference on Machine Learning ({ICML})},
  year={2021},
}

@inproceedings{JoLH22,
  author       = {Jaehyeong Jo and
                  Seul Lee and
                  Sung Ju Hwang},
  title        = {Score-based Generative Modeling of Graphs via the System of Stochastic Differential Equations},
  booktitle    = {International Conference on Machine Learning ({ICML})},
  year         = {2022},
}

@inproceedings{vignac2023digress,
    title={{DiGress}: Discrete Denoising diffusion for graph generation},
    author={Clement Vignac and Igor Krawczuk and Antoine Siraudin and Bohan Wang and Volkan Cevher and Pascal Frossard},
    booktitle={International Conference on Learning Representations (ICLR)},
    year={2023},
}

@inproceedings{xu_discrete-state_2024,
	title = {Discrete-state {Continuous}-time {Diffusion} for {Graph} {Generation}},
	booktitle = {Advances in {Neural} {Information} {Processing} {Systems} ({NeurIPS})},
	author = {Xu, Zhe and Qiu, Ruizhong and Chen, Yuzhong and Chen, Huiyuan and Fan, Xiran and Pan, Menghai and Zeng, Zhichen and Das, Mahashweta and Tong, Hanghang},
	year = {2024},
}

@inproceedings{qin2025defog,
    title={{DeFoG}: Discrete Flow Matching for Graph Generation},
    author={Yiming Qin and Manuel Madeira and Dorina Thanou and Pascal Frossard},
    booktitle={International Conference on Machine Learning ({ICML})},
    year={2025},
}

@inproceedings{lu_sampling_2012,
	title = {Sampling {Connected} {Induced} {Subgraphs} {Uniformly} at {Random}},
	booktitle = {International Conference on Scientific and {Statistical} {Database} {Management} ({SSDBM})},
	author = {Lu, Xuesong and Bressan, Stéphane},
	year = {2012},
}

@article{wernicke_efficient_2006,
	title = {Efficient {Detection} of {Network} {Motifs}},
	volume = {3},
	number = {4},
	journal = {IEEE/ACM Transactions on Computational Biology and Bioinformatics},
	author = {Wernicke, Sebastian},
	year = {2006},
	pages = {347--359},
}

@inproceedings{ribeiro_efficient_2010,
	title = {Efficient {Subgraph} {Frequency} {Estimation} with {G}-{Tries}},
	booktitle = {Algorithms in {Bioinformatics}},
	author = {Ribeiro, Pedro and Silva, Fernando},
	year = {2010},
	pages = {238--249},
}

@article{paredes_rand-fase_2015,
	title = {Rand-{FaSE}: Fast approximate subgraph census},
	volume = {5},
	number = {1},
	journal = {Social Network Analysis and Mining},
	author = {Paredes, Pedro and Ribeiro, Pedro},
	year = {2015},
	pages = {17},
}

@inproceedings{salameh_autogo_2023,
	title = {{AutoGO}: {Automated} Computation Graph Optimization for Neural Network Evolution},
	booktitle = {Advances in {Neural} {Information} {Processing} {Systems} (NeurIPS)},
	author = {Salameh, Mohammad and Mills, Keith and Hassanpour, Negar and Han, Fred and Zhang, Shuting and Lu, Wei and Jui, Shangling and ZHOU, CHUNHUA and Sun, Fengyu and Niu, Di},
	year = {2023},
}

@inproceedings{Plus,
    author = {Wu, Ruofan and Zheng, Zhen and Zhang, Feng and Liu, Chuanjie and Pan, Zaifeng and Zhai, Jidong and Du, Xiaoyong},
    title = {{PluS}: Highly efficient and expandable ML compiler with pluggable graph schedules},
    year = {2025},
    booktitle = {USENIX Annual Technical Conference (USENIX ATC)},
    articleno = {39},
    numpages = {17},
}

@InProceedings{pmlr-v119-li20c,
  title = 	 {Neural Architecture Search in A Proxy Validation Loss Landscape},
  author =       {Li, Yanxi and Dong, Minjing and Wang, Yunhe and Xu, Chang},
  booktitle = 	 {International Conference on Machine Learning (ICML)},
  year = 	 {2020}
}

@ARTICLE{10013693,
  author={Li, Jianxin and Sun, Qingyun and Peng, Hao and Yang, Beining and Wu, Jia and Yu, Philip S.},
  journal={IEEE Transactions on Pattern Analysis and Machine Intelligence (TPAMI)}, 
  title={Adaptive Subgraph Neural Network With Reinforced Critical Structure Mining}, 
  year={2023},
  volume={45},
  number={7},
  pages={8063-8080},
}

@inproceedings{wan2022on,
    title={On Redundancy and Diversity in Cell-based Neural Architecture Search},
    author={Xingchen Wan and Binxin Ru and Pedro M Esperan{\c{c}}a and Zhenguo Li},
    booktitle={International Conference on Learning Representations (ICLR)},
    year={2022},
}

@inproceedings{yu2025mage,
    title={{MAGE}: Model-Level Graph Neural Networks Explanations via Motif-based Graph Generation},
    author={Zhaoning Yu and Hongyang Gao},
    booktitle={International Conference on Learning Representations (ICLR)},
    year={2025},
}

@inproceedings{Ying2019NASBench101TR,
  title={{NAS-Bench-101}: Towards Reproducible Neural Architecture Search},
  author={Chris Ying and Aaron Klein and Esteban Real and Eric Christiansen and Kevin P. Murphy and Frank Hutter},
  booktitle={International Conference on Machine Learning (ICML)},
  year={2019}
}

@inproceedings{Dong020,
  author       = {Xuanyi Dong and
                  Yi Yang},
  title        = {{NAS-Bench-201}: Extending the Scope of Reproducible Neural Architecture
                  Search},
  booktitle    = {International Conference on Learning Representations ({ICLR})},
  year         = {2020},
}

@ARTICLE{9762315,
  author={Klyuchnikov, Nikita and Trofimov, Ilya and Artemova, Ekaterina and Salnikov, Mikhail and Fedorov, Maxim and Filippov, Alexander and Burnaev, Evgeny},
  journal={IEEE Access}, 
  title={{NAS-Bench-NLP}: Neural Architecture Search Benchmark for Natural Language Processing}, 
  year={2022},
  volume={10},
  number={},
  pages={45736-45747}
}

@inproceedings{zela2022surrogatenasbenchmarksgoing,
    title={Surrogate {NAS} Benchmarks: Going Beyond the Limited Search Spaces of Tabular {NAS} Benchmarks},
    author={Arber Zela and Julien Niklas Siems and Lucas Zimmer and Jovita Lukasik and Margret Keuper and Frank Hutter},
    booktitle={International Conference on Learning Representations (ICLR)},
    year={2022}
}

@article{Yang2024YoungerTF,
  title={Younger: The First Dataset for Artificial Intelligence-Generated Neural Network Architecture},
  author={Zhengxin Yang and Wanling Gao and Luzhou Peng and Yunyou Huang and Fei Tang and Jianfeng Zhan},
  journal={ArXiv},
  year={2024},
  volume={abs/2406.15132},
}

@inproceedings{Gtries,
	author = {Ribeiro, Pedro and Silva, Fernando},
	title = {{G-tries}: An efficient data structure for discovering network motifs},
	year = {2010},
	booktitle = {ACM Symposium on Applied Computing ({SAC})},
}

@INPROCEEDINGS{FASE,
  author={Paredes, Pedro and Ribeiro, Pedro},
  booktitle={2013 IEEE/ACM International Conference on Advances in Social Networks Analysis and Mining (ASONAM)}, 
  title={Towards a faster network-centric subgraph census}, 
  year={2013},
}

@ARTICLE{SCMD,
  author={Wang, Jianxin and Huang, Yuannan and Wu, Fang-Xiang and Pan, Yi},
  journal={IEEE/ACM Transactions on Computational Biology and Bioinformatics}, 
  title={Symmetry Compression Method for Discovering Network Motifs}, 
  year={2012},
  volume={9},
  number={6},
  pages={1776-1789}
}

@article{kashtan_efficient_2004,
	title = {Efficient sampling algorithm for estimating subgraph concentrations and detecting network motifs},
	volume = {20},
	language = {en},
	number = {11},
	journal = {Bioinformatics},
	author = {Kashtan, N. and Itzkovitz, S. and Milo, R. and Alon, U.},
	month = jul,
	year = {2004},
	pages = {1746--1758},
}

@inproceedings{saha_finding_2015,
	title = {Finding Network Motifs Using {MCMC} Sampling},
	booktitle = {Complex {Networks} {VI}},
	author = {Saha, Tanay Kumar and Hasan, Mohammad Al},
	year = {2015},
	pages = {13--24},
}

@inproceedings{Fan2023GenerativeDM,
  title={Generative Diffusion Models on Graphs: Methods and Applications},
  author={Wenqi Fan and Cheng-Yun Karen Liu and Yunqing Liu and Jiatong Li and Hang Li and Hui Liu and Jiliang Tang and Qing Li},
  booktitle={International Joint Conference on Artificial Intelligence},
  year={2023},
}

@inproceedings{Simonovsky2018GraphVAETG,
  title={{GraphVAE}: Towards Generation of Small Graphs Using Variational Autoencoders},
  author={Martin Simonovsky and Nikos Komodakis},
  booktitle={International Conference on Artificial Neural Networks (ICANN)},
  year={2018},
}

@inproceedings{You2018GraphRNNGR,
  title={{GraphRNN}: Generating Realistic Graphs with Deep Auto-regressive Models},
  author={Jiaxuan You and Rex Ying and Xiang Ren and William L. Hamilton and Jure Leskovec},
  booktitle={International Conference on Machine Learning (ICML)},
  year={2018},
}

@inproceedings{BojchevskiSZG18,
  author       = {Aleksandar Bojchevski and
                  Oleksandr Shchur and
                  Daniel Z{\"{u}}gner and
                  Stephan G{\"{u}}nnemann},
  title        = {{NetGAN}: Generating Graphs via Random Walks},
  booktitle    = {International Conference on Machine Learning ({ICML})},
  year         = {2018},
}

@article{DeCao2018MolGANAI,
  title={{MolGAN}: An implicit generative model for small molecular graphs},
  author={Nicola De Cao and Thomas Kipf},
  journal={ArXiv},
  year={2018},
  volume={abs/1805.11973},
}

@inproceedings{LiuKBKS19,
  author       = {Jenny Liu and
                  Aviral Kumar and
                  Jimmy Ba and
                  Jamie Kiros and
                  Kevin Swersky},
  title        = {Graph Normalizing Flows},
  booktitle    = {Advances in Neural Information Processing Systems ({NeurIPS})},
  year         = {2019},
}

@article{siraudin2025cometh,
	title={Cometh: A continuous-time discrete-state graph diffusion model},
	author={Antoine Siraudin and Fragkiskos D. Malliaros and Christopher Morris},
	journal={Transactions on Machine Learning Research},
	year={2025},
}

@inproceedings{DFM,
    author = {Campbell, Andrew and Yim, Jason and Barzilay, Regina and Rainforth, Tom and Jaakkola, Tommi},
    title = {Generative flows on discrete state-spaces: Enabling multimodal flows with applications to protein co-design},
    year = {2024},
    booktitle = {International Conference on Machine Learning (ICML)},
}

@article{asthana2024multiconditioned,
    title={Multi-conditioned Graph Diffusion for Neural Architecture Search},
    author={Rohan Asthana and Joschua Conrad and Youssef Dawoud and Maurits Ortmanns and Vasileios Belagiannis},
    journal={Transactions on Machine Learning Research},
    year={2024},
}

@inproceedings{TVM,
    author = {Chen, Tianqi and Moreau, Thierry and Jiang, Ziheng and Zheng, Lianmin and Yan, Eddie and Cowan, Meghan and Shen, Haichen and Wang, Leyuan and Hu, Yuwei and Ceze, Luis and Guestrin, Carlos and Krishnamurthy, Arvind},
    title = {{TVM}: An automated end-to-end optimizing compiler for deep learning},
    year = {2018},
    booktitle = {USENIX Conference on Operating Systems Design and Implementation (OSDI)},
}

@ARTICLE{VF2,
  author={Cordella, L.P. and Foggia, P. and Sansone, C. and Vento, M.},
  journal={IEEE Transactions on Pattern Analysis and Machine Intelligence (TPAMI)}, 
  title={A (sub)graph isomorphism algorithm for matching large graphs}, 
  year={2004},
  volume={26},
  number={10},
  pages={1367-1372}
}

@article{jensen1906fonctions,
  title={Sur les fonctions convexes et les in{\'e}galit{\'e}s entre les valeurs moyennes},
  author={Jensen, Johan Ludwig William Valdemar},
  journal={Acta mathematica},
  volume={30},
  number={1},
  pages={175--193},
  year={1906},
  publisher={Springer}
}

@ARTICLE{Micro_Graph,
  author={Zhang, Shichang and Hu, Ziniu and Subramonian, Arjun and Sun, Yizhou},
  journal={IEEE Transactions on Knowledge and Data Engineering}, 
  title={Motif-Driven Contrastive Learning of Graph Representations}, 
  year={2024},
  volume={36},
  number={8},
  pages={4063-4075}
}

@misc{oliver_approximate_2022,
	title = {Approximate Network Motif Mining Via Graph Learning},
	publisher = {arXiv},
	author = {Oliver, Carlos and Chen, Dexiong and Mallet, Vincent and Philippopoulos, Pericles and Borgwardt, Karsten},
	year = {2022},
	note = {arXiv:2206.01008}
}

@misc{vieira_studying_2025,
	title = {Studying and Improving Graph Neural Network-based Motif Estimation},
	publisher = {arXiv},
	author = {Vieira, Pedro C. and Silva, Miguel E. P. and Ribeiro, Pedro Manuel Pinto},
	year = {2025},
	note = {arXiv:2506.15709}
}
